\documentclass[]{zgca}

\usepackage{amsmath}
\usepackage{amssymb}
\usepackage{array}
\usepackage{enumitem}
\usepackage{booktabs}
\usepackage{graphicx}
\usepackage{subcaption}
\usepackage{hyperref}
\AtBeginDocument{\providecommand{\doi}{}%
  \renewcommand{\doi}[1]{\href{https://doi.org/#1}{doi: #1}}}
\usepackage{multirow}
\usepackage{colortbl}
\usepackage{tabularx}
\usepackage{xcolor}
\usepackage{wrapfig}
\usepackage{float}
\usepackage{setspace}
\usepackage{needspace}
\usepackage{tcolorbox}
\tcbuselibrary{skins,breakable}
\usetikzlibrary{positioning,arrows.meta,calc}

\newcolumntype{L}[1]{>{\raggedright\arraybackslash}p{#1}}
\newcolumntype{C}[1]{>{\centering\arraybackslash}p{#1}}
\newcolumntype{M}[1]{>{\centering\arraybackslash}m{#1}}
\newcolumntype{Z}{>{\raggedright\arraybackslash}X}

\definecolor{findingbg}{HTML}{F4F7FA}
\definecolor{findingborder}{HTML}{415882}
\newtcolorbox{findingbox}{
  enhanced, breakable,
  colback=findingbg, colframe=findingborder,
  boxrule=0.9pt, arc=3pt,
  left=10pt, right=10pt, top=7pt, bottom=7pt,
}

\newcommand{\zgcm}{ZGCM-1}

\newcommand{\zgcmeval}{ZGCM-1-7B}
\newcommand{\para}[1]{\noindent\textbf{#1}}
\definecolor{TableBestBlue}{HTML}{B8D7EA}
\definecolor{TableSecondBlue}{HTML}{D8E7F1}
\definecolor{TableGroupBlue}{HTML}{EEF5F8}
\newtcolorbox{casebox}[1]{
  enhanced, breakable,
  colback=TableGroupBlue, colframe=findingborder,
  boxrule=0.6pt, arc=2pt,
  title=\textbf{#1}, fonttitle=\small,
  left=7pt, right=7pt, top=5pt, bottom=5pt,
}
\newcommand{\bestscore}[1]{\cellcolor{TableBestBlue}\textbf{#1}}
\newcommand{\secondscore}[1]{\cellcolor{TableSecondBlue}#1}

\newcommand{\Rmnum}[1]{\uppercase\expandafter{\romannumeral #1}}
\crefname{section}{Section}{Sections}
\crefname{subsection}{Section}{Sections}
\crefname{subsubsection}{Section}{Sections}
\crefname{figure}{Figure}{Figures}
\crefname{subfigure}{Figure}{Figures}
\crefname{table}{Table}{Tables}
\crefname{subtable}{Table}{Tables}
\crefname{equation}{Equation}{Equations}
\crefname{appendix}{Appendix}{Appendices}

\title{ZGCM-1: A Fully Open and Extremely Efficient Foundation Model for Math and Agentic Search}

\author[1,2]{ZGCM Team}
\affiliation[1]{Zhongguancun Academy}
\affiliation[2]{Zhongguancun Institute of Artificial Intelligence}

\abstract{
  
While foundation models continue to push the frontiers of mathematical reasoning and agentic problem solving, the broader academic community has been largely excluded from this progress due to prohibitive compute requirements and closed training recipes. In this work, we present \textbf{ZGCM-1}, a fully open 7B dense foundation model trained from scratch with extreme data, system, and algorithmic efficiency. ZGCM-1 is founded on a core premise: compact models cannot passively memorize the open web, but can overcome parametric capacity limits by coupling deliberate internal thinking with active external tool use. To support this paradigm across a 256K context, we develop an end-to-end, high-efficiency open training recipe: (1) Architecture \& System Co-design: interleaved gated sliding-window and full attention, and a stable FP8 Muon optimizer; (2) Progressive Curriculum \& MDP Mid-Training: context scaling across 16K, 64K, and 256K, and the reformulation of interaction traces into Markov Decision Processes. Furthermore, we establish an AI-native R\&D workflow where agent swarms autonomously manage cluster operations, data curation, and rapid diagnostic evaluation. Extensive evaluations show that ZGCM-1-7B is competitive across 7B model family on general benchmarks. On several challenging mathematical reasoning and agentic search suites, it remains competitive with frontier models orders of magnitude larger, such as Qwen3-235B-A22B and GLM-5.1. We also show that our pre-training design offers a $\sim$4.2$\times$ efficiency improvement in 16K pre-training time-to-loss. Across the full development lifecycle, we distill eight actionable empirical findings—spanning architectural scaling, SFT quality pruning, long-context generalization, and agentic co-training dynamics. To facilitate community research, we open-source model weights from the pre-training, mid-training, and post-training stages, intermediate checkpoints, training code, per-stage data and data recipes, and W\&B logs.

}

\code{https://github.com/zgcagi/ZGCM-1}
\newcommand\model[1]{\damodata[Model]{#1}}
\model{https://huggingface.co/zgcagi/ZGCM-1-7B}
\newcommand\data[1]{\damodata[Data]{#1}}
\data{https://huggingface.co/datasets/zgcagi/ZGCM-1-Data}

\begin{document}
\thispagestyle{firstheader}
\maketitle
\begingroup\renewcommand{\thefootnote}{}\footnotetext{Corresponding author: Jiyan He (\href{mailto:hejiyan@zgci.ac.cn}{hejiyan@zgci.ac.cn}).}\addtocounter{footnote}{-1}\endgroup
\pagestyle{plain}

\begin{figure}[H]
  \centering
  \includegraphics[width=0.78\textwidth]{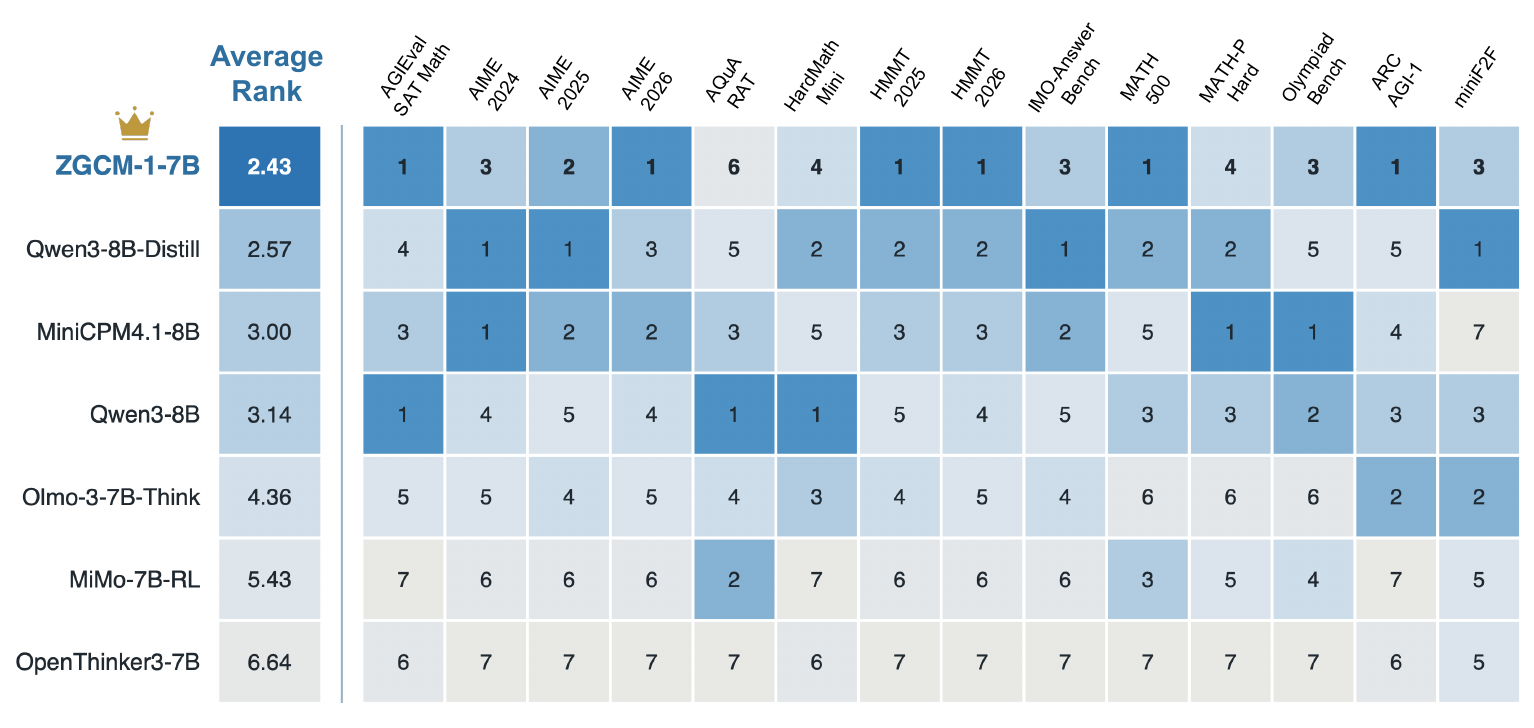}
  \caption{Per-benchmark ranks for \zgcmeval{} and models at comparable scale across
  14 reasoning benchmarks. Full results appear in
  \cref{sec:posttraining-evaluation}.}
  \label{fig:homepage-evaluation-overview}
\end{figure}

\newpage

\begin{figure}[ht]
  \centering
  \includegraphics[width=\textwidth]{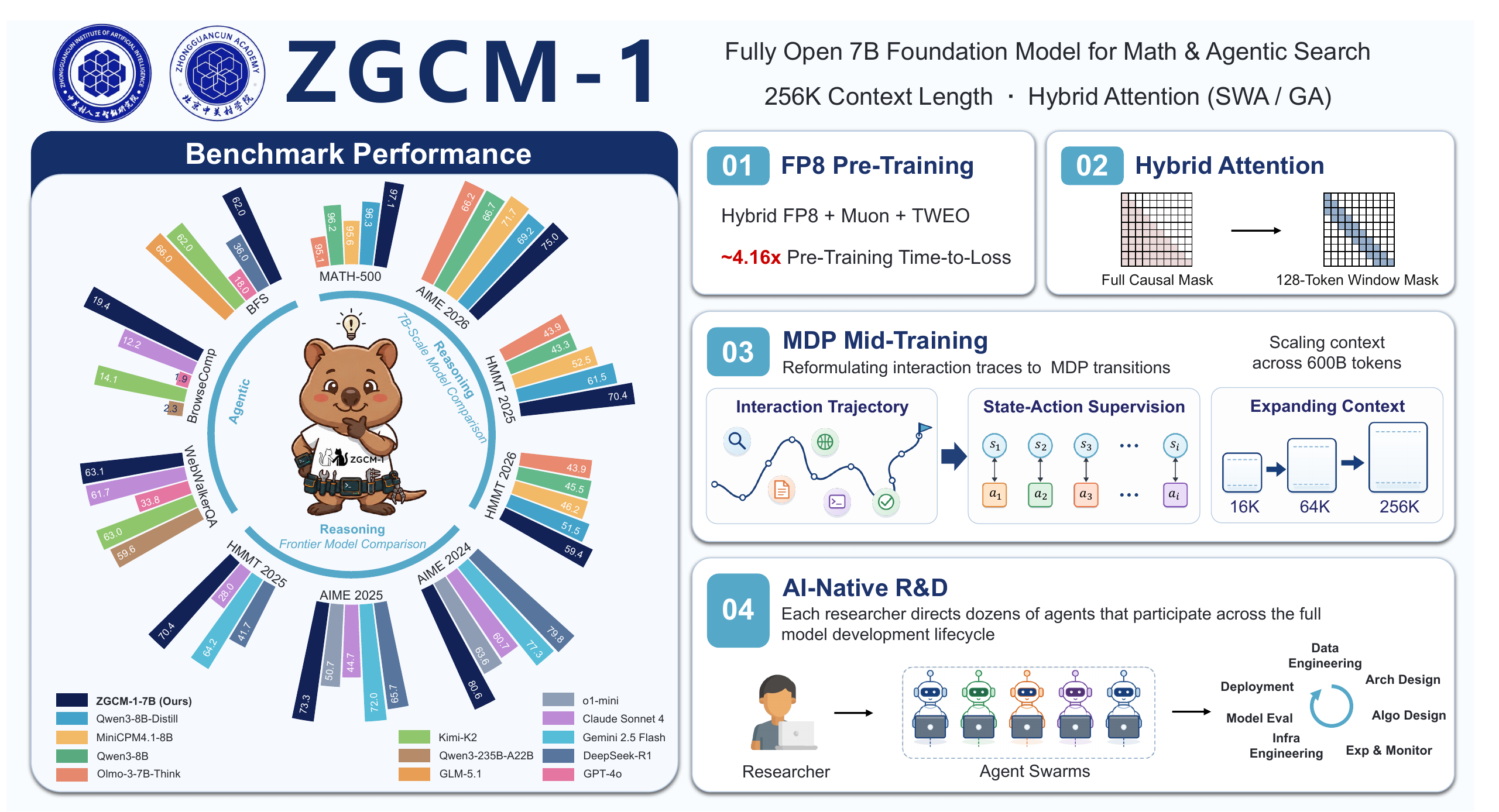}
  \caption{Overview of \zgcm{}. Left: benchmark performance across reasoning
  and agentic tasks compared with 7B-scale and frontier models. Right: four
  key technical highlights: (1) FP8 pre-training with Muon and TWEO achieving
  ${\sim}4.2\times$ time-to-loss speedup, (2) hybrid sliding-window/global
  attention, (3) MDP mid-training that reformulates interaction traces into
  state-action supervision with context scaling to 256K, and (4) AI-native
  R\&D where each researcher directs agent swarms across the full development
  lifecycle. BFS denotes Binary Function Search. Qwen3-8B-Distill denotes DeepSeek-R1-0528-Qwen3-8B. Full results and
  evaluation protocols appear in \cref{sec:posttraining-evaluation}.}
  \label{fig:introduction-evaluation-overview}
\end{figure}

\section{Introduction}

Foundation models are advancing at an extraordinary pace, pushing the frontier of long-horizon reasoning \citep{deepseekv4, kimik3} and tool-augmented agency in real-world environments \citep{glm5, qwen38, openai2026gpt56, anthropic2026cryptoweaknesses}. Despite these breakthroughs, foundational research remains encumbered by two practical bottlenecks:

\begin{itemize}[leftmargin=*,noitemsep]
    \item \textbf{The Scale Barrier}: Frontier reasoning and deep agentic search are widely seen as the exclusive preserve of hundred-billion-parameter systems, locking compute-constrained researchers out of training and exploring frontier-grade intelligence.
    \item \textbf{The Opacity Barrier}: Most competitive models are released strictly as \textit{open-weight} rather than \textit{fully open-source}. Upstream filtering recipes, mid-training curricula, long-context schedules, and multi-turn agent traces remain proprietary black boxes, preventing systematic study of training dynamics and capacity limits.
\end{itemize}

\paragraph{Our Motivation \& Core Thesis.} To address these barriers, we present \textbf{ZGCM-1}, a 7.39B dense foundation model trained from scratch under a transparent, open-science paradigm. We challenge the notion that advanced intelligence strictly requires massive parameter scales, centering our design on a straightforward thesis:
\begin{quote}
\centering
\textit{Compact models are inherently bounded by static parametric capacity, but they can transcend this limitation through a dual engine of deliberate internal thinking and active external seeking.}
\end{quote}
Rather than relying on passive memorization of the open web, ZGCM-1 bridges knowledge gaps by coupling long-horizon chain-of-thought reasoning with autonomous tool use---actively gathering web evidence, interacting with system terminals, and analyzing stripped binary programs.

\paragraph{High-Efficiency Open Recipes.} To make training and inference tractable on academic compute budgets, we build an efficient full-stack pipeline across 256K contexts:
\begin{enumerate}[leftmargin=*,noitemsep]
    \item \textbf{Hybrid Attention Architecture}: We interleave gated sliding-window attention (SWA) with global attention at a 5:1 ratio, reducing per-token KV-cache footprint by \textbf{6.4$\times$} and delivering a \textbf{3.94$\times$ throughput speedup} at 256K context over standard full attention.
    \item \textbf{System-Algorithm Co-Design}: We combine the Muon optimizer \citep{jordan2024muon, liu2025muon}, hybrid FP8 precision, and TWEO outlier regularization \citep{liang2025tweo}, achieving a \textbf{$\sim$4.2$\times$ pre-training time-to-loss speedup} over an AdamW/BF16 baseline.
    \item \textbf{Curriculum Mid-Training with MDP Supervision}: We progressively scale context across 600B tokens (16K $\rightarrow$ 64K $\rightarrow$ 256K) while reformulating interaction traces into Markov Decision Process (MDP) state-action transitions to provide dense, step-level supervision.
    \item \textbf{Execution-Grounded Alignment \& Mixed SFT}: We apply mixed think/no-think fine-tuning on execution-verified trajectories balancing deep reasoning with direct-response efficiency.
\end{enumerate}

\paragraph{Empirical Feasibility.} Evaluations across 20 standard benchmarks confirm our thesis (\cref{fig:introduction-evaluation-overview}, \cref{sec:posttraining-evaluation}):
\begin{itemize}[leftmargin=*,noitemsep]
    \item \textbf{Reasoning \& Mathematics}: ZGCM-1-7B ranks first on average across 14 reasoning benchmarks at the 7B--8B scale while scoring \textbf{75.0\% on AIME 2026}, \textbf{97.1\% on MATH-500}, and \textbf{70.4\% on HMMT 2025}.
    \item \textbf{Agentic Search \& System Agency}: Deliberate thinking coupled with tool interaction allows ZGCM-1 to contend with frontier models orders of magnitude larger (e.g., Claude 4 Sonnet, Kimi-K2, and GLM-5.1), achieving \textbf{63.1\% on WebWalkerQA}, \textbf{19.4\% on BrowseComp}, and \textbf{62.0\% on Binary Function Search}.
\end{itemize}

\paragraph{AI-Native R\&D.} We integrate researcher-directed AI agents throughout the model development lifecycle, from data processing and experimentation to evaluation and deployment. A shared agent harness combines human context artifacts with validated scripts, workflows, and debugging experience, enabling agents to execute tasks, inspect feedback, and iterate. Our Atomic Capability Evaluation suite provides rapid diagnostic feedback to guide development. Assessments from nine core contributors characterize both the benefits and limitations of this workflow: experimentation, monitoring, and deployment receive higher autonomy ratings, while architecture and learning algorithm design remain more dependent on human judgment and direction.

\paragraph{Empirical Findings and Open Science.} Across the development lifecycle, we distill \textbf{eight empirical findings} spanning architectural efficiency, training dynamics, post-training data selection, long-context generalization, agentic co-training, and the benefits and limitations of AI-native R\&D. To support community research, we release model weights from the pre-training, mid-training, and post-training stages, intermediate checkpoints, training code and configurations, data recipes, W\&B logs, and evaluation harnesses at \url{https://github.com/zgcagi/ZGCM-1}.

\section{Architecture}
\label{sec:architecture}

\subsection{Model Overview}

\zgcm{} follows a decoder-only Transformer
architecture with Grouped-Query
Attention~(GQA;~\citealt{ainslie2023gqa}),
RMSNorm~\citep{zhang2019root}, SwiGLU
activation~\citep{shazeer2020glu}, and Rotary Position
Embedding~(RoPE;~\citealt{su2024roformer}). The model has approximately 7.39 billion
parameters, 32 Transformer layers, a hidden dimension of 4,096, a SwiGLU
intermediate dimension of 11,008, 32 query heads and 8 key-value heads
(head dimension 128), and a maximum context length of 256K tokens.

The architecture uses a hybrid causal-attention backbone that interleaves
gated~\citep{qiu2025gatedattention} sliding-window attention
(SWA;~\citealt{child2019generating,beltagy2020longformer})
with global attention at the same 5:1 local-to-global ratio previously deployed
at production scale by Gemma 3~\citep{gemma3}. Of the 32 layers, 27 use gated SWA with a 128-token window and
five use global causal attention. The global layers are placed at layers 6,
12, 18, 24, and 30, giving five repeated blocks of five local layers followed
by one global layer, plus two trailing local layers. The model further applies
QK normalization~\citep{dehghani2023scalingvit}, implemented with RMSNorm, and Partial RoPE with a rotary fraction of 0.33.

\begin{figure}[t]
  \centering
  \includegraphics[width=0.85\textwidth]{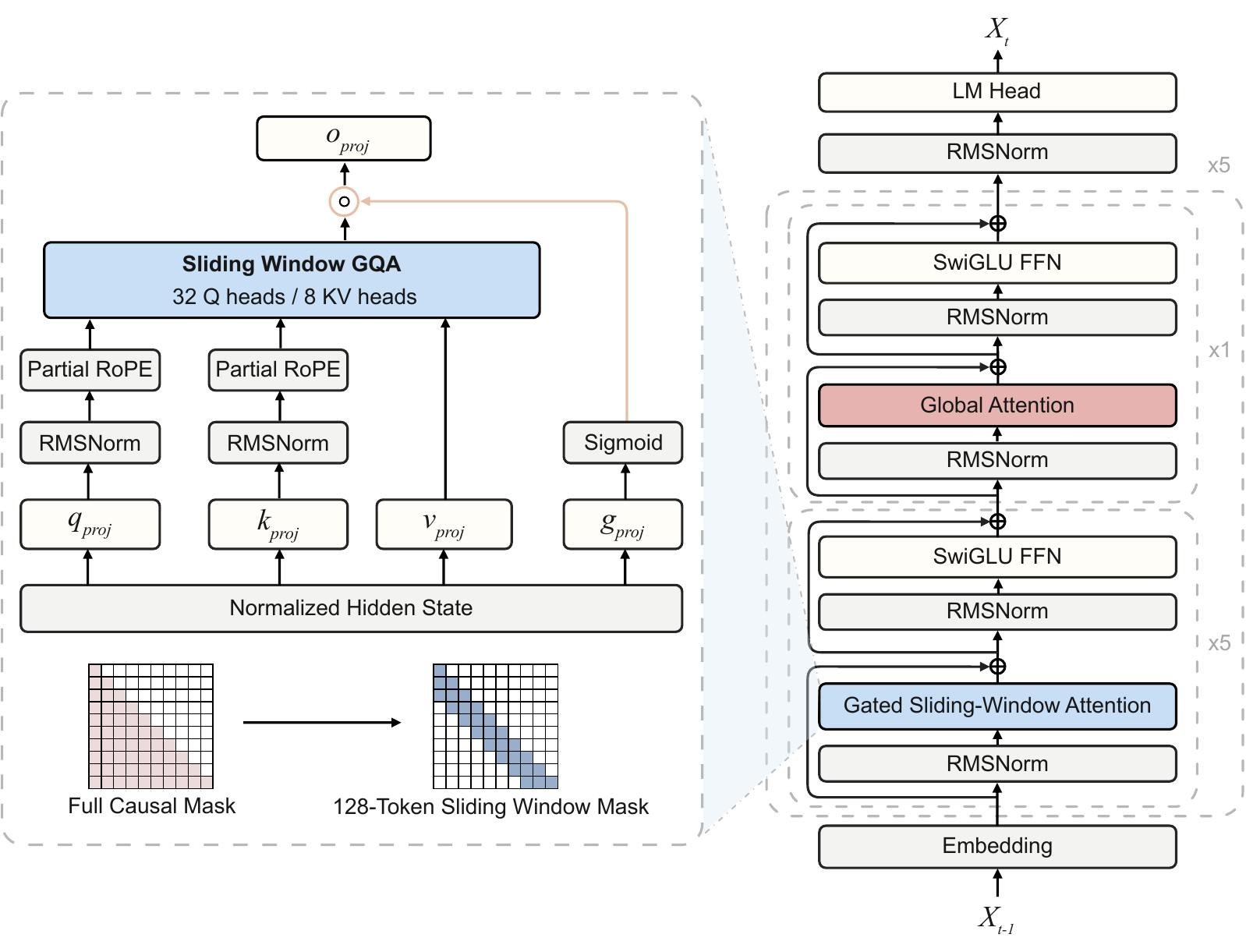}
  \caption{Hybrid attention architecture of \zgcm{}. The right panel shows
  the backbone formed by gated sliding-window and global attention layers;
  the left panel details the gated sliding-window attention module and its
  corresponding attention masks. The last global layer is layer~29
  (0-indexed), followed by two trailing SWA layers.}
  \label{fig:zgcm1-architecture}
\end{figure}

As illustrated in Figure~\ref{fig:zgcm1-architecture}, for a normalized
hidden state $h$, the gated SWA module forms query, key, value, and gate
projections in parallel. RMS normalization and Partial RoPE are applied to the
query and key branches. If $A_{\mathrm{SWA}}(q,k,v)$ denotes the 128-token
sliding-window GQA output, the gated attention output is

\[
  \operatorname{GatedSWA}(h)
  = o_{\mathrm{proj}}\!\left(
      A_{\mathrm{SWA}}(q,k,v) \odot \sigma(g_{\mathrm{proj}}(h))
    \right).
\]

The learned sigmoid gate modulates the local-attention output element-wise
before the output projection. Global-attention layers omit this gating and
attend over the full context, allowing information flow beyond the local
window.

\begin{figure}[t]
  \centering
  \includegraphics[width=\linewidth]{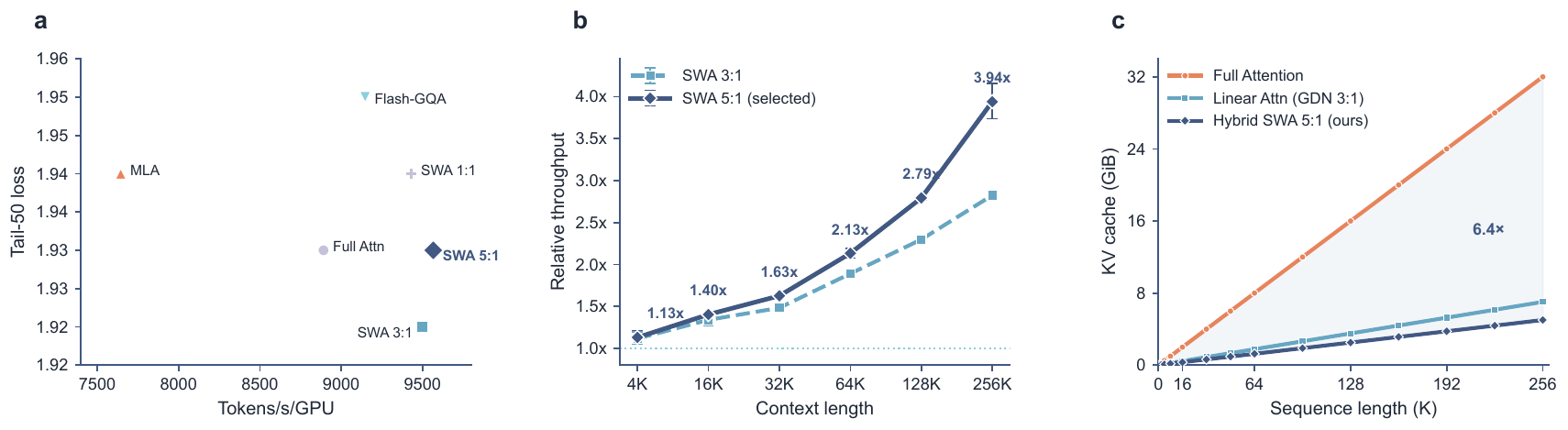}
  \caption{Architecture experiments. (a) Quality-throughput trade-off
  among non-parameter-matched 7B attention configurations trained for 10B
  tokens (lower loss and higher throughput are better). (b) Throughput
  speedup of SWA schedules over full attention as context length grows from 4K
  to 256K. (c) KV cache memory footprint of three iso-parameter architectures
  (batch\,=\,1, bf16): full attention (32 global layers), linear attention
  (GDN 3:1, 7 global + 21 linear, L=28), and our hybrid SWA 5:1 (5 global +
  27 SWA).}
  \label{fig:architecture-experiments}
\end{figure}

\subsection{Hybrid-Attention Experiments}
\label{sec:attention-experiments}

To select the production attention schedule, we compare full attention,
GQA with FlashAttention-2
(Flash-GQA;~\citealt{ainslie2023gqa,dao2023flashattention2}),
MLA~\citep{deepseekv2}, and three SWA schedules with local-to-global ratios of 1:1,
3:1, and 5:1. All configurations are trained for 10B tokens at sequence length
4,096 on eight H100 GPUs with a common training budget.

We measure optimization quality by mean loss over the final 50 training steps
and training efficiency by tokens per second per GPU.
As shown in \cref{fig:architecture-experiments}a, SWA 5:1
achieves the highest throughput (9,566 tokens/s/GPU) while matching the tail
loss of the full-attention baseline (1.93). SWA 3:1 reaches the lowest tail
loss (1.92) at slightly lower throughput. MLA incurs a substantial throughput
penalty (7,645 tokens/s/GPU) without a corresponding loss improvement. We adopt SWA 5:1 for the production model: it offers the highest observed
throughput at a competitive tail loss. The production architecture
additionally incorporates a 128-token window, QK RMS normalization, GQA, and
local output gating.

We further compare full attention, SWA 3:1, and SWA 5:1 as context length
increases from 4K to 256K on the same 7B backbone with a constant token
budget per step. \Cref{fig:architecture-experiments}b shows
that the throughput advantage of SWA 5:1 over full attention widens with
context length, growing from \(1.13\times\) at 4K to \(3.94\times\) at 256K.
SWA 3:1 provides an intermediate speedup profile. This widening gap
makes gated SWA well suited for long-context training and
inference, where full attention becomes the dominant computational bottleneck.

\subsection{KV Cache Analysis}
\label{sec:kv-cache-analysis}

The hybrid architecture yields substantial memory savings at inference time.
Because the 27 SWA layers retain only a fixed 128-token window in their KV
cache while only the 5 global layers store the full sequence, the per-token KV
footprint drops from 128\,KiB (full attention) to 20\,KiB.
\Cref{fig:architecture-experiments}c compares three iso-parameter architectures:
full attention (32 global layers, 7.39B), a linear-attention baseline (GDN 3:1
with 7 global + 21 linear layers, 7.47B), and our hybrid SWA 5:1. At 256K
context, full attention requires 32.0\,GiB of KV cache, GDN 3:1 requires
7.0\,GiB, and our hybrid requires only 5.0\,GiB (a 6.4\(\times\) reduction
over full attention).
Since autoregressive decoding is memory-bandwidth-bound, this smaller KV
footprint directly translates to higher decode throughput, making the
architecture particularly efficient for long-reasoning and agentic-search
workloads that routinely operate at long context lengths.

\section{Pre-Training}
\label{sec:pretraining}

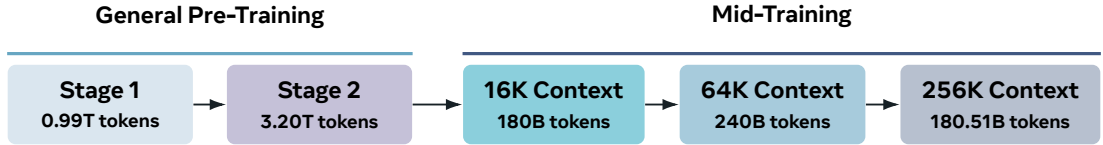
\begin{figure}[t]
  \centering
  \definecolor{ptStageOne}{HTML}{DAE6EF}
  \definecolor{ptStageTwo}{HTML}{C5C1D8}
  \definecolor{ptContextShort}{HTML}{8BCFDC}
  \definecolor{ptContextMedium}{HTML}{67A6C2}
  \definecolor{ptContextLong}{HTML}{415882}
  \definecolor{ptInk}{HTML}{172126}
  \begin{tikzpicture}[
      stage/.style={
        rounded corners=3pt,
        minimum height=1.05cm,
        inner xsep=8pt,
        inner ysep=5pt,
        align=center,
        draw=none,
        text=black,
        font=\sffamily
      },
      phase/.style={text=black,font=\sffamily\bfseries},
      flow/.style={-{Latex[length=2.2mm,width=1.5mm]},draw=ptInk,
        line width=0.7pt}
    ]
    \node[stage,fill=ptStageOne,minimum width=2.45cm] (stage1) at (1.25,0)
      {Stage 1\\[-1pt]{\footnotesize 0.99T tokens}};
    \node[stage,fill=ptStageTwo,minimum width=2.45cm] (stage2) at (4.15,0)
      {Stage 2\\[-1pt]{\footnotesize 3.20T tokens}};

    \node[stage,fill=ptContextShort,minimum width=2.20cm] (ctx16) at (7.25,0)
      {16K Context\\[-1pt]{\footnotesize 180B tokens}};
    \node[stage,fill=ptContextMedium!58,minimum width=2.20cm,
      right=0.45cm of ctx16] (ctx64)
      {64K Context\\[-1pt]{\footnotesize 240B tokens}};
    \node[stage,fill=ptContextLong!38,minimum width=2.20cm,
      right=0.45cm of ctx64] (ctx256)
      {256K Context\\[-1pt]{\footnotesize 180.51B tokens}};

    \node[phase] at (2.70,1.15) {General Pre-Training};
    \draw[draw=ptContextMedium,line width=1.2pt]
      ([yshift=4pt]stage1.north west) -- ([yshift=4pt]stage2.north east);
    \node[phase] at
      ($(ctx16.north west)!0.5!(ctx256.north east)+(0,0.625cm)$)
      {Mid-Training};
    \draw[draw=ptContextLong,line width=1.2pt]
      ([yshift=4pt]ctx16.north west) -- ([yshift=4pt]ctx256.north east);

    \draw[flow] (stage1.east) -- (stage2.west);
    \draw[flow] (stage2.east) -- (ctx16.west);
    \draw[flow] (ctx16.east) -- (ctx64.west);
    \draw[flow] (ctx64.east) -- (ctx256.west);
  \end{tikzpicture}
  \caption{General Pre-Training is organized into two data stages, followed by three
  Mid-Training stages with progressively longer context lengths.}
  \label{fig:pretraining-stage-structure}
\end{figure}

Pre-Training consists of two consecutive phases.
As illustrated in \cref{fig:pretraining-stage-structure}, General Pre-Training
builds broad language, knowledge, mathematics, and code capabilities across
two data stages. Mid-Training then retains the full-sequence causal
language-modeling objective while introducing denser reasoning, instruction,
and agentic data and progressively extending the context from 16K to
64K and 256K. Detailed corpus construction, curriculum experiments, and
training configurations are reported in \cref{app:pretraining-details}.

\subsection{General Pre-Training}

\subsubsection{Data Mixture}

We first search for data mixtures with a 0.3B-parameter proxy model trained
on approximately 30B tokens. Training loss and evaluation signals for
knowledge, code, and mathematics guide iterative adjustments to candidate
mixtures, reducing the cost of mixture exploration at the target model scale,
following the broader practice of proxy-model mixture optimization
\citep{regmix,olmo3}.
Based on the mixture-search results, we balance capabilities across domains
against loss-convergence efficiency to define the final training data recipe.

\noindent\textbf{Web data.}
Curated English web corpora~\citep{wang2025ultrafineweb} form the main source,
complemented by a controlled allocation of Chinese web data. We retain upstream
document-level quality scores and quality strata produced by validation-driven
filtering, and normalize accepted text and language metadata into a common
document representation.

\noindent\textbf{Academic and OCR data.}
We combine educationally filtered PDF views~\citep{kydlicek2025finepdfs} with
OCR-derived scientific content released by
Ai2~\citep{allenai2025olmocrpes2o}, produced with the olmOCR
pipeline~\citep{poznanski2025olmocr}. For scientific papers
collected directly from arXiv, our in-house pipeline extracts and cleans the
text while retaining the technical content required for training.

\noindent\textbf{Code data.}
The code mixture covers both code-rich web pages and open-source repository
files~\citep{nvidia2025nemotroncccode,nvidia2025nemotroncode}. We convert
structured releases into text views,
normalize file-level metadata, and clean repository artifacts and non-source
content while preserving programs, technical documentation, and explanatory
code text.

\noindent\textbf{Mathematics data.}
We prioritize higher-quality tiers from mathematical
corpora~\citep{zhou2026ultradatamath}, together with classifier-filtered web
mathematics and textbook collections following the math data recipe of
SmolLM2~\citep{allal2025smollm2}. Processing combines
heuristic
cleaning, quality-model selection, and formula-preserving normalization so
that LaTeX expressions remain embedded in their surrounding reasoning context,
as in Proof-Pile-2~\citep{azerbayev2023llemma}.

\noindent\textbf{LaTeX papers.}
We combine the arXiv LaTeX-source slice of
RedPajama-1T~\citep{together2023redpajama} with
filtered recent TeX sources. The accepted views
recover the main textual stream, filter malformed or content-poor documents,
and preserve equations and scientific document structure.

\noindent\textbf{Specialized reasoning data.}
Reasoning-oriented material is maintained as a separate source family rather
than folded into the general web or mathematics pools. We draw on
Nemotron-Pretraining-Specialized-v1~\citep{nvidia2025nemotronspecialized},
validate source schemas and versions, select the designated high-quality
reasoning views, and admit them to Stage 2 as an independently controlled
mixture component.

Each source family undergoes source-specific language and quality filtering,
together with text extraction or repository cleaning where required. Accepted
documents are normalized into a common record schema, materialized as
versioned shards, and registered in source manifests. We then tokenize and
index the shards with the GLM-5.1 tokenizer. Cross-stage deduplication excludes
all Stage-1 content from Stage-2 selection before each source is sampled
according to the final data recipe.

The final General Pre-Training corpus contains approximately 0.99T tokens in
Stage 1 and 3.20T tokens in Stage 2. As shown in the upper donuts of
\cref{fig:pretraining-data-mixture}, web data remains the largest source in
both stages. Stage 2 increases the relative contribution of code and
mathematics, introduces a specialized reasoning component, and retains broad
coverage of academic, OCR, and LaTeX content.

\subsubsection{Curriculum Pretraining}

Recent work shows that ordering training examples by difficulty rather than
sampling randomly can improve pre-training efficiency within a fixed token
budget~\citep{zhang2026curriculum}. We use the 0.99T-token Stage 1 as a
curriculum pretraining phase. General-language documents are presented from
lower to higher lexical complexity, while code and mathematics are interleaved
separately. The controlled comparison and qualitative examples are reported in
\cref{app:curriculum-learning,app:curriculum-case-studies}. We find that lexical-complexity ordering is cheap to compute at corpus scale for
general-language data. However, it does not reliably reflect the intrinsic
difficulty of code or mathematics. We therefore apply the ordering only to
non-code, non-mathematics data, remove extreme high-complexity outliers that are
usually corrupted or garbled text, and interleave code and mathematics
independently. In the 7B probe, this schedule lowers coding BPB from 1.99 to
0.81 and mathematics BPB from 0.97 to 0.94, while general-benchmark BPB rises by
0.03 to 0.09. The mathematics comparison uses matched evaluation sampling, while
the coding gap indicates direction rather than a matched effect size. The
ordering applies only to Stage 1, and later stages train on the full mixture
without complexity ordering.

\Needspace{10\baselineskip}
\begin{findingbox}
\textbf{\textit{Finding 1 (Curriculum Pre-training):}}
Lexical complexity is a cheap corpus-scale ordering signal for general-language text, but a poor difficulty proxy for code and mathematics, where surface statistics track boilerplate rather than algorithmic or reasoning depth. The resulting schedule, which sorts general-language documents while interleaving code and mathematics independently, improves BPB on both technical domains and trades $0.03$--$0.09$ BPB of general-domain fit for gains concentrated where the model is targeted.
\end{findingbox}

\subsubsection{Hyperparameters}

\noindent\textbf{Optimization.}
We use NVIDIA Megatron Core~\citep{nvidia2026megatroncore} for distributed
training on H100 GPUs. Matrix parameters are optimized with
Muon~\citep{jordan2024muon,liu2025muon} (momentum 0.9, spectral scaling, five Newton--Schulz
steps, constant learning rate $2\times10^{-4}$, weight decay 0.1, gradient
clipping 1.0). Scalar parameters use Adam.

\noindent\textbf{Numerical precision.}
Matrix multiplications use Transformer Engine hybrid FP8 (E4M3 forward,
E5M2 backward)~\citep{micikevicius2022fp8} with delayed scaling; scaling
factors are updated from the maximum absolute value over a 1,024-step
history. All other operations retain
BF16 or FP32 precision. TWEO~\citep{liang2025tweo} is applied as an activation
regularizer to suppress extreme intermediate values, complementing the
delayed-scaling rule. The 16K production run sustains approximately 585 model TFLOP/s/GPU,
i.e.\ approximately 60\% BF16-equivalent MFU against the 989 TFLOP/s
H100 BF16 dense peak. Full parallelism configurations are reported in
\cref{app:training-configurations}.

\Needspace{8\baselineskip}
\begin{findingbox}
\textbf{\textit{Finding 2 (TWEO enables FP8 Pretraining):}} Suppressing intermediate activation outliers is crucial for FP8 training stability; combining delayed scaling with TWEO~\citep{liang2025tweo} regularization prevents numerical divergence while sustaining ~60\% BF16-equivalent MFU at scale.
\end{findingbox}

\noindent\textbf{Training efficiency.}
We estimate 16K pre-training time-to-loss relative to a comparable
OLMo~3-style 7B BF16/AdamW baseline~\citep{olmo3}. Four factors contribute:
SWA 5:1 yields a $1.4\times$ throughput gain over full attention at
16K (\cref{sec:attention-experiments}, \cref{fig:architecture-experiments}); the FP8 precision-and-systems configuration contributes
approximately $1.5\times$; Muon provides approximately $1.8\times$
step-to-loss efficiency over AdamW; and the Pre-LN contributes an
estimated $1.1\times$ data efficiency from exploratory 7B evidence.
Multiplying these gives
\[
1.4\times 1.5\times 1.8\times 1.1 \approx 4.2.
\]
This roughly indicates a fourfold improvement in 16K pre-training
time-to-loss.

\Needspace{8\baselineskip}
\begin{findingbox}
\textbf{\textit{Finding 3 (Pretraining Efficiency Gain):}} The co-design of hybrid SWA ($1.4\times$), FP8 mixed-precision systems ($1.5\times$), the Muon optimizer ($1.8\times$), and Pre-LN ($1.1\times$) yields a cumulative $\sim\!4.2\times$ speedup in 16K pre-training time-to-loss over a standard BF16/AdamW baseline.
\end{findingbox}

\subsection{Mid-Training}

\begin{figure}[t]
  \centering
  \includegraphics[width=\textwidth]{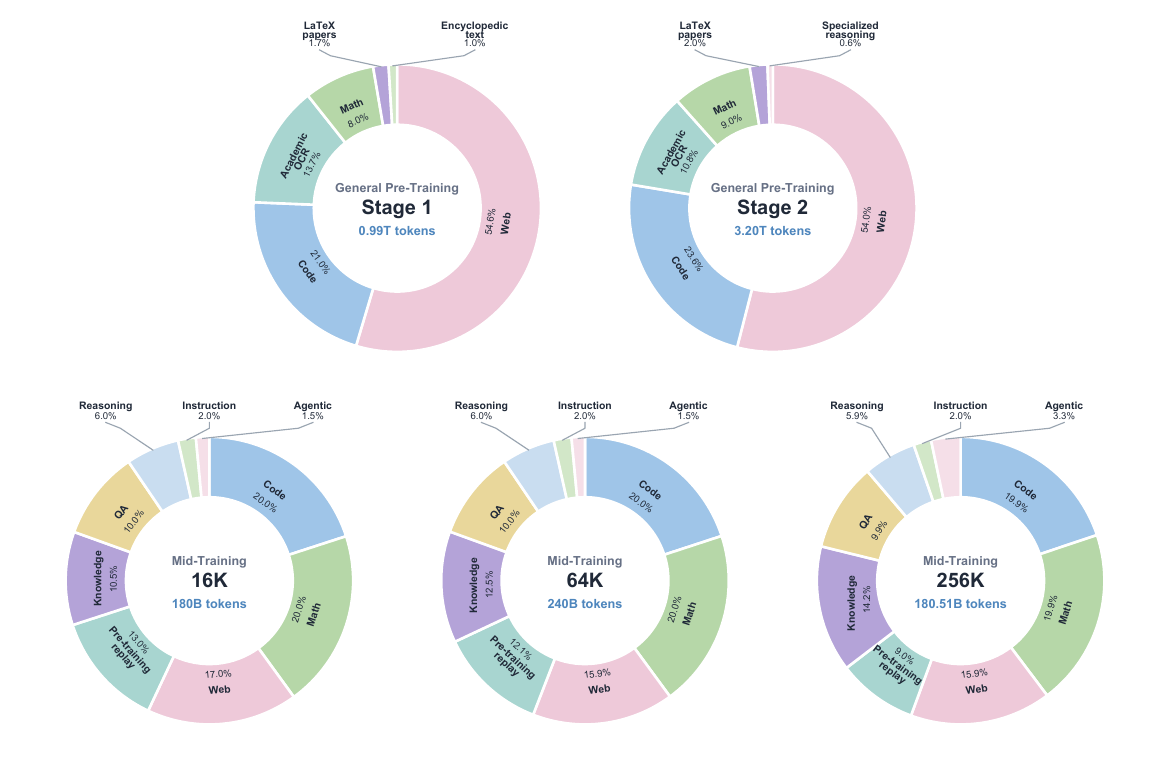}
  \caption{Data mixture composition shown as donut charts for the two General
  Pre-Training stages and the three Mid-Training context stages. Wedge areas
  represent each category's share of the corresponding stage's total tokens.}
  \label{fig:pretraining-data-mixture}
\end{figure}

\subsubsection{Data Mixture}

Mid-Training is capability-oriented continued pre-training. We apply pool-wide deduplication to improve token efficiency and
control repeated content \citep{lee2022dedup}. From the resulting 2.86T-token
candidate pool, we construct a 600.51B-token sampled schedule organized into
16K, 64K, and 256K context stages. The maximum sequence length increases
progressively across these stages, following the multi-stage context expansion
of Qwen2.5-1M~\citep{qwen2025qwen25_1m}; other open technical reports extend
the context in a single final pre-training
stage~\citep{mimo,qwen2025qwen3}. Each later stage remains cumulative
rather than replacing shorter sequences: the 64K stage contains 180.89B tokens
at up to 16K and 59.11B tokens in the 16K--64K range, while the 256K stage combines
127.81B tokens at up to 16K, 21.72B tokens in the 16K--64K range, and 30.98B
tokens above 64K.

The 16K stage trains at the standard context length with a balanced mixture of
code, mathematics, knowledge, and reasoning. The 64K stage introduces longer
documents and reasoning sequences while retaining a substantial share of
shorter-context data. The 256K stage adds ultra-long documents, cross-document
information, and long-horizon agentic trajectories, and continues to replay
data from the 16K and 64K length ranges to preserve coverage of
conventional-context tasks.

The lower donuts of \cref{fig:pretraining-data-mixture} summarize the
top-level capability mixture. Code and mathematics each remain close to 20\%
throughout the curriculum, preserving core programming and reasoning
capabilities. As the context length grows, knowledge data increases from
10.50\% to 14.23\%, and agentic data increases from 1.50\% to 3.30\%, raising
the density of long-document, tool-interaction, and multi-step task examples.
Pre-training replay decreases from 13.00\% to 9.00\% but remains present to
maintain broad coverage as higher-density data is introduced. Web, QA,
reasoning, and instruction data remain comparatively stable across the three
stages. Exact length accounting and construction details are given in
\cref{app:midtraining-data}.

\begin{figure}[htbp]
  \centering
  \includegraphics[width=\textwidth]{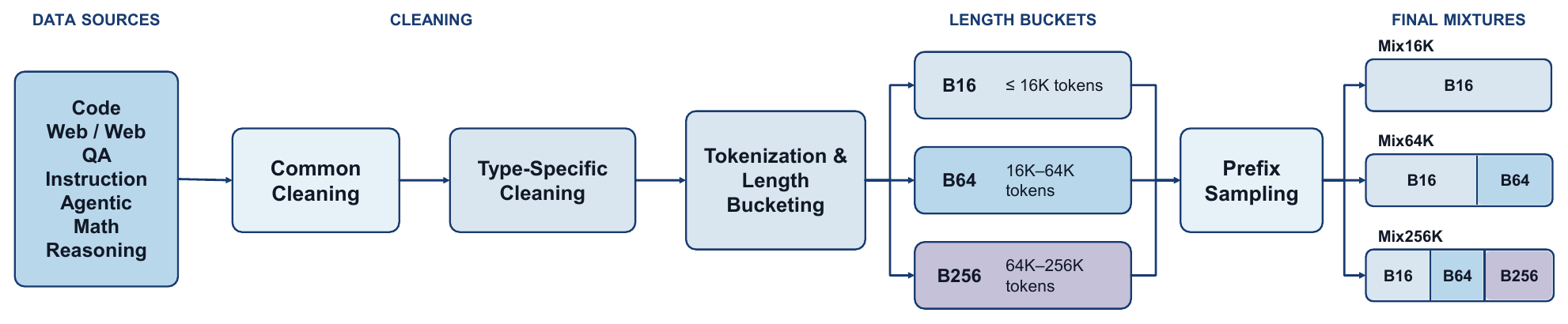}
  \caption{Mid-Training data processing pipeline.}
  \label{fig:midtraining-data-processing-pipeline}
\end{figure}
\FloatBarrier

\subsubsection{Reasoning Data}

Reasoning sources are normalized to a common schema and routed into accepted,
rewrite, pending, holdout, or rejected views according to self-containment,
answer evidence, and reasoning value. Rule-cleaned examples form the main
pool.

Valuable questions whose original traces are unsuitable for training
are extracted as standalone problems, reconstructed or improved by teacher
models, and admitted only after checks for parsing validity, completeness,
template and source leakage risks, duplication, and trainability. This separates
question value from the quality of an upstream reasoning trace. Details are reported in
\cref{app:reasoning-instruction-data}.

\subsubsection{Instruction Data}

We collect a large-scale instruction corpus that includes Tulu- and
FLAN-derived data \citep{lambert2024tulu3,wei2021flan}. These examples are
converted into pre-training-compatible raw context. The pipeline preserves legitimate
short-label tasks, routes translation-risk subsets through model review, and
rewrites selected single-turn examples as multi-turn discussions. These
transformations increase coverage of conversational state and sequential
reasoning while retaining the full-sequence training objective. Further
construction details are provided in \cref{app:reasoning-instruction-data}.

\subsubsection{Agentic Data}

\begin{figure}[htbp]
  \centering
  \includegraphics[width=\textwidth]{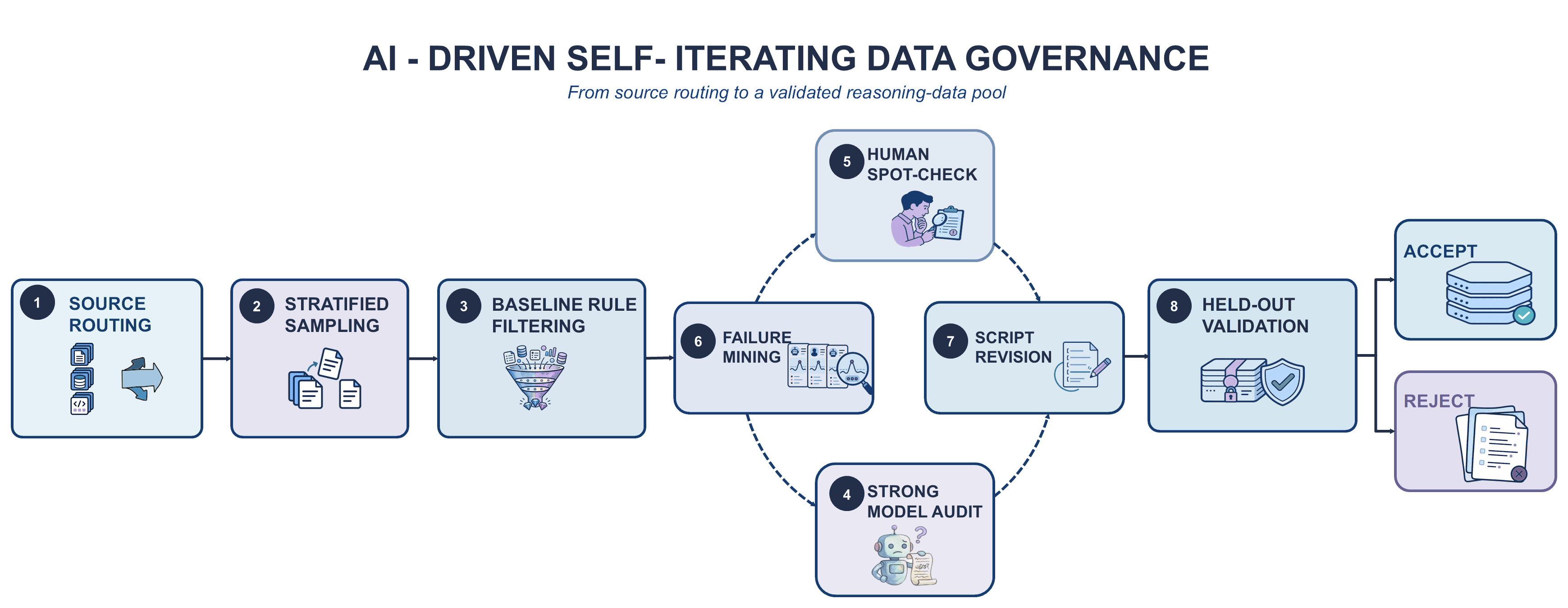}
  \caption{AI-driven self-iterating data governance pipeline. Each source
  family is routed and sampled independently, then processed through
  source-specific rule filtering, strong-model audit, failure mining, human
  spot checks, iterative script revision, and held-out validation before
  accepted examples enter the validated reasoning-data pool.}
  \label{fig:ai-self-iterating-data-governance}
\end{figure}

Agentic data for Mid-Training covers both general interaction traces and
trajectories collected from software-engineering tasks. We retain high-quality
complete interactions and, following agentic continual
pre-training~\citep{agentfounder}, reformulate general traces as Markov
decision process-style state-conditioned next-action prediction examples, combining
full-trajectory context with denser supervision for local decision making. For
software-engineering trajectories, we apply execution-aware filtering and
preserve the interaction context needed to learn planning, tool use, and
iterative refinement. Detailed construction procedures, filtering criteria,
and corpus statistics are provided in
\cref{app:agentic-midtraining-data}. For each source family, we use the
agent-driven self-iterating governance process shown in
\cref{fig:ai-self-iterating-data-governance} to develop and validate a
source-specific processing recipe.

\FloatBarrier

\subsubsection{Hyperparameters}

All Mid-Training stages use full-sequence causal language modeling over raw
context (distinct from the assistant-only supervision used in SFT, as detailed
in \cref{sec:sft-hyperparameters}). We maintain approximately 12.6M tokens per
optimizer step across all three stages by adjusting global batch sizes (768,
192, and 48 for 16K, 64K, and 256K respectively). The learning rate is
$10^{-4}$, and the 256K stage uses a RoPE base of 10M with full activation
recomputation. We evaluate both a direct 256K route (30B tokens) and a staged
route (10B at 64K then 20B at 256K); the staged route achieves a slightly
lower final loss. Full configurations are in \cref{app:training-configurations}.

\begin{figure}[t]
  \centering
  \includegraphics[width=\textwidth]{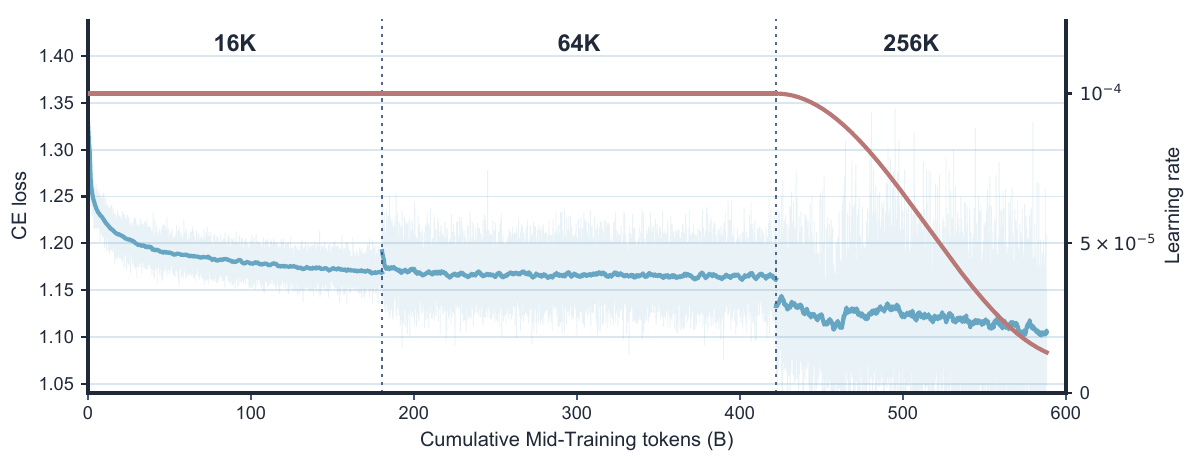}
  \caption{Token-level cross-entropy (CE) loss (blue, left axis) and
  learning-rate schedule (terracotta, right axis) for the staged Mid-Training
  run. The solid blue curve is a 200-point moving average, while the light-blue
  curve shows raw loss values. Both metrics share the cumulative-token axis;
  vertical dotted lines mark transitions between the 16K, 64K, and 256K
  context stages.}
  \label{fig:midtraining-c1-curves}
\end{figure}

The resulting C1 staged run is summarized by its loss trajectory and learning-rate
schedule in \cref{fig:midtraining-c1-curves}.

\subsection{Monitor During Training}
\label{sec:monitor-during-training}

Throughout General Pre-Training and Mid-Training, we continuously monitor
training dynamics, numerical stability, hardware and systems status, training
efficiency, and resource utilization.

To monitor capability development during General Pre-Training, we directly
evaluate saved base checkpoints with a fixed capability suite covering
knowledge, question answering, mathematics, reasoning, coding, and basic
symbolic operations. No supervised fine-tuning or other post-training
adaptation is applied before evaluation. In parallel, we monitor perplexity,
bits per byte (BPB), normalized loss, key-token logits, and optimizer-related
statistics that reflect training dynamics. These signals help diagnose
optimization and systems issues, detect capability regressions, and inform
adjustments to data mixtures, curricula, and stage transitions.

Checkpoint-level evaluations reveal heterogeneous capability development across
domains. Knowledge and question-answering benchmarks generally improve over the
monitored training window, but the rate and timing of improvement differ across
benchmarks. Mathematics and coding capabilities also improve, although their
gains are concentrated in different intervals and accompanied by larger
short-term fluctuations. We therefore treat these curves as descriptive
diagnostics of domain-dependent learning dynamics rather than as evidence of a
universal monotonic trend. Detailed trajectories for all benchmarks are
provided in \cref{app:pretraining-capability-dynamics}.

As an example, continuous throughput monitoring after resuming a training
job revealed a gradual decline across matched windows of
approximately 80 steps, from 585 to 554 model TFLOP/s/GPU. Releasing
overlap buffers reduced but did not eliminate the decline. Adding periodic
CUDA allocator cache clearing and garbage collection every 100 iterations
restored stable throughput at 585 model TFLOP/s/GPU. This fix was
deployed into the production training loop.

A complementary short-SFT probe evaluates successive training milestones after
lightweight supervised adaptation. Unlike the direct base-checkpoint
evaluation above, this probe measures the downstream capabilities that can be
elicited after supervised adaptation. Its results are reported in
\cref{app:midtraining-short-sft-probe}.

\section{Post-Training}
\label{sec:posttraining}

The post-training pipeline consists of supervised fine-tuning (SFT) followed
by mixed reinforcement learning (RL). SFT establishes response modes with
general and agentic data, while RL yields capability gains on single-domain
tasks such as mathematics and code. Additional data and implementation details
are provided in \cref{app:posttraining-details}.

\subsection{Supervised Fine-Tuning}
\label{sec:sft}

\begin{figure*}[t]
  \centering
  \includegraphics[width=\textwidth]{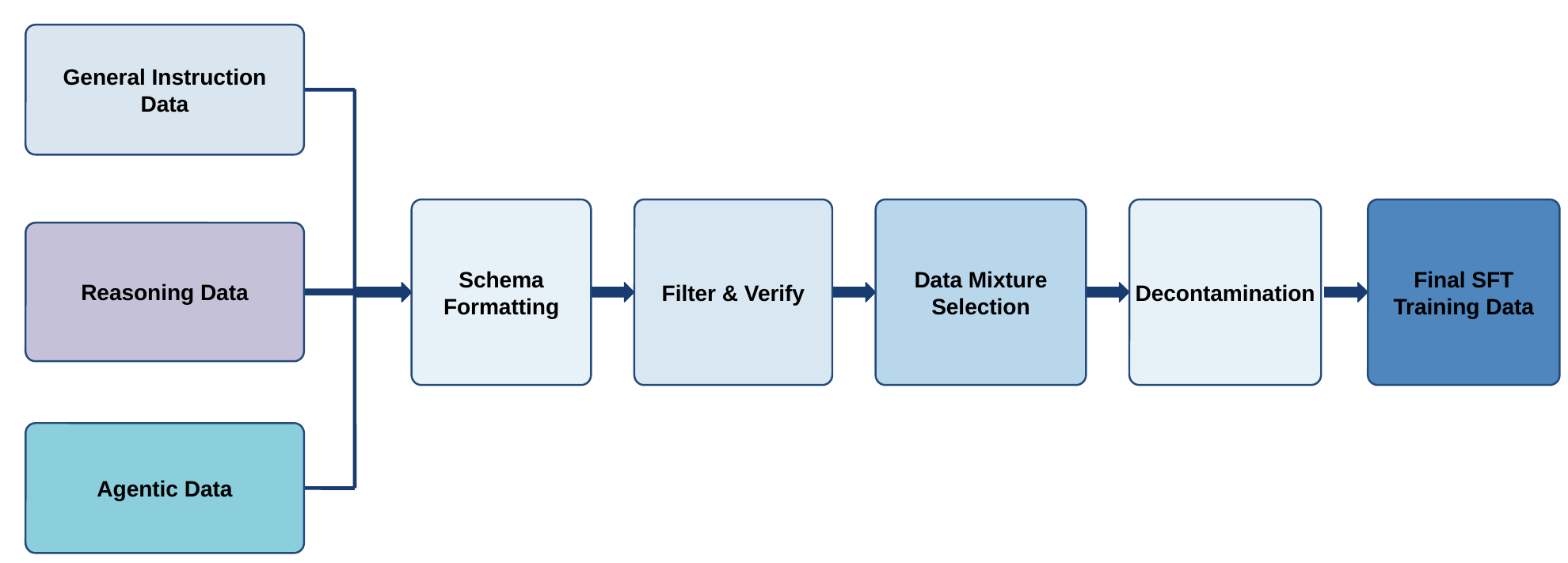}
  \caption{Supervised fine-tuning data curation pipeline. General data
  comprises general instruction and reasoning examples, while agentic data
  forms a separate branch. All sources undergo schema formatting, filtering and
  verification, mixture selection, benchmark decontamination, and preparation
  as packed training sequences with assistant-only loss.}
  \label{fig:sft-data-curation}
\end{figure*}

\Cref{fig:sft-data-curation} summarizes our SFT data curation pipeline. We
organize the corpus into general and agentic data, with general data comprising
general instruction and reasoning examples. Before mixture selection, all
examples are normalized to a common message and tool schema and filtered for
structural validity, response correctness, and tool-turn consistency.
Candidate-run evaluations then determine a mixture that balances capability
coverage, long-reasoning supervision, and joint general-agentic training.
Finally, we remove benchmark contamination and convert the retained examples
into packed sequences with assistant-only loss masks and length-based routing.

\subsubsection{General Data}

Our SFT corpus combines open-source datasets with internally distilled data and
contains 4,921,933 examples organized into general and agentic data. As shown
in \cref{tab:sft-composition}, general data covers instruction following,
knowledge, mathematics, science, code, dialogue, and reasoning, while agentic
data targets interaction with tools and external environments.

\begin{table}[h]
  \centering
  \caption{Composition of our supervised fine-tuning corpus.}
  \label{tab:sft-composition}
  \small
  \begin{tabular}{@{}L{0.18\linewidth}C{0.18\linewidth}L{0.43\linewidth}@{}}
    \toprule
    Component & Row share & Primary role \\
    \midrule
    General data & 96.46\% & Instruction following, knowledge, mathematics,
      science, code, dialogue, and reasoning \\
    Agentic data & 3.54\% & Deep research, software engineering, and terminal
      interaction through structured tools \\
    \midrule
    Total & 100.00\% & All supervised fine-tuning data \\
    \bottomrule
  \end{tabular}
\end{table}

\noindent\textbf{Data Quality Filtering and Tiered Curation.} We establish a multi-stage data curation pipeline combining deterministic heuristic rules with model-based quality scoring. In the initial stage, rule-based checks systematically eliminate structural defects, including missing dialogue turns, invalid role assignments, duplicate generations, prompt template leakages, and malformed reasoning traces.Following rule-based filtering, we deploy a model-based evaluator calibrated against human-annotated pilot benchmarks to grade candidate instances along multiple pedagogical and technical axes: \textit{educational value}, \textit{logical soundness}, \textit{step-by-step reasoning coherence}, \textit{factual consistency}, and \textit{safety compliance}. Rather than applying rigid binary thresholds, fixed multi-criteria scoring rules classify examples into discrete quality tiers. High-scoring instances offering dense supervision and diverse capability coverage are prioritized, borderline cases are routed to auxiliary verification pipelines, and noisy, uninformative, or high-risk entries are strictly discarded. Refusal samples undergo dedicated isolation and filtering to eliminate over-refusals and environment-dependent artifacts while preserving essential safety alignment.Empirical comparisons across candidate training runs indicate that aggressive tiered filtering—pruning approximately 50\% of the raw SFT candidates—outperforms training on the uncurated full corpus in aggregate, though individual benchmarks do not all move in the same direction. A controlled three-way comparison of filtering strength is reported in \cref{app:sft-quality-ablation}.

\begin{findingbox}
\textbf{\textit{Finding 4 (Data Quality Trumps Volume in Posttraining):}} SFT quality matters
more than data volume. In controlled comparisons, pruning roughly half of the
candidate pool improved the six-benchmark mean from 67.78 to 68.83, with gains
concentrated in reasoning (BBH~\citep{suzgun2023bbh} $+10.10$) and losses confined to a subset of
tasks. We therefore prioritize reliable, high-value supervision
over maximizing the number of retained examples.
\end{findingbox}

\noindent\textbf{Decontamination.} To prevent benchmark contamination and ensure reliable generalization metrics, we enforce a strict cross-benchmark decontamination protocol. The entire candidate SFT corpus is scrubbed against all downstream evaluation and validation sets using an $8$-gram matching sliding window. Any training sample exhibiting greater than 50\% $8$-gram overlap with any evaluation prompt or reference response is permanently purged from the corpus.

\noindent\textbf{Capability Balancing and Reasoning Mixture.} To determine the optimal data composition across domain capabilities, we conduct controlled mixture ablation sweeps on candidate SFT models across mathematics, code generation, multi-turn dialogue, and complex instruction following. These sweeps reveal that disproportionately scaling long chain-of-thought (CoT) trajectories introduces verbosity bias and impairs instruction adherence. We systematically perform iterative grid calibrations to harmonize the proportions of long-form reasoning, direct-response QA, and strict formatting directives, arriving at a balanced Pareto-optimal mixture.

\begin{findingbox}\textbf{\textit{Finding 5 (Long-CoT Trade-offs and Calibration):}} Long chain-of-thought (CoT) supervision exhibits non-monotonic utility: oversaturating the training mixture with extended reasoning trajectories degrades general instruction-following performance. Dynamically calibrating data mixture ratios mitigates this degradation, preserving strong instruction adherence without compromising long-CoT reasoning gains.\end{findingbox}

\noindent\textbf{Unified Think and Direct-Response Supervision.}The resulting SFT corpus deliberately interweaves \textit{think} examples (containing explicit, end-to-end reasoning chains encased in structured reasoning tags) with \textit{no-think} examples (featuring concise, immediate responses). This dual-mode design equips a single set of model weights to dynamically toggle between full deliberation and efficient zero-shot responses depending on user-specified system instructions or runtime budgets. Crucially, intermediate checkpoint evaluations confirm that joint training induces positive cross-modal transfer: exposure to structured reasoning trajectories directly sharpens the accuracy of no-think responses across coding, mathematics, and logical reasoning benchmarks (detailed in \cref{app:thinking-to-direct-transfer}).

\subsubsection{Agentic Data}

Our agentic SFT data comprises three parallel branches: deep research,
software engineering, and terminal interaction. During data construction, we
align each task family's training schema with its downstream inference
environment,
including system instructions, message roles, tool definitions, structured
calls, observation placement, and final-answer conventions. Our experiments
show that schema mismatches substantially degrade downstream agent performance.
The Deep Research schema is documented in \cref{app:deepresearch-schema}.
Within these aligned schemas, multi-turn action-observation trajectories
supervise tool selection, information gathering, environment interaction, and
final response generation.

\paragraph{Tool-Protocol Validation.}
Across agentic branches, we convert source-specific and legacy tool-call
formats into the target structured representation. We validate argument
structure, tool-call-observation pairing, unique call identifiers, and complete
assistant termination, and remove textual instructions that conflict with the
target protocol. Automatic repairs are limited to examples with unambiguous
call-result correspondence; irrecoverable examples are discarded. Stable
example identifiers preserve the links among source records, quality
decisions, and protocol revisions.

Agentic behavior also depends on instruction following, reasoning, and domain
knowledge. We therefore compare a sequential schedule that first trains on
general-capability data and then continues on agentic data with a joint schedule
that interleaves general and agentic examples throughout SFT. The joint schedule
yields stronger downstream performance and is used for the final training run.

\begin{findingbox}
\textbf{\textit{Finding 6 (The Necessity of General Data for Agent Training):}} Agent-only fine-tuning leads to degraded interaction fidelity, as agentic proficiency inherently relies on broad generalist capabilities. Co-training with general instruction data provides indispensable grounding in step-by-step reasoning and precise constraint following, yielding substantially superior agentic performance.
\end{findingbox}

\paragraph{Deep Research Trajectories.}
We collect 20,217 multi-step research trajectories and normalize them into a
shared interaction environment with \texttt{search} and \texttt{visit} tools.
Structured tool calls and environment observations follow a common
representation. We remove trajectories with invalid arguments, unmatched
action-observation pairs, or malformed tool returns. Each accepted trajectory
is provided in aligned think and no-think forms.

\paragraph{Software-Engineering Trajectories.}
We construct the software-engineering branch from collected repository-
interaction trajectories. The execution-grounded subset covers repository
interaction in executable environments, whereas the execution-free subset
provides supervision for repository understanding, file localization, tool
selection, and patch planning. After subset-specific filtering and scoring, we
retain 30,014 execution-grounded and 30,000 execution-free trajectories and
project them to the GLM-5.1 structured-tool format. The construction pipeline
is detailed in \cref{app:swe-trajectories}.

\paragraph{Terminal Trajectories.}
We collect terminal-agent trajectories from terminal task environments through
multiple independent collection routes. Each task runs in an isolated Docker
environment with a persistent shell and a structured \texttt{bash} tool,
followed by an environment-side verifier. We deduplicate trajectories within
each collection route at the task-environment level and normalize
source-specific reasoning and tool protocols. For a later controlled Agent-SFT
experiment, we materialize a paired-source view containing 22,309 trajectories
and a 15,748-trajectory verifier-successful subset. This view is an ablation
artifact and is not the terminal subset used in the main SFT run. Further
details are provided in \cref{app:terminal-trajectories}.

\subsubsection{Training Objective and Hyperparameters}
\label{sec:sft-hyperparameters}

SFT renders all examples in a common GLM-5.1 conversation format and applies
assistant-only supervision. Depending on the mode, the target contains either
an explicit reasoning trajectory followed by a final answer or a direct answer
without an exposed reasoning trace. Assistant reasoning, responses, and
structured tool calls contribute to the loss; system and user messages and tool
observations remain masked context. This mixed think/no-think objective trains
reasoning, response generation, and action selection without teaching the
model to reproduce environment feedback.
Packing and loss-mask validation are detailed in
\cref{app:sft-packing-validation}.

\begin{figure*}[h]
  \centering
  \includegraphics[width=0.49\textwidth]{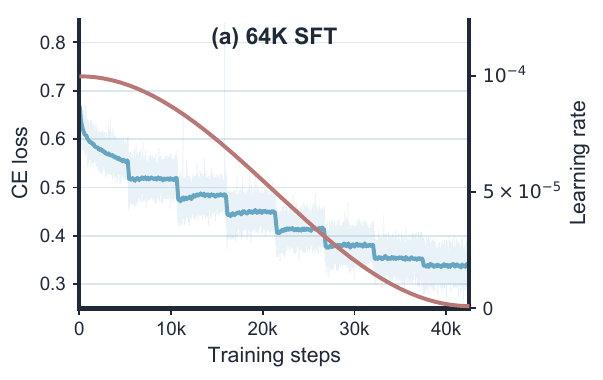}\hfill
  \includegraphics[width=0.49\textwidth]{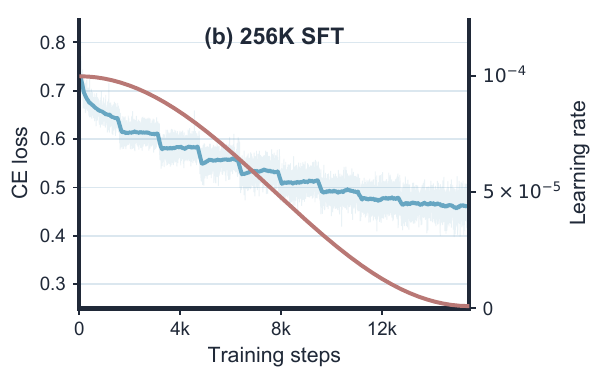}
  \caption{Training dynamics of the two length-specific SFT variants. The 64K
  run (left) and the 256K run (right) exhibit similar loss-reduction patterns:
  a rapid initial decrease followed by stepwise convergence as the learning
  rate follows a cosine decay schedule. Curves are shown from optimizer step 50
  to focus on steady-state training dynamics. Solid blue curves show a
  160-step moving average of token-level cross-entropy (CE) loss, light-blue
  curves show raw loss values, and salmon curves show the learning rate. Both
  panels use the same CE-loss scale. The released model uses the 256K SFT
  variant.}
  \label{fig:sft-training-dynamics}
\end{figure*}

We train two length-specific SFT variants with maximum sequence lengths of
65,536 and 262,144 tokens. The 64K variant is trained for eight epochs, while
the 256K variant is trained for ten epochs over 19.46B packed tokens and is used
for the released model. Both runs use a global batch size of 48. We apply
layer-wise Muon~\citep{liu2025muon} to matrix parameters and Adam to scalar
parameters, with the learning rate cosine-decayed from \(1\times10^{-4}\) to
\(1\times10^{-6}\), no warmup, a weight decay of 0.01, and gradient clipping
at 1.0. The corresponding optimization dynamics are shown in
\cref{fig:sft-training-dynamics}.

\begin{findingbox}
\textbf{\textit{Finding 7 (Long Agentic Mid-Training Bypasses Ultra-Long SFT):}}
The long-context foundation established during Mid-Training allows robust reasoning and agentic behaviors to be reliably elicited at the 256K scale using only moderate-length SFT data. This effectively bypasses the need for expensive 256K-long reasoning trajectories during post-training, demonstrating that shorter-sequence supervision suffices to activate and extrapolate long-sequence capabilities established in the mid-training.
\end{findingbox}

\subsection{Mixed Reinforcement Learning}
\label{sec:rl}

\subsubsection{RL Data Curation}

We construct a mixed reinforcement-learning corpus covering mathematics, code,
and general-capability tasks. We estimate prompt difficulty through rollouts
from the initial policy and remove problems that the model already solves
frequently, concentrating training on prompts that continue to provide useful
learning signals.

\subsubsection{Reward Design}

Rewards follow the evaluation semantics of each domain. Mathematics uses binary
answer-correctness rewards, while code receives rewards based on the fraction
of executable tests passed. General-capability tasks use corresponding
correctness or instruction-following criteria. Invalid or truncated responses
do not receive positive rewards.
Under a maximum response length of 64K tokens, we further explore combining
outcome-based rewards with the reference-policy KL regularization of
GRPO~\citep{shao2024deepseekmath} and a mild length
penalty to stabilize long-horizon optimization. The KL term controls policy
drift, while the length penalty reduces reward noise from overlong or truncated
responses~\citep{yu2025dapo}, supporting reliable multi-step reasoning without
encouraging unnecessary response expansion.

\subsubsection{Hyperparameters}

\noindent\textbf{Optimization.}
We optimize the policy with Group Relative Policy Optimization
(GRPO)~\citep{shao2024deepseekmath}. For each problem, we sample multiple
responses and construct relative advantages within the corresponding response
group. Groups with zero reward variance do not provide effective relative
learning signals; dynamic sampling therefore filters and replaces them during
training~\citep{yu2025dapo}. The actor learning rate is $2\times10^{-6}$.

\noindent\textbf{Sampling and rollouts.}
Across experiments, we explore two sampling scales. The first samples 24
problems per training step and generates 16 responses per problem, yielding
$24\times16=384$ trajectories. The second samples 384 problems and generates
8 responses per problem, yielding up to $384\times8=3{,}072$ trajectories per
step. To support long-horizon reasoning, rollouts allow up to 65,536 newly
generated tokens within a total prompt--response context of 98,304 tokens; the
actor microbatch uses the same token budget.

\section{Evaluation}
\label{sec:posttraining-evaluation}
\begin{table}[h]
  \centering
  \caption{Thinking-mode benchmark results (\%) for \zgcmeval{} and comparable-scale
  models. Dark blue cells in bold indicate the best result; light blue
  cells indicate the second-best result.}
  \label{tab:posttraining-thinking-7b-comparison}
  \newcommand{\posttrainingbenchmarkrows}{%
\rowcolor{TableGroupBlue}
\multicolumn{8}{c}{\textbf{Reasoning \& General}} \\
MATH-500 & \bestscore{97.13} & \secondscore{96.32} & 95.60 & 96.20 & 95.10 & 96.20 & 88.40 \\
AIME 2024 & \secondscore{80.62} & \bestscore{83.33} & \bestscore{83.33} & 80.00 & 71.60 & 66.67 & 43.33 \\
AIME 2025 & \secondscore{73.33} & \bestscore{75.21} & \secondscore{73.33} & 63.33 & 64.60 & 53.33 & 43.33 \\
AIME 2026 & \bestscore{75.00} & 69.17 & \secondscore{71.67} & 66.67 & 66.16 & 56.67 & 40.00 \\
HMMT 2025 & \bestscore{70.42} & \secondscore{61.50} & 52.50 & 43.33 & 43.89 & 40.00 & 23.33 \\
HMMT 2026 & \bestscore{59.48} & \secondscore{51.52} & 46.21 & 45.45 & 43.94 & 39.39 & 21.21 \\
AGIEval SAT Math & \bestscore{99.09} & 91.14 & \secondscore{98.98} & \bestscore{99.09} & 90.45 & 58.64 & 68.64 \\
AQuA-RAT & 90.57 & 90.88 & 91.39 & \bestscore{92.62} & 90.98 & \secondscore{91.80} & 81.15 \\
HARDMath-mini & 56.34 & \secondscore{61.92} & 51.89 & \bestscore{63.31} & 58.90 & 37.41 & 37.45 \\
IMO-AnswerBench & 53.00 & \bestscore{55.00} & \secondscore{54.25} & 45.00 & 45.25 & 41.50 & 26.00 \\
MATH-P-Hard & 79.21 & \secondscore{82.53} & \bestscore{84.41} & 82.08 & 77.42 & 78.14 & 59.50 \\
OlympiadBench & 76.06 & 74.75 & \bestscore{82.75} & \secondscore{82.47} & 69.55 & 74.89 & 56.37 \\
ARC-AGI-1 & \bestscore{3.50} & 1.69 & 2.58 & 3.00 & \secondscore{3.25} & 0.50 & 1.50 \\
miniF2F & 2.87 & \bestscore{5.12} & 0.00 & 2.87 & \secondscore{3.28} & 1.23 & 1.23 \\
\addlinespace[1pt]
\rowcolor{TableGroupBlue}
\multicolumn{8}{c}{\textbf{Code}} \\
HumanEval+ & \bestscore{90.24} & 88.87 & 89.63 & 80.20 & \secondscore{89.90} & 88.95 & 87.40 \\
MBPP+ & 63.23 & \secondscore{65.54} & 63.96 & \bestscore{69.10} & 64.70 & 63.96 & 61.40 \\
LiveCodeBench v6 & 46.86 & \bestscore{53.57} & 52.14 & \secondscore{52.20} & 49.26 & 47.42 & 42.43 \\
\addlinespace[1pt]
\rowcolor{TableGroupBlue}
\multicolumn{8}{c}{\textbf{Knowledge}} \\
MMLU & 73.88 & 82.09 & \secondscore{83.51} & \bestscore{85.40} & 77.80 & 78.39 & 77.40 \\
GPQA-Diamond & 47.87 & \bestscore{60.10} & 47.98 & \secondscore{59.09} & 49.94 & 54.40 & 53.70 \\
\addlinespace[1pt]
\rowcolor{TableGroupBlue}
\multicolumn{8}{c}{\textbf{Instruction Following}} \\
IFEval & 75.42 & 72.37 & 73.24 & \secondscore{87.40} & \bestscore{88.20} & 61.00 & 51.70 \\
}

  \begingroup
  \scriptsize
  \setlength{\tabcolsep}{2.0pt}
  \renewcommand{\arraystretch}{1.3}
  \newcommand{\evalhead}[1]{{\scriptsize\bfseries\centering #1}}
  \begin{tabularx}{\textwidth}{L{0.20\textwidth}*{7}{M{0.105\textwidth}}}
    \toprule
    \evalhead{Benchmark}
      & \evalhead{\shortstack{ZGCM-1\\[1.5pt]7B}}
      & \evalhead{\shortstack{DeepSeek-R1\\[1.5pt]0528-Qwen3\\[1.5pt]8B}}
      & \evalhead{\shortstack{MiniCPM\\[1.5pt]4.1-8B}}
      & \evalhead{\shortstack{Qwen3\\[1.5pt]8B}}
      & \evalhead{\shortstack{Olmo-3-7B\\[1.5pt]Think}}
      & \evalhead{\shortstack{MiMo-7B\\[1.5pt]RL}}
      & \evalhead{\shortstack{Open\\[1.5pt]Thinker\\[1.5pt]3-7B}} \\
    \midrule
    \posttrainingbenchmarkrows
    \bottomrule
  \end{tabularx}
  \endgroup
\end{table}

\subsection{Setup}

We evaluate \zgcm{} in think mode. Reported results use the released 256K SFT
checkpoint (\cref{sec:sft}). Non-agentic and web-environment
deep-research runs use a
sampling temperature of 1.0 and top-\(p\) of 1.0. Their total context budget is
262,144 tokens, comprising up to 4,096 prompt tokens and up to 258,048
generated tokens. For non-agentic benchmarks, we report mean pass@1 over 32
runs unless the benchmark specifies another aggregation rule. For
DeepSeek-R1-0528-Qwen3-8B and MiniCPM4.1-8B, GPQA-Diamond, HumanEval+,
MBPP+, and LiveCodeBench v6 are four-run means; MMLU is a single run and
IFEval uses Prompt Strict accuracy.

The non-agentic suite covers 20 benchmarks across four categories. For
mathematical and abstract reasoning, we evaluate
MATH-500~\citep{lightman2023verify}, AIME 2024/2025/2026~\citep{maa2026aime},
HMMT 2025/2026~\citep{hmmt2026}, AGIEval SAT Math~\citep{zhong2023agieval},
AQuA-RAT~\citep{ling2017aqua}, HARDMath-mini~\citep{fan2024hardmath},
IMO-AnswerBench~\citep{luong2025imoanswerbench},
MATH-P-Hard~\citep{huang2025mathperturb}, and the text-only mathematics
subset of OlympiadBench~\citep{he2024olympiadbench}, together with formal
theorem proving on miniF2F~\citep{zheng2022minif2f} and abstract grid
reasoning on ARC-AGI-1~\citep{chollet2019measure}. For code generation, we
use HumanEval+ and MBPP+~\citep{liu2023evalplus}, together with
LiveCodeBench v6~\citep{jain2024livecodebench}. For knowledge, we use
MMLU~\citep{hendrycks2021mmlu} and GPQA-Diamond~\citep{rein2024gpqa};
instruction following is measured with IFEval~\citep{zhou2023ifeval}.

Web-environment deep research uses a thinking-enabled ReAct
policy~\citep{yao2023react} with at most 64 search-and-read steps. Binary
Function Search uses a separate Ghidra-based~\citep{nsa2019ghidra} interaction protocol and decoding
configuration. Both harnesses are specified in \cref{app:agentic-evaluation}.
Web baseline values other than WebDancer are taken from public technical
reports and model releases; WebDancer and Binary Function Search baselines are
evaluated under our shared harness.

\subsection{7B Scale Model Comparison}

We compare \zgcm{} with six reasoning models at the 7B--8B scale. As shown in
\cref{tab:posttraining-thinking-7b-comparison}, the comparison spans 20
benchmarks across reasoning and general capabilities, code, knowledge, and
instruction following.

\subsection{Agentic Evaluation}

We evaluate \zgcm{} in two agentic research settings: open-web information
seeking and binary-program analysis.

\paragraph{Web-Environment Deep Research.}
Web-environment deep research requires agents to iteratively search, inspect,
and synthesize external evidence.
\Cref{tab:posttraining-agentic-comparison} compares specialized research
agents, open-weight models with tools, and proprietary systems using reported
results on WebWalkerQA~\citep{webwalker}, BrowseComp~\citep{browsecomp}, and
GAIA~\citep{mialon2024gaia}, with WebDancer~\citep{webdancer} as the
specialized research-agent baseline. \zgcm{} obtains 63.09\% on
WebWalkerQA, 19.43\% on BrowseComp, and 42.52\% on GAIA text-only.

\begin{table}[h]
  \centering
  \caption{Web-environment deep-research results (\%). Dark blue cells in bold
  indicate the best result; light blue cells indicate the second-best result.
  Dashes denote unavailable reported results.}
  \label{tab:posttraining-agentic-comparison}
  \footnotesize
  \setlength{\tabcolsep}{5pt}
  \begin{tabular}{lccc}
    \toprule
    Model / System
      & \multicolumn{1}{c}{WebWalkerQA}
      & \multicolumn{1}{c}{BrowseComp}
      & \multicolumn{1}{c}{GAIA text-only} \\
    \midrule
    \multicolumn{4}{l}{\textit{Our Model}} \\
    \textbf{\zgcmeval{}} & \bestscore{63.09} & \secondscore{19.43} & 42.52 \\
    \midrule
    \multicolumn{4}{l}{\textit{Open-source Research Agent}} \\
    WebDancer-32B & 38.40 & 2.50 & 40.70 \\
    \midrule
    \multicolumn{4}{l}{\textit{Open-weight Models with Tools}} \\
    Qwen3-235B-A22B & 59.60 & 2.30 & 45.60 \\
    Kimi-K2 & \secondscore{63.00} & 14.10 & 57.30 \\
    DeepSeek-R1 & 10.00 & 2.00 & 31.10 \\
    QwQ-32B & 4.30 & 0.50 & 22.30 \\
    \midrule
    \multicolumn{4}{l}{\textit{Proprietary Models with Tools}} \\
    GPT-4o & 33.80 & 1.90 & 34.60 \\
    GPT-5 & -- & \bestscore{54.90} & \bestscore{76.40} \\
    Claude 4 Sonnet & 61.70 & 12.20 & \secondscore{68.30} \\
    \bottomrule
  \end{tabular}
\end{table}

\paragraph{Binary Function Search.} This benchmark evaluates whether an agent can recover a target
function from a stripped ELF binary using only a behavior description.
We construct the benchmark from a broad collection effort spanning thousands
of open-source C/C++ projects, selecting several semantically meaningful
functions from each project when suitable candidates are available. The
current validated data pool contains more than 11,000 tasks from hundreds of
projects. We use a 50-task subset for evaluation, formed by randomly sampling
five tasks from each of 10 representative projects (tmux, tree, GNU Wget2, XZ
Utils, YAJL, zlib, libuv, libgit2, nm, and Lua). All 10 projects are held out from the training data and do not appear in any training split; consequently, no evaluation task is derived from a project seen during training. We plan to publicly release the
Binary Function
Search dataset to support reproducible research on tool-assisted binary
analysis.

\begin{figure}[H]
    \centering
    \includegraphics[width=\textwidth]
        {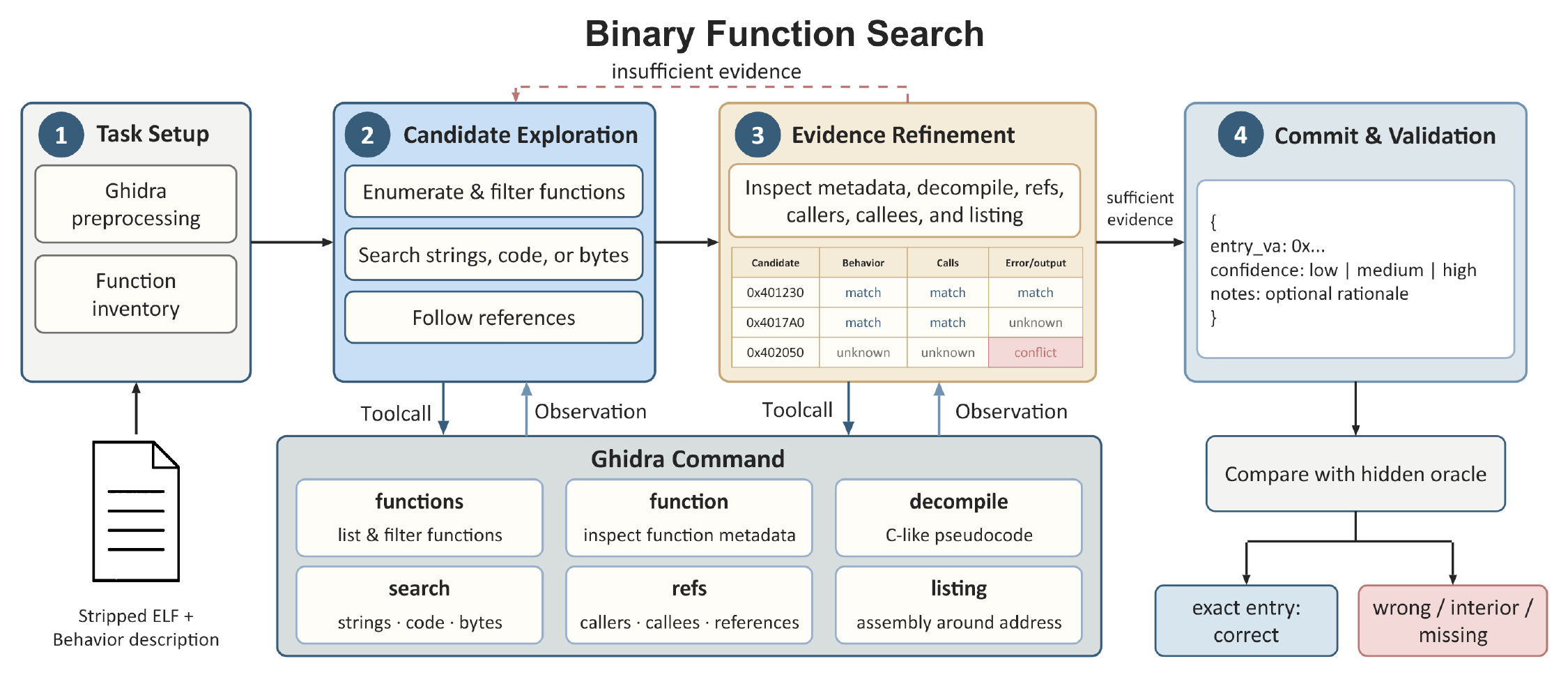}
    \caption{Binary Function Search workflow and Ghidra command
    interface. The agent iterates between candidate exploration and evidence
    refinement, then submits an exact function-entry address for hidden-oracle
    validation.}
    \label{fig:binary-function-search-workflow}
\end{figure}

As shown in \cref{fig:binary-function-search-workflow}, the agent receives
a behavior description containing clues about control flow, callees,
constants, error handling, or outputs. Through a structured Ghidra-backed
interface, it enumerates functions, searches literals and error messages,
decompiles candidates, follows references, and inspects assembly. The agent
iteratively narrows the candidate set by comparing this evidence with the
description, then submits the exact function-entry ELF virtual address. We
validate the submitted address against a hidden oracle constructed from the
corresponding unstripped binary. A submission is judged correct only if it precisely matches the exact function-entry ELF virtual address; incorrect or missing submissions are judged incorrect. Further construction and scoring details appear in
\cref{app:binary-function-search}. \zgcmeval{} achieves 62\% accuracy, far exceeding similarly sized baselines (12\% for Qwen3-8B and 0\% for the others). It remains competitive with larger frontier models, approaching GLM-5.1 (66\%) and outperforming DeepSeek-R1 and GPT-4o.

\begin{table}[t]
\centering
\caption{Binary Function Search results on 50 tasks. Valid submissions
counts answers in the required format; Correct counts exact oracle matches;
 Accuracy is Correct/50. The dark blue cell in bold and the light blue cell
indicate the best and second-best accuracy.}
\label{tab:binary-search}
\small
\begin{tabular}{lccc}
\toprule
Model / System & Valid submissions & Correct & Accuracy (\%) \\
\midrule
\multicolumn{4}{l}{\textit{Larger frontier models}} \\
DeepSeek-V4 Flash   & 50/50 & 38/50 & \bestscore{76.00} \\
Qwen3.5-397B-A17B   & 50/50 & 38/50 & \bestscore{76.00} \\
GLM-5.1             & 50/50 & 33/50 & \secondscore{66.00} \\
\textbf{\zgcmeval{}} & 50/50 & 31/50 & \textbf{62.00} \\
Kimi-K2             & 50/50 & 31/50 & 62.00 \\
DeepSeek-R1         & 46/50 & 18/50 & 36.00 \\
GPT-4o              & 50/50 &  9/50 & 18.00 \\
\midrule
\multicolumn{4}{l}{\textit{Comparable-scale models}} \\
\textbf{\zgcmeval{}} & 50/50 & 31/50 & \bestscore{62.00} \\
Qwen3-8B             & 50/50 &  6/50 & \secondscore{12.00} \\
\bottomrule
\end{tabular}
\end{table}

\section{AI-Native Research and Development}
\label{sec:ai-native-rd}
\providecommand{\ACE}{\textsc{ACE}}

\subsection{AI Assistance Across the R\&D Lifecycle}
\begin{figure*}[t]
  \centering
  \includegraphics[width=\textwidth]{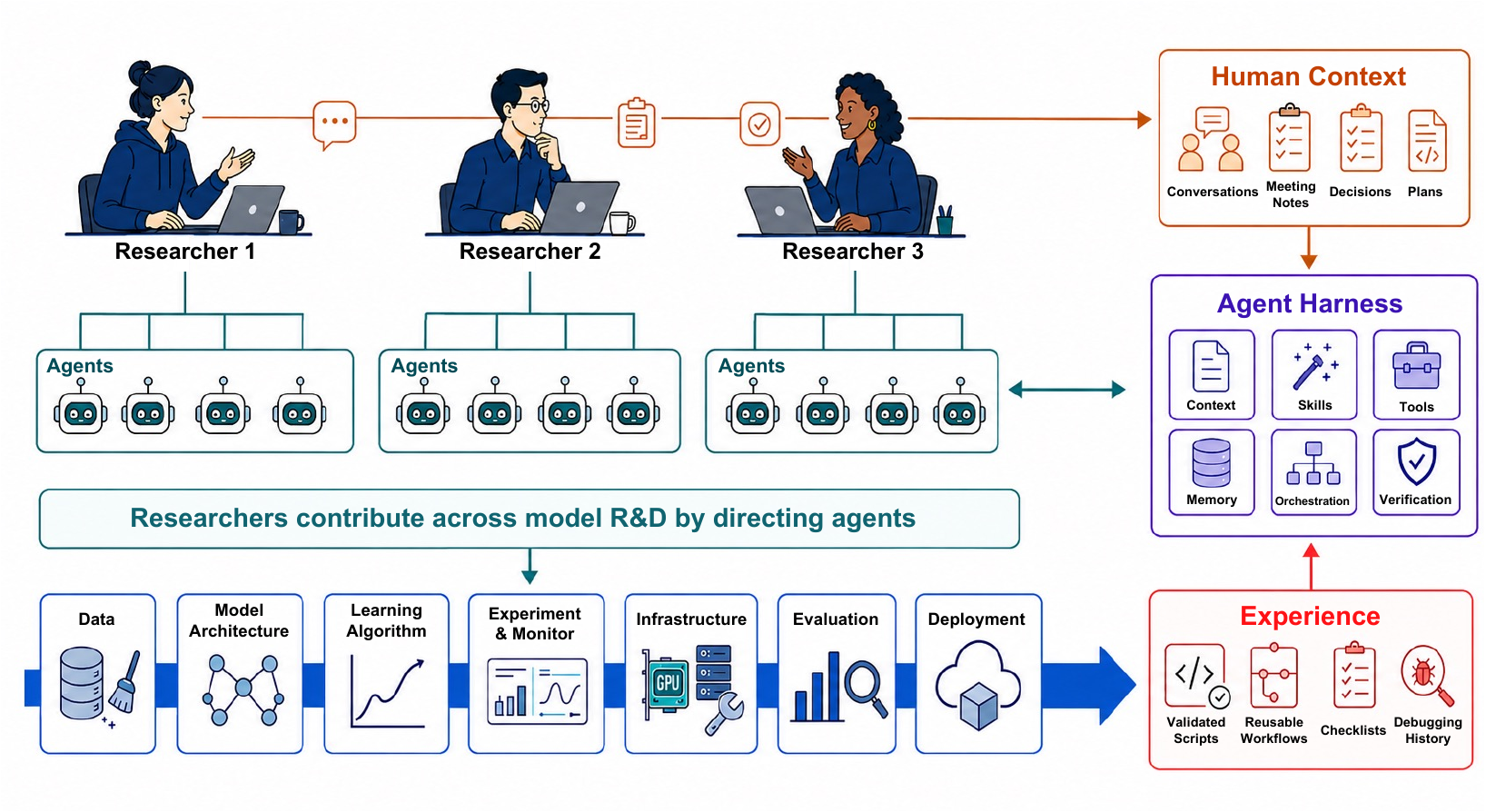}
  \caption{AI-native R\&D workflow. Each researcher directs agents across
  seven core stages of model development: data engineering, model architecture
  design, learning algorithm design, experimentation \& monitoring,
  infrastructure engineering, evaluation, and deployment engineering.
  Completed agent work produces reusable experience (validated scripts,
  workflows, checklists, debugging history), and researcher discussions
  produce human context (conversations, meeting notes, decisions, plans);
  both feed into a shared agent harness that equips all agents with
  accumulated context, skills, tools, memory, orchestration, and
  verification capabilities.}
  \label{fig:ai-native-rd}
\end{figure*}

We adopt an AI-native methodology throughout the entire model development
lifecycle. As illustrated in \cref{fig:ai-native-rd}, each researcher directs
LLM-based agents that contribute across all stages of development, from data
engineering and architecture design to learning algorithms, experimentation,
infrastructure, evaluation, and deployment. Researcher discussions produce
human context artifacts (conversations, meeting notes, decisions, plans),
while completed agent work produces reusable experience (validated scripts,
workflows, checklists, debugging history). Both streams feed into a shared
agent harness that equips every agent with the context, skills, and tools
needed for effective autonomous operation.

\para{Data cleaning.}
AI agents participate in the full data-processing pipeline. Given a set of
quality requirements, an agent first writes heuristic filtering and
transformation scripts, then executes them on the target corpus. After
processing, the agent inspects the resulting data composition and
distribution statistics to verify whether the output meets the predefined
criteria. If discrepancies are found, the agent iterates on the processing
scripts automatically, refining rules and thresholds until the data
distribution satisfies all requirements.

\para{Cluster management.}
Data processing, model training, and inference all run on a unified cluster
platform where users submit jobs, inspect logs, and manage development
machines through a single web interface. To enable AI agents to operate this
platform autonomously, we iteratively co-developed a comprehensive skill
package with our agents. The package exposes structured interfaces for
creating development machines, submitting training jobs, modifying resource
configurations, streaming logs, and filtering status fields. With these
skills, agents can independently launch experiments, monitor their progress,
retrieve and diagnose failures from logs, and adjust hyperparameters or
resource allocations without human intervention.

\para{Auto research.}
Current LLM-based agents possess sufficient long-horizon task capability to
conduct semi-autonomous research when equipped with literature search and
cluster management tools. Agents can explore algorithmic alternatives and
hyperparameter configurations, launch training runs, analyze results, and
iterate without continuous human supervision. We routinely use this
auto-research workflow during \zgcm{} development. For example, agents
profiled the data-transfer characteristics of our storage cluster and made
targeted adjustments to the training framework's I/O pipeline, yielding
substantial training-speed improvements entirely through autonomous
agent work.

\para{Evaluation.}
Model evaluation is part of our AI-native R\&D workflow, serving as a
development tool rather than only a final-stage assessment. We use
Atomic Capability Evaluation (\ACE{}), a
fine-grained diagnostic benchmark that rapidly characterizes model
capabilities and feeds back into training iterations. \ACE{}
organizes its 2,503 probes into 183 atomic capabilities across 18 categories.
Under our standard setup, a complete \ACE{} run takes approximately two to
three minutes, giving a first-order profile of strengths and weaknesses at
the category, capability, and probe levels. This makes \ACE{}
well suited to frequent evaluation during ablation studies: models trained with
different data mixtures, objectives, hyper parameters, or inference
configurations can be compared efficiently, and regressions can be
localized to affected capabilities. \ACE{} thus closes the loop between
evaluation and development, identifying which interventions help, which
introduce regressions, and where further adjustments are needed.
We construct the benchmark through iterative human-LLM collaboration. LLMs
assist with capability decomposition, probe generation, boundary-case
discovery, and scorer implementation, while researchers define capability
boundaries, audit generated artifacts, and approve their inclusion. Benchmark
execution is fully automated: deterministic programs score rule-expressible
probes, while a separate model applies predefined rubrics when correctness
cannot be expressed reliably as an executable rule. \Cref{app:ace} details the
taxonomy, scoring policies, validation, and results.

\para{Human context.}
Researchers and developers engage in frequent project discussions that,
although information-sparse, contain critical design decisions, action items,
and technical rationale. We integrate our self-developed agent platform,
ZGent, into the team's regular meeting channels. ZGent records and distills
the content of each discussion, accumulating context over time and
automatically dispatching resulting tasks to the appropriate agents for
execution. This closes the gap between human deliberation and agent action,
significantly reducing the overhead of translating meeting conclusions into
concrete engineering work.

\subsection{AI4AI Autonomy Across the R\&D Lifecycle}

\label{sec:ai-autonomy-assessment}

AI4AI (AI for AI) refers to using AI systems to build, evaluate, and improve other AI systems. To assess how much responsibility AI agents can assume across the R\&D lifecycle, nine core contributors each rated 11 task categories against the five-level rubric in Table~\ref{tab:ai-autonomy-levels}, yielding 99 ratings. The L4/L5 boundary is autonomous objective formation: L4 executes independently within human-defined objectives, whereas L5 also identifies research objectives and coordinates work across stages. The rubric is ordinal, so small differences between means are not meaningful (\cref{app:ai-autonomy-rubric} gives the task-specific criteria and all individual ratings).

\begin{table*}[t]
    \centering
    \small
    \caption{Levels of AI autonomy in R\&D workflows. The levels describe autonomy rather than productivity gains or output quality. L5 provides a reference for full autonomy, rather than an assertion of demonstrated capability.}
    \label{tab:ai-autonomy-levels}
    \begin{tabularx}{\textwidth}{@{}l l X@{}}
        \toprule
        \textbf{Level} & \textbf{Name} & \textbf{Defining criterion} \\
        \midrule
        L1 & Basic Assistance & Humans direct and execute the workflow; AI assists with individual steps. \\[4pt]
        L2 & Partial Automation & AI executes predefined workflows; humans specify procedures and handle exceptions. \\[4pt]
        L3 & Conditional Autonomy & AI adapts and manages routine workflows; consequential decisions require human approval. \\[4pt]
        L4 & High Autonomy & AI independently plans, executes, and iterates within human-defined objectives and constraints. \\[4pt]
        L5 & Full Autonomy & AI also identifies research objectives and coordinates sustained improvement across R\&D stages without routine human direction. \\
        \bottomrule
    \end{tabularx}
\end{table*}

\begin{figure*}[!t]
    \centering
    \includegraphics[width=0.80\textwidth]{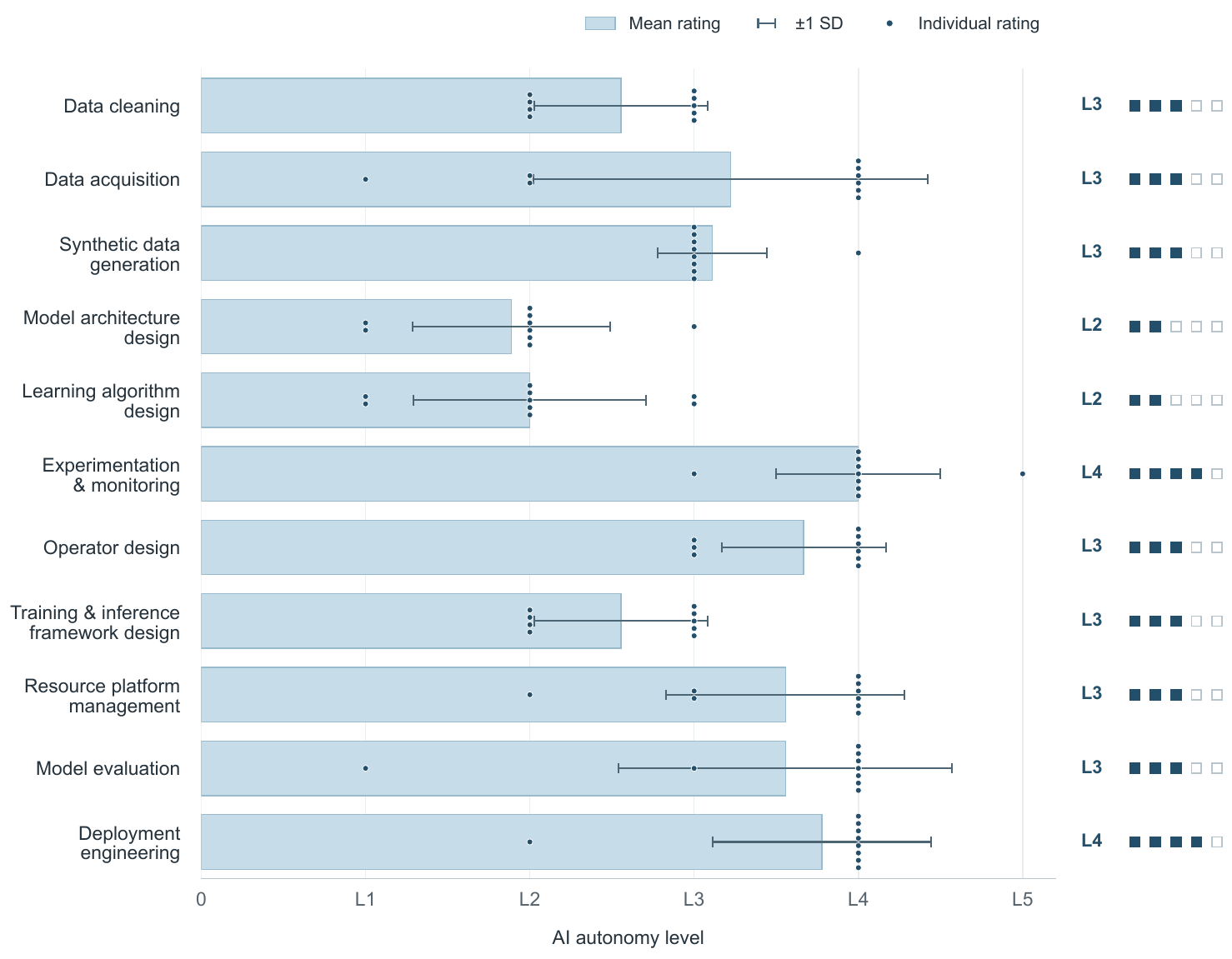}
    \caption{AI4AI autonomy across R\&D tasks, based on core contributor
    assessments. Bars show mean ratings, error bars indicate $\pm 1$ sample
    standard deviation across nine contributors, and points represent
    individual ratings. Vertical offsets separate coincident points and have
    no quantitative meaning. Levels L1--L5 are encoded as 1--5; zero is a
    plotting baseline, not a rating category. Error bars describe
    inter-contributor variation, not confidence intervals. The right column
    gives the autonomy level the team assigned to each task, with the number of
    filled markers indicating the level. Assignments are consensus judgments
    informed by the ratings.}
    \label{fig:ai-autonomy-assessment}
\end{figure*}

Autonomy is uneven across tasks (Figure~\ref{fig:ai-autonomy-assessment}). The team assigns each task a level by consensus: experimentation and monitoring and deployment engineering are L4; model architecture design and learning algorithm design are L2, and neither receives any rating above L3; the remaining seven tasks are L3. Contributors see substantially more scope for autonomous execution in operational tasks than in design tasks.

These are contributor judgments, not standardized measurements. Only one of the 99 ratings is L5, and no task is assigned above L4. Because the assigned levels range from L2 to L4, autonomy is better described per task than as a single maturity level for the whole workflow.

\begin{findingbox}
\textbf{\textit{Finding 8 (AI Autonomy Is Task-Dependent):}}
Shared context, reusable skills, and integrated tools let agents iterate with limited human intervention, but autonomy is uneven: operational tasks such as experimentation and deployment reach L4, whereas model architecture and learning algorithm design sit at L2 and never exceed L3.
\end{findingbox}
\FloatBarrier

\section{Conclusion, Limitations, and Future Directions}
\label{sec:conclusion}
\subsection{Conclusion}
In this report, we introduced \textsc{ZGCM-1}, a fully open-source 7.39B dense foundation model optimized for complex mathematical reasoning and agentic search across a 256K context. Challenging the prevailing assumption that frontier reasoning strictly necessitates hundreds of billions of parameters, \textsc{ZGCM-1} is established upon a clear core thesis: \textit{compact models can overcome inherent parametric capacity limits by coupling deliberate internal reasoning with active external tool use.} 

To realize this vision on accessible academic compute, we developed an end-to-end, high-efficiency open recipe:
\begin{itemize}[leftmargin=*]
    \item \textbf{System-Architecture Co-Design:} Interleaving gated sliding-window attention with global attention at a 5:1 ratio delivers a 3.94$\times$ throughput speedup and a 6.4$\times$ KV-cache reduction at 256K. Paired with FP8 precision, the Muon optimizer, and TWEO outlier suppression, our pipeline achieves a $\sim$4.2$\times$ pre-training time-to-loss acceleration over standard BF16/AdamW baselines.
    \item \textbf{Progressive Curriculum \& MDP Mid-Training:} Progressively expanding context windows across 600B tokens (16K $\rightarrow$ 64K $\rightarrow$ 256K) alongside Markov Decision Process (MDP) state-action modeling injects dense per-step decision supervision into raw pre-training.
    \item \textbf{Calibrated Post-Training \& AI-Native R\&D:} Rigorous tiered data pruning and unified think/no-think co-training deliver robust reasoning and direct-response efficiency from moderate-length SFT data, complemented by a mixed RL stage that yields gains on single-domain tasks such as mathematics and code. Concurrently, an autonomous AI-native multi-agent workflow drastically compressed iteration cycles.
\end{itemize}
Extensive evaluations demonstrate that \textsc{ZGCM-1-7B} achieves competitive or state-of-the-art results among sub-10B models on challenging benchmarks (e.g., 75.0\% on AIME 2026, 63.1\% on WebWalkerQA, and 62.0\% on Binary Function Search), standing toe-to-toe with models orders of magnitude larger. To foster transparent and reproducible open-science research, model weights from the pre-training, mid-training, and post-training stages, intermediate training checkpoints, data recipes, telemetry logs, and evaluation suites are made publicly available.

\subsection{Limitations}
Despite its competitive reasoning and agentic performance, \textsc{ZGCM-1} operates under several identifiable limitations:
\begin{enumerate}[leftmargin=*]
    \item \textbf{Parametric Knowledge Bound:} While deliberate thinking and external search compensate for many factual omissions, the model's static memory capacity remains fundamentally bounded by its 7.39B dense scale. In purely closed-book, recall-intensive tasks without retrieval support, \textsc{ZGCM-1} naturally trails massive frontier systems.
    \item \textbf{Instruction Adherence vs. Reasoning Verbosity:} As revealed by our fine-grained Atomic Capability Evaluation (ACE) and data mixture ablations, heavy reasoning supervision can induce verbosity and slightly impair strict, non-reasoning instruction following (e.g., IFEval and complex surface-level constraints) if not continuously calibrated.
    \item \textbf{Nascent General Software and Terminal Agency:} While \textsc{ZGCM-1} excels in structured agentic environments like web deep research and binary analysis, broader long-horizon repository-level engineering (e.g., SWE-bench Verified) and unstructured Linux terminal navigation (Terminal-Bench 2.0) present steep challenges where completion rates remain modest.
    \item \textbf{Environment and Protocol Brittleness:} The model's agentic execution relies heavily on strict schema alignment and stable environment feedback. Extreme observation noise, tool-calling format deviations, or external search API latency can still disrupt multi-step rollout trajectories.
\end{enumerate}

\subsection{Future Directions}
Our findings pave the way for several high-impact directions in efficient open-source intelligence:
\begin{enumerate}[leftmargin=*]
    \item \textbf{Extension to Sparse MoE Architectures:} Scaling the hybrid SWA, Muon optimization, and MDP mid-training recipes to sparse Mixture-of-Experts (MoE) backbones will allow expanding parametric capacity and domain specialization while preserving low inference FLOPs.
    \item \textbf{End-to-End Interactive Agentic RL:} Transitioning from token-level SFT and outcome-based reasoning rewards toward multi-turn, interactive reinforcement learning directly inside real-world execution sandboxes (e.g., terminal, web, and compiler environments).
    \item \textbf{Autonomous Dynamic Knowledge Retrieval:} Further unifying pre-training representations with on-the-fly autonomous retrieval, enabling the model to dynamically trigger search sub-routines whenever parametric uncertainty is detected.
    \item \textbf{Self-Evolving AI4AI R\&D Ecosystem:} Extending the AI-native agent harness from cluster telemetry, data curation, and atomic evaluation into autonomous hypothesis formulation, automatic kernel optimization, and closed-loop synthetic environment design.
\end{enumerate}

\section{Contributions}
\label{sec:contributions}

\textbf{Project Lead.}
Jiyan He (\href{mailto:hejiyan@zgci.ac.cn}{hejiyan@zgci.ac.cn}).

\textbf{Core Contributors} (alphabetical).
Guang Liang, Hao Liu, Haoxiang Guan, Jinbo Sun, Junyi Guo, Wenjun Feng,
Yantai Xie, Yifei Shen.

\textbf{Contributors} (alphabetical).
Bin Shao, Chuyang Wei, Kai Chen, Kexin Zhou, Minghang Zhu, Shuxin Zheng,
Tie-Yan Liu, Taine Zhao, Wenhui Zhu, Xueyin Xu, Xiaoqing Zhang, Yatao Li, Yuxuan Ren.

\bibliographystyle{assets/plainnat}
\bibliography{references}

\begin{thebibliography}{76}
\providecommand{\natexlab}[1]{#1}
\providecommand{\url}[1]{\texttt{#1}}
\expandafter\ifx\csname urlstyle\endcsname\relax
  \providecommand{\doi}[1]{doi: #1}\else
  \providecommand{\doi}{doi: \begingroup \urlstyle{rm}\Url}\fi

\bibitem[Ainslie et~al.(2023)Ainslie, Lee-Thorp, de~Jong, Zemlyanskiy,
  Lebr{\'o}n, and Sanghai]{ainslie2023gqa}
Joshua Ainslie, James Lee-Thorp, Michiel de~Jong, Yury Zemlyanskiy, Federico
  Lebr{\'o}n, and Sumit Sanghai.
\newblock {GQA}: Training generalized multi-query transformer models from
  multi-head checkpoints.
\newblock \emph{arXiv preprint arXiv:2305.13245}, 2023.
\newblock \url{https://arxiv.org/abs/2305.13245}.

\bibitem[Allal et~al.(2025)Allal, Lozhkov, Bakouch, Bl{\'a}zquez, Penedo,
  Tunstall, Marafioti, Kydl{\'\i}{\v{c}}ek, et~al.]{allal2025smollm2}
Loubna~Ben Allal, Anton Lozhkov, Elie Bakouch, Gabriel~Mart{\'\i}n
  Bl{\'a}zquez, Guilherme Penedo, Lewis Tunstall, Andr{\'e}s Marafioti, Hynek
  Kydl{\'\i}{\v{c}}ek, et~al.
\newblock {SmolLM2}: When smol goes big---data-centric training of a small
  language model, 2025.
\newblock \url{https://arxiv.org/abs/2502.02737}.

\bibitem[{Allen Institute for AI}(2025)]{allenai2025olmocrpes2o}
{Allen Institute for AI}.
\newblock {olmOCR-peS2o}.
\newblock Hugging Face dataset release, 2025.
\newblock \url{https://huggingface.co/datasets/allenai/olmOCR-pes2o-0225}.

\bibitem[{Anthropic}(2026)]{anthropic2026cryptoweaknesses}
{Anthropic}.
\newblock Discovering cryptographic weaknesses with {Claude}.
\newblock Blog post, July 2026.
\newblock
  \url{https://www.anthropic.com/research/discovering-cryptographic-weaknesses}.

\bibitem[Azerbayev et~al.(2023)Azerbayev, Schoelkopf, Paster, Dos~Santos,
  McAleer, Jiang, Deng, Biderman, and Welleck]{azerbayev2023llemma}
Zhangir Azerbayev, Hailey Schoelkopf, Keiran Paster, Marco Dos~Santos, Stephen
  McAleer, Albert~Q. Jiang, Jia Deng, Stella Biderman, and Sean Welleck.
\newblock {Llemma}: An open language model for mathematics, 2023.
\newblock \url{https://arxiv.org/abs/2310.10631}.

\bibitem[Beltagy et~al.(2020)Beltagy, Peters, and Cohan]{beltagy2020longformer}
Iz~Beltagy, Matthew~E Peters, and Arman Cohan.
\newblock {Longformer}: The long-document transformer.
\newblock \emph{arXiv preprint arXiv:2004.05150}, 2020.
\newblock \url{https://arxiv.org/abs/2004.05150}.

\bibitem[Child et~al.(2019)Child, Gray, Radford, and
  Sutskever]{child2019generating}
Rewon Child, Scott Gray, Alec Radford, and Ilya Sutskever.
\newblock Generating long sequences with sparse transformers.
\newblock \emph{arXiv preprint arXiv:1904.10509}, 2019.
\newblock \url{https://arxiv.org/abs/1904.10509}.

\bibitem[Chollet(2019)]{chollet2019measure}
Fran{\c{c}}ois Chollet.
\newblock On the measure of intelligence.
\newblock \emph{arXiv preprint arXiv:1911.01547}, 2019.
\newblock \url{https://arxiv.org/abs/1911.01547}.

\bibitem[Dao(2023)]{dao2023flashattention2}
Tri Dao.
\newblock {FlashAttention-2}: Faster attention with better parallelism and work
  partitioning.
\newblock \emph{arXiv preprint arXiv:2307.08691}, 2023.
\newblock \url{https://arxiv.org/abs/2307.08691}.

\bibitem[{DeepSeek-AI}(2024)]{deepseekv2}
{DeepSeek-AI}.
\newblock {DeepSeek-V2}: A strong, economical, and efficient mixture-of-experts
  language model.
\newblock \emph{arXiv preprint arXiv:2405.04434}, 2024.
\newblock \url{https://arxiv.org/abs/2405.04434}.

\bibitem[{DeepSeek-AI}(2026)]{deepseekv4}
{DeepSeek-AI}.
\newblock {DeepSeek-V4}: Towards highly efficient million-token context
  intelligence.
\newblock \emph{arXiv preprint arXiv:2606.19348}, 2026.
\newblock \url{https://arxiv.org/abs/2606.19348}.

\bibitem[Dehghani et~al.(2023)Dehghani, Djolonga, Mustafa, Padlewski, Heek,
  et~al.]{dehghani2023scalingvit}
Mostafa Dehghani, Josip Djolonga, Basil Mustafa, Piotr Padlewski, Jonathan
  Heek, et~al.
\newblock Scaling vision transformers to 22 billion parameters.
\newblock \emph{arXiv preprint arXiv:2302.05442}, 2023.
\newblock \url{https://arxiv.org/abs/2302.05442}.

\bibitem[Fan et~al.(2024)Fan, Martinson, Wang, Hausknecht, Brenner, Liu, Peng,
  Wang, and Brenner]{fan2024hardmath}
Jingxuan Fan, Sarah Martinson, Erik~Y. Wang, Kaylie Hausknecht, Jonah Brenner,
  Danxian Liu, Nianli Peng, Corey Wang, and Michael~P. Brenner.
\newblock {HARDMath}: A benchmark dataset for challenging problems in applied
  mathematics.
\newblock \emph{arXiv preprint arXiv:2410.09988}, 2024.
\newblock \url{https://arxiv.org/abs/2410.09988}.

\bibitem[{Gemma Team}(2025)]{gemma3}
{Gemma Team}.
\newblock Gemma 3 technical report.
\newblock \emph{arXiv preprint arXiv:2503.19786}, 2025.
\newblock \url{https://arxiv.org/abs/2503.19786}.

\bibitem[{GLM-5-Team}(2026)]{glm5}
{GLM-5-Team}.
\newblock {GLM-5}: from vibe coding to agentic engineering.
\newblock \emph{arXiv preprint arXiv:2602.15763}, 2026.
\newblock \url{https://arxiv.org/abs/2602.15763}.

\bibitem[{Harbor Framework Team}(2026)]{harbor2026}
{Harbor Framework Team}.
\newblock {Harbor}: A framework for evaluating and optimizing agents and models
  in container environments.
\newblock Software release, 2026.
\newblock \url{https://github.com/harbor-framework/harbor}.

\bibitem[{Harvard--MIT Mathematics Tournament}(2026)]{hmmt2026}
{Harvard--MIT Mathematics Tournament}.
\newblock {HMMT} february tournament archives.
\newblock Competition archive, 2026.
\newblock \url{https://www.hmmt.org/www/archive/problems}.
\newblock February 2025 and 2026 tournaments; accessed 2026-08-06.

\bibitem[He et~al.(2024)He, Luo, Bai, Hu, Thai, et~al.]{he2024olympiadbench}
Chaoqun He, Renjie Luo, Yuzhuo Bai, Shengding Hu, Zhen~Leng Thai, et~al.
\newblock {OlympiadBench}: A challenging benchmark for promoting {AGI} with
  olympiad-level bilingual multimodal scientific problems.
\newblock \emph{arXiv preprint arXiv:2402.14008}, 2024.
\newblock \url{https://arxiv.org/abs/2402.14008}.

\bibitem[Hendrycks et~al.(2021)Hendrycks, Burns, Basart, Zou, Mazeika,
  et~al.]{hendrycks2021mmlu}
Dan Hendrycks, Collin Burns, Steven Basart, Andy Zou, Mantas Mazeika, et~al.
\newblock Measuring massive multitask language understanding.
\newblock In \emph{International Conference on Learning Representations}, 2021.
\newblock \url{https://arxiv.org/abs/2009.03300}.

\bibitem[Huang et~al.(2025)Huang, Guo, Li, Ji, Ge, Li, Guo, Cai, Yuan, Wang,
  et~al.]{huang2025mathperturb}
Kaixuan Huang, Jiacheng Guo, Zihao Li, Xiang Ji, Jiawei Ge, Wenzhe Li, Yingqing
  Guo, Tianle Cai, Hui Yuan, Runzhe Wang, et~al.
\newblock {MATH-Perturb}: Benchmarking {LLM}s' math reasoning abilities against
  hard perturbations.
\newblock \emph{arXiv preprint arXiv:2502.06453}, 2025.
\newblock \url{https://arxiv.org/abs/2502.06453}.

\bibitem[Jain et~al.(2024)Jain, Han, Gu, Li, Yan,
  et~al.]{jain2024livecodebench}
Naman Jain, King Han, Alex Gu, Wen-Ding Li, Fanjia Yan, et~al.
\newblock {LiveCodeBench}: Holistic and contamination free evaluation of large
  language models for code.
\newblock \emph{arXiv preprint arXiv:2403.07974}, 2024.
\newblock \url{https://arxiv.org/abs/2403.07974}.

\bibitem[Jimenez et~al.(2024)Jimenez, Yang, Wettig, Yao, Pei, Press, and
  Narasimhan]{jimenez2024swebench}
Carlos~E. Jimenez, John Yang, Alexander Wettig, Shunyu Yao, Kexin Pei, Ofir
  Press, and Karthik Narasimhan.
\newblock {SWE-bench}: Can language models resolve real-world {GitHub} issues?
\newblock In \emph{The Twelfth International Conference on Learning
  Representations}, 2024.
\newblock \url{https://openreview.net/forum?id=VTF8yNQM66}.

\bibitem[Jordan et~al.(2024)Jordan, Jin, Boza, You, Cesista, Newhouse, and
  Bernstein]{jordan2024muon}
Keller Jordan, Yuchen Jin, Vlado Boza, Jiacheng You, Franz Cesista, Laker
  Newhouse, and Jeremy Bernstein.
\newblock Muon: An optimizer for hidden layers in neural networks.
\newblock Blog post, 2024.
\newblock \url{https://kellerjordan.github.io/posts/muon/}.
\newblock Accessed 2026-09-08.

\bibitem[{Kimi Team}(2026)]{kimik3}
{Kimi Team}.
\newblock Kimi k3: Open frontier intelligence.
\newblock \emph{arXiv preprint arXiv:2607.24653}, 2026.
\newblock \url{https://arxiv.org/abs/2607.24653}.

\bibitem[Kydl{\'\i}{\v{c}}ek et~al.(2025)Kydl{\'\i}{\v{c}}ek, Penedo, and von
  Werra]{kydlicek2025finepdfs}
Hynek Kydl{\'\i}{\v{c}}ek, Guilherme Penedo, and Leandro von Werra.
\newblock {FinePDFs-Edu}.
\newblock Hugging Face dataset release, 2025.
\newblock \url{https://huggingface.co/datasets/HuggingFaceFW/finepdfs-edu}.

\bibitem[Lambert et~al.(2024)Lambert, Morrison, Pyatkin, Huang, Ivison,
  et~al.]{lambert2024tulu3}
Nathan Lambert, Jacob Morrison, Valentina Pyatkin, Shengyi Huang, Hamish
  Ivison, et~al.
\newblock Tulu 3: Pushing frontiers in open language model post-training.
\newblock \emph{arXiv preprint arXiv:2411.15124}, 2024.
\newblock \url{https://arxiv.org/abs/2411.15124}.

\bibitem[Lee et~al.(2022)Lee, Ippolito, Nystrom, Zhang, Eck, Callison-Burch,
  and Carlini]{lee2022dedup}
Katherine Lee, Daphne Ippolito, Andrew Nystrom, Chiyuan Zhang, Douglas Eck,
  Chris Callison-Burch, and Nicholas Carlini.
\newblock Deduplicating training data makes language models better.
\newblock In \emph{Proceedings of the 60th Annual Meeting of the Association
  for Computational Linguistics}, pages 8424--8445, 2022.
\newblock \url{https://arxiv.org/abs/2107.06499}.

\bibitem[Liang et~al.(2025)Liang, Shao, Tang, Liu, and Wu]{liang2025tweo}
Guang Liang, Jie Shao, Ningyuan Tang, Xinyao Liu, and Jianxin Wu.
\newblock {TWEO}: Transformers without extreme outliers enables {FP8} training
  and quantization for dummies.
\newblock \emph{arXiv preprint arXiv:2511.23225}, 2025.
\newblock \url{https://arxiv.org/abs/2511.23225}.

\bibitem[Lightman et~al.(2023)Lightman, Kosaraju, Burda, Edwards, Baker,
  et~al.]{lightman2023verify}
Hunter Lightman, Vineet Kosaraju, Yura Burda, Harri Edwards, Bowen Baker,
  et~al.
\newblock Let's verify step by step.
\newblock \emph{arXiv preprint arXiv:2305.20050}, 2023.
\newblock \url{https://arxiv.org/abs/2305.20050}.

\bibitem[Ling et~al.(2017)Ling, Yogatama, Dyer, and Blunsom]{ling2017aqua}
Wang Ling, Dani Yogatama, Chris Dyer, and Phil Blunsom.
\newblock Program induction by rationale generation: Learning to solve and
  explain algebraic word problems.
\newblock In \emph{Proceedings of the 55th Annual Meeting of the Association
  for Computational Linguistics}, pages 158--167, 2017.
\newblock \doi{10.18653/v1/P17-1015}.
\newblock \url{https://aclanthology.org/P17-1015/}.

\bibitem[Liu et~al.(2023)Liu, Xia, Wang, and Zhang]{liu2023evalplus}
Jiawei Liu, Chunqiu~Steven Xia, Yuyao Wang, and Lingming Zhang.
\newblock Is your code generated by {ChatGPT} really correct? rigorous
  evaluation of large language models for code generation.
\newblock In \emph{Advances in Neural Information Processing Systems}, 2023.
\newblock \url{https://arxiv.org/abs/2305.01210}.

\bibitem[Liu et~al.(2025)Liu, Su, Yao, Jiang, Lai, et~al.]{liu2025muon}
Jingyuan Liu, Jianlin Su, Xingcheng Yao, Zhejun Jiang, Guokun Lai, et~al.
\newblock Muon is scalable for {LLM} training.
\newblock \emph{arXiv preprint arXiv:2502.16982}, 2025.
\newblock \url{https://arxiv.org/abs/2502.16982}.

\bibitem[Liu et~al.(2024)Liu, Zheng, Muennighoff, Zeng, Dou, Pang, Jiang, and
  Lin]{regmix}
Qian Liu, Xiaosen Zheng, Niklas Muennighoff, Guangtao Zeng, Longxu Dou, Tianyu
  Pang, Jing Jiang, and Min Lin.
\newblock {RegMix}: Data mixture as regression for language model pre-training.
\newblock \emph{arXiv preprint arXiv:2407.01492}, 2024.
\newblock \url{https://arxiv.org/abs/2407.01492}.

\bibitem[{LLM-Core-Team Xiaomi}(2025)]{mimo}
{LLM-Core-Team Xiaomi}.
\newblock {MiMo}: Unlocking the reasoning potential of language model -- from
  pretraining to posttraining.
\newblock \emph{arXiv preprint arXiv:2505.07608}, 2025.
\newblock \url{https://arxiv.org/abs/2505.07608}.

\bibitem[Luong et~al.(2025)Luong, Hwang, Nguyen, Ghiasi, Chervonyi, Seo, Kim,
  Bingham, Lee, Mishra, et~al.]{luong2025imoanswerbench}
Thang Luong, Dawsen Hwang, Hoang~H. Nguyen, Golnaz Ghiasi, Yuri Chervonyi,
  Insuk Seo, Junsu Kim, Garrett Bingham, Jonathan Lee, Swaroop Mishra, et~al.
\newblock Towards robust mathematical reasoning.
\newblock In \emph{Proceedings of the 2025 Conference on Empirical Methods in
  Natural Language Processing}, 2025.
\newblock \url{https://aclanthology.org/2025.emnlp-main.1794/}.

\bibitem[{Mathematical Association of America}(2026)]{maa2026aime}
{Mathematical Association of America}.
\newblock American invitational mathematics examination.
\newblock Mathematics competition, 2026.
\newblock
  \url{https://artofproblemsolving.com/wiki/index.php/AIME_Problems_and_Solutions}.
\newblock 2024, 2025, and 2026 competitions; problems and solutions archived by
  the Art of Problem Solving; accessed 2026-09-08.

\bibitem[Merrill et~al.(2026)Merrill, Shaw, Carlini, Li, Raj,
  et~al.]{merrill2026terminalbench}
Mike~A. Merrill, Alexander~G. Shaw, Nicholas Carlini, Boxuan Li, Harsh Raj,
  et~al.
\newblock {Terminal-Bench}: Benchmarking agents on hard, realistic tasks in
  command line interfaces.
\newblock \emph{arXiv preprint arXiv:2601.11868}, 2026.
\newblock \url{https://arxiv.org/abs/2601.11868}.
\newblock Introduces Terminal-Bench 2.0, an 89-task benchmark; release
  announcement at \url{https://www.tbench.ai/news/announcement-2-0}.

\bibitem[Mialon et~al.(2024)Mialon, Fourrier, Swift, Wolf, LeCun, and
  Scialom]{mialon2024gaia}
Gr{\'e}goire Mialon, Cl{\'e}mentine Fourrier, Craig Swift, Thomas Wolf, Yann
  LeCun, and Thomas Scialom.
\newblock {GAIA}: A benchmark for general {AI} assistants.
\newblock In \emph{International Conference on Learning Representations}, 2024.
\newblock \url{https://arxiv.org/abs/2311.12983}.

\bibitem[Micikevicius et~al.(2022)Micikevicius, Stosic, Burgess, Cornea, Dubey,
  et~al.]{micikevicius2022fp8}
Paulius Micikevicius, Dusan Stosic, Neil Burgess, Marius Cornea, Pradeep Dubey,
  et~al.
\newblock {FP8} formats for deep learning.
\newblock \emph{arXiv preprint arXiv:2209.05433}, 2022.
\newblock \url{https://arxiv.org/abs/2209.05433}.

\bibitem[{National Security Agency}(2019)]{nsa2019ghidra}
{National Security Agency}.
\newblock {Ghidra} software reverse engineering framework.
\newblock Software release, 2019.
\newblock \url{https://github.com/NationalSecurityAgency/ghidra}.

\bibitem[{NVIDIA Corporation}(2025{\natexlab{a}})]{nvidia2025nemotroncccode}
{NVIDIA Corporation}.
\newblock {Nemotron-CC-Code-v1}.
\newblock Hugging Face dataset release, 2025{\natexlab{a}}.
\newblock \url{https://huggingface.co/datasets/nvidia/Nemotron-CC-Code-v1}.

\bibitem[{NVIDIA Corporation}(2025{\natexlab{b}})]{nvidia2025nemotroncode}
{NVIDIA Corporation}.
\newblock {Nemotron Pretraining Code v1 and v2}.
\newblock Hugging Face dataset releases, 2025{\natexlab{b}}.
\newblock
  \url{https://huggingface.co/datasets/nvidia/Nemotron-Pretraining-Code-v2}.
\newblock See also Nemotron-Pretraining-Code-v1.

\bibitem[{NVIDIA
  Corporation}(2025{\natexlab{c}})]{nvidia2025nemotronspecialized}
{NVIDIA Corporation}.
\newblock {Nemotron-Pretraining-Specialized-v1}.
\newblock Hugging Face dataset release, 2025{\natexlab{c}}.
\newblock
  \url{https://huggingface.co/datasets/nvidia/Nemotron-Pretraining-Specialized-v1}.

\bibitem[{NVIDIA Corporation}(2026)]{nvidia2026megatroncore}
{NVIDIA Corporation}.
\newblock {Megatron-LM and Megatron Core}: Gpu-optimized library for training
  transformer models at scale.
\newblock Software repository and documentation, 2026.
\newblock \url{https://github.com/NVIDIA/Megatron-LM}.
\newblock Accessed 2026-08-06.

\bibitem[{OpenAI}(2024)]{openai2024swebenchverified}
{OpenAI}.
\newblock {Introducing SWE-bench Verified}.
\newblock OpenAI blog, 2024.
\newblock \url{https://openai.com/index/introducing-swe-bench-verified/}.
\newblock Accessed 2026-09-08.

\bibitem[{OpenAI}(2026)]{openai2026gpt56}
{OpenAI}.
\newblock {GPT-5.6}: Frontier intelligence that scales with your ambition.
\newblock Model release, July 2026.
\newblock \url{https://openai.com/index/gpt-5-6/}.

\bibitem[Poznanski et~al.(2025)Poznanski, Rangapur, Borchardt, Dunkelberger,
  Huff, Lin, Wilhelm, Lo, and Soldaini]{poznanski2025olmocr}
Jake Poznanski, Aman Rangapur, Jon Borchardt, Jason Dunkelberger, Regan Huff,
  Daniel Lin, Christopher Wilhelm, Kyle Lo, and Luca Soldaini.
\newblock {olmOCR}: Unlocking trillions of tokens in {PDFs} with vision
  language models, 2025.
\newblock \url{https://arxiv.org/abs/2502.18443}.

\bibitem[Pyatkin et~al.(2025)Pyatkin, Malik, Graf, Ivison, Huang, Dasigi,
  Lambert, and Hajishirzi]{pyatkin2025ifbench}
Valentina Pyatkin, Saumya Malik, Victoria Graf, Hamish Ivison, Shengyi Huang,
  Pradeep Dasigi, Nathan Lambert, and Hannaneh Hajishirzi.
\newblock Generalizing verifiable instruction following.
\newblock In \emph{Advances in Neural Information Processing Systems, Datasets
  and Benchmarks Track}, 2025.
\newblock \url{https://arxiv.org/abs/2507.02833}.

\bibitem[Qiu et~al.(2025)Qiu, Wang, Zheng, Huang, Wen, Yang, Men, Yu, Huang,
  Huang, Liu, Zhou, and Lin]{qiu2025gatedattention}
Zihan Qiu, Zekun Wang, Bo~Zheng, Zeyu Huang, Kaiyue Wen, Songlin Yang, Rui Men,
  Le~Yu, Fei Huang, Suozhi Huang, Dayiheng Liu, Jingren Zhou, and Junyang Lin.
\newblock Gated attention for large language models: Non-linearity, sparsity,
  and attention-sink-free, 2025.
\newblock \url{https://arxiv.org/abs/2505.06708}.

\bibitem[{Qwen Team}(2026)]{qwen38}
{Qwen Team}.
\newblock {Qwen3.8-Max}: A new bar for coding and cowork.
\newblock Blog post, August 2026.
\newblock \url{https://qwen.ai/blog?id=qwen3.8}.

\bibitem[Rein et~al.(2024)Rein, Hou, Stickland, Petty, Pang,
  et~al.]{rein2024gpqa}
David Rein, Betty~Li Hou, Asa~Cooper Stickland, Jackson Petty, Richard~Yuanzhe
  Pang, et~al.
\newblock {GPQA}: A graduate-level google-proof {Q\&A} benchmark.
\newblock In \emph{First Conference on Language Modeling}, 2024.
\newblock \url{https://arxiv.org/abs/2311.12022}.

\bibitem[Shao et~al.(2024)Shao, Wang, Zhu, Xu, Song,
  et~al.]{shao2024deepseekmath}
Zhihong Shao, Peiyi Wang, Qihao Zhu, Runxin Xu, Junxiao Song, et~al.
\newblock {DeepSeekMath}: Pushing the limits of mathematical reasoning in open
  language models.
\newblock \emph{arXiv preprint arXiv:2402.03300}, 2024.
\newblock \url{https://arxiv.org/abs/2402.03300}.

\bibitem[Shazeer(2020)]{shazeer2020glu}
Noam Shazeer.
\newblock {GLU} variants improve transformer.
\newblock \emph{arXiv preprint arXiv:2002.05202}, 2020.
\newblock \url{https://arxiv.org/abs/2002.05202}.

\bibitem[Su et~al.(2024)Su, Ahmed, Lu, Pan, Bo, and Liu]{su2024roformer}
Jianlin Su, Murtadha Ahmed, Yu~Lu, Shengfeng Pan, Wen Bo, and Yunfeng Liu.
\newblock {RoFormer}: Enhanced transformer with rotary position embedding.
\newblock \emph{Neurocomputing}, 568:\penalty0 127063, 2024.
\newblock \url{https://arxiv.org/abs/2104.09864}.

\bibitem[Su et~al.(2025)Su, Zhang, Li, Chen, Wang, Song, Wang, Li, Wu, Chen,
  et~al.]{agentfounder}
Liangcai Su, Zhen Zhang, Guangyu Li, Zhuo Chen, Chenxi Wang, Maojia Song, Xinyu
  Wang, Kuan Li, Jialong Wu, Xuanzhong Chen, et~al.
\newblock Scaling agents via continual pre-training.
\newblock \emph{arXiv preprint arXiv:2509.13310}, 2025.
\newblock \url{https://arxiv.org/abs/2509.13310}.

\bibitem[Suzgun et~al.(2023)Suzgun, Scales, Sch{\"a}rli, Gehrmann, Tay, Chung,
  Chowdhery, Le, Chi, Zhou, and Wei]{suzgun2023bbh}
Mirac Suzgun, Nathan Scales, Nathanael Sch{\"a}rli, Sebastian Gehrmann, Yi~Tay,
  Hyung~Won Chung, Aakanksha Chowdhery, Quoc~V. Le, Ed~H. Chi, Denny Zhou, and
  Jason Wei.
\newblock Challenging {BIG-Bench} tasks and whether chain-of-thought can solve
  them.
\newblock In \emph{Findings of the Association for Computational Linguistics:
  ACL 2023}, pages 13003--13051, 2023.
\newblock \doi{10.18653/v1/2023.findings-acl.824}.
\newblock \url{https://arxiv.org/abs/2210.09261}.

\bibitem[{SWE-agent}(2026)]{sweagentmini}
{SWE-agent}.
\newblock {mini-SWE-agent}.
\newblock Software repository, version v2, 2026.
\newblock \url{https://github.com/SWE-agent/mini-swe-agent}.
\newblock Accessed 2026-09-08.

\bibitem[{Team Olmo}(2025)]{olmo3}
{Team Olmo}.
\newblock Olmo 3.
\newblock \emph{arXiv preprint arXiv:2512.13961}, 2025.
\newblock \url{https://arxiv.org/abs/2512.13961}.

\bibitem[{Together Computer}(2023)]{together2023redpajama}
{Together Computer}.
\newblock {RedPajama-Data-1T}: An open source recipe to reproduce the {LLaMA}
  training dataset.
\newblock Hugging Face dataset release, 2023.
\newblock
  \url{https://huggingface.co/datasets/togethercomputer/RedPajama-Data-1T}.
\newblock ArXiv slice, 28B tokens.

\bibitem[{vLLM Team}(2023)]{vllm2023}
{vLLM Team}.
\newblock {vLLM}: A high-throughput and memory-efficient inference and serving
  engine for {LLMs}.
\newblock Software repository, 2023.
\newblock \url{https://github.com/vllm-project/vllm}.
\newblock See also Kwon et al., SOSP 2023, arXiv:2309.06180.

\bibitem[Wang et~al.(2025)Wang, Fu, Cai, Tang, Lyu, Fang, Zheng, Zhou, Zeng,
  Xiao, Han, and Liu]{wang2025ultrafineweb}
Yudong Wang, Zixuan Fu, Jie Cai, Peijun Tang, Hongya Lyu, Yewei Fang, Zhi
  Zheng, Jie Zhou, Guoyang Zeng, Chaojun Xiao, Xu~Han, and Zhiyuan Liu.
\newblock {Ultra-FineWeb}: Efficient data filtering and verification for
  high-quality {LLM} training data, 2025.
\newblock \url{https://arxiv.org/abs/2505.05427}.

\bibitem[Wei et~al.(2022)Wei, Bosma, Zhao, Guu, Yu, Lester, Du, Dai, and
  Le]{wei2021flan}
Jason Wei, Maarten Bosma, Vincent~Y. Zhao, Kelvin Guu, Adams~Wei Yu, Brian
  Lester, Nan Du, Andrew~M. Dai, and Quoc~V. Le.
\newblock Finetuned language models are zero-shot learners.
\newblock In \emph{International Conference on Learning Representations}, 2022.
\newblock \url{https://arxiv.org/abs/2109.01652}.

\bibitem[Wei et~al.(2025)Wei, Sun, Papay, McKinney, Han, Fulford, Chung,
  Passos, Fedus, and Glaese]{browsecomp}
Jason Wei, Zhiqing Sun, Spencer Papay, Scott McKinney, Jeffrey Han, Isa
  Fulford, Hyung~Won Chung, Alex~Tachard Passos, William Fedus, and Amelia
  Glaese.
\newblock {BrowseComp}: A simple yet challenging benchmark for browsing agents.
\newblock \emph{arXiv preprint arXiv:2504.12516}, 2025.
\newblock \url{https://arxiv.org/abs/2504.12516}.

\bibitem[Wu et~al.(2025{\natexlab{a}})Wu, Li, Fang, Yin, Zhang, Tao, Zhang, Xi,
  Fu, Jiang, Xie, Huang, and Zhou]{webdancer}
Jialong Wu, Baixuan Li, Runnan Fang, Wenbiao Yin, Liwen Zhang, Zhengwei Tao,
  Dingchu Zhang, Zekun Xi, Gang Fu, Yong Jiang, Pengjun Xie, Fei Huang, and
  Jingren Zhou.
\newblock {WebDancer}: Towards autonomous information seeking agency.
\newblock \emph{arXiv preprint arXiv:2505.22648}, 2025{\natexlab{a}}.
\newblock \url{https://arxiv.org/abs/2505.22648}.

\bibitem[Wu et~al.(2025{\natexlab{b}})Wu, Yin, Jiang, Wang, Xi,
  et~al.]{webwalker}
Jialong Wu, Wenbiao Yin, Yong Jiang, Zhenglin Wang, Zekun Xi, et~al.
\newblock {WebWalker}: Benchmarking {LLM}s in web traversal.
\newblock \emph{arXiv preprint arXiv:2501.07572}, 2025{\natexlab{b}}.
\newblock \url{https://arxiv.org/abs/2501.07572}.

\bibitem[Yang et~al.(2025{\natexlab{a}})Yang, Li, Yang, Zhang, Hui,
  et~al.]{qwen2025qwen3}
An~Yang, Anfeng Li, Baosong Yang, Beichen Zhang, Binyuan Hui, et~al.
\newblock {Qwen3} technical report.
\newblock \emph{arXiv preprint arXiv:2505.09388}, 2025{\natexlab{a}}.
\newblock \url{https://arxiv.org/abs/2505.09388}.

\bibitem[Yang et~al.(2025{\natexlab{b}})Yang, Yu, Li, Liu, Huang,
  et~al.]{qwen2025qwen25_1m}
An~Yang, Bowen Yu, Chengyuan Li, Dayiheng Liu, Fei Huang, et~al.
\newblock {Qwen2.5-1M} technical report.
\newblock \emph{arXiv preprint arXiv:2501.15383}, 2025{\natexlab{b}}.
\newblock \url{https://arxiv.org/abs/2501.15383}.

\bibitem[Yang et~al.(2024)Yang, Jimenez, Wettig, Lieret, Yao, Narasimhan, and
  Press]{yang2024sweagent}
John Yang, Carlos~E. Jimenez, Alexander Wettig, Kilian Lieret, Shunyu Yao,
  Karthik~R. Narasimhan, and Ofir Press.
\newblock {SWE}-agent: Agent-computer interfaces enable automated software
  engineering.
\newblock In \emph{The Thirty-eighth Annual Conference on Neural Information
  Processing Systems}, 2024.
\newblock \url{https://arxiv.org/abs/2405.15793}.

\bibitem[Yao et~al.(2023)Yao, Zhao, Yu, Du, Shafran, et~al.]{yao2023react}
Shunyu Yao, Jeffrey Zhao, Dian Yu, Nan Du, Izhak Shafran, et~al.
\newblock {ReAct}: Synergizing reasoning and acting in language models.
\newblock In \emph{International Conference on Learning Representations}, 2023.
\newblock \url{https://arxiv.org/abs/2210.03629}.

\bibitem[Yu et~al.(2025)Yu, Zhang, Zhu, Yuan, Zuo, et~al.]{yu2025dapo}
Qiying Yu, Zheng Zhang, Ruofei Zhu, Yufeng Yuan, Xiaochen Zuo, et~al.
\newblock {DAPO}: An open-source {LLM} reinforcement learning system at scale.
\newblock \emph{arXiv preprint arXiv:2503.14476}, 2025.
\newblock \url{https://arxiv.org/abs/2503.14476}.

\bibitem[Zhang and Sennrich(2019)]{zhang2019root}
Biao Zhang and Rico Sennrich.
\newblock Root mean square layer normalization.
\newblock \emph{Advances in Neural Information Processing Systems}, 32, 2019.
\newblock \url{https://arxiv.org/abs/1910.07467}.

\bibitem[Zhang et~al.(2026)Zhang, Mohamed, Abdine, Shang, and
  Vazirgiannis]{zhang2026curriculum}
Yang Zhang, Amr Mohamed, Hadi Abdine, Guokan Shang, and Michalis Vazirgiannis.
\newblock Beyond random sampling: Efficient language model pretraining via
  curriculum learning.
\newblock In \emph{Proceedings of the 19th Conference of the European Chapter
  of the Association for Computational Linguistics (Volume 1: Long Papers)},
  pages 5776--5794. Association for Computational Linguistics, 2026.
\newblock \doi{10.18653/v1/2026.eacl-long.271}.
\newblock \url{https://aclanthology.org/2026.eacl-long.271/}.

\bibitem[Zheng et~al.(2022)Zheng, Han, and Polu]{zheng2022minif2f}
Kunhao Zheng, Jesse~Michael Han, and Stanislas Polu.
\newblock {miniF2F}: A cross-system benchmark for formal olympiad-level
  mathematics.
\newblock In \emph{International Conference on Learning Representations}, 2022.
\newblock \url{https://arxiv.org/abs/2109.00110}.

\bibitem[Zhong et~al.(2023)Zhong, Cui, Guo, Liang, Lu, Wang, Saied, Chen, and
  Duan]{zhong2023agieval}
Wanjun Zhong, Ruixiang Cui, Yiduo Guo, Yaobo Liang, Shuai Lu, Yanlin Wang, Amin
  Saied, Weizhu Chen, and Nan Duan.
\newblock {AGIEval}: A human-centric benchmark for evaluating foundation
  models.
\newblock \emph{arXiv preprint arXiv:2304.06364}, 2023.
\newblock \url{https://arxiv.org/abs/2304.06364}.

\bibitem[Zhou et~al.(2026)Zhou, Lyu, Lin, Zhao, Guo, Zhang, Xue, Ma, Zhou,
  Wang, and Liu]{zhou2026ultradatamath}
Chuyue Zhou, Hongya Lyu, Xinle Lin, Hengyu Zhao, Junshao Guo, Xueren Zhang,
  Shuaikang Xue, Qiang Ma, Jie Zhou, Yudong Wang, and Zhiyuan Liu.
\newblock {UltraData-Math}.
\newblock Hugging Face dataset release, 2026.
\newblock \url{https://huggingface.co/datasets/openbmb/UltraData-Math}.

\bibitem[Zhou et~al.(2023)Zhou, Lu, Mishra, Brahma, Basu,
  et~al.]{zhou2023ifeval}
Jeffrey Zhou, Tianjian Lu, Swaroop Mishra, Siddhartha Brahma, Sujoy Basu,
  et~al.
\newblock Instruction-following evaluation for large language models.
\newblock \emph{arXiv preprint arXiv:2311.07911}, 2023.
\newblock \url{https://arxiv.org/abs/2311.07911}.

\end{thebibliography}
\newpage
{\hypersetup{linkcolor=black}
\begin{spacing}{1.35}
\fontsize{11pt}{14pt}\selectfont
\tableofcontents
\end{spacing}
}

\newpage
\beginappendix
\crefalias{section}{appendix}
\crefalias{subsection}{appendix}
\crefalias{subsubsection}{appendix}
\section{Pre-Training Details}
\label{app:pretraining-details}

\subsection{Training Configurations}
\label{app:training-configurations}

\Cref{tab:appendix-training-configurations} lists the distributed-training
configurations for general pre-training (16K context) and long-context
qualification (64K and 256K). All runs use H100 GPUs with FP8
computation. We evaluate two routes to 256K: a direct route that trains 30B
tokens at 256K, and a staged route that first trains 10B tokens at 64K then
20B tokens at 256K. The staged route reaches a slightly lower final loss
(1.16 vs.\ 1.19) at comparable throughput.

\begin{table}[htbp]
  \centering
  \caption{Distributed-training configurations on H100 GPUs. TP = tensor
  parallelism, PP = pipeline parallelism, CP = context parallelism, DP = data
  parallelism, MBS = micro-batch size, GBS = global batch size.}
  \label{tab:appendix-training-configurations}
  \small
  \resizebox{\linewidth}{!}{%
  \begin{tabular}{@{}lcccccccl@{}}
    \toprule
    \textbf{Stage} & \textbf{Context} & \textbf{TP/PP/CP/DP} & \textbf{MBS/GBS}
      & \textbf{Recompute} & \textbf{LR} & \textbf{RoPE base} & \textbf{Notes} \\
    \midrule
    General pre-training & 16K & 2/1/1/96 & 1/768 & None
      & $2\times10^{-4}$ & 5M & ${\sim}$60\% BF16-eq.\ MFU ($\approx$585 TFLOP/s/GPU) \\
    Staged (step 1) & 64K & 2/1/4/24 & 1/192 & Selective
      & $1\times10^{-4}$ & 5M & 10B tokens \\
    Direct 256K & 256K & 2/2/4/12 & 1/48 & Full
      & $1\times10^{-4}$ & 10M & 30B tokens; 203 TFLOP/s/GPU \\
    Staged (step 2) & 256K & 2/2/4/12 & 1/48 & Full
      & $1\times10^{-4}$ & 10M & 20B tokens; 204 TFLOP/s/GPU \\
    \bottomrule
  \end{tabular}}
\end{table}

\subsection{Capability Dynamics during General Pre-Training}
\label{app:pretraining-capability-dynamics}

\begin{figure}[htbp]
  \centering
  \includegraphics[width=\textwidth]{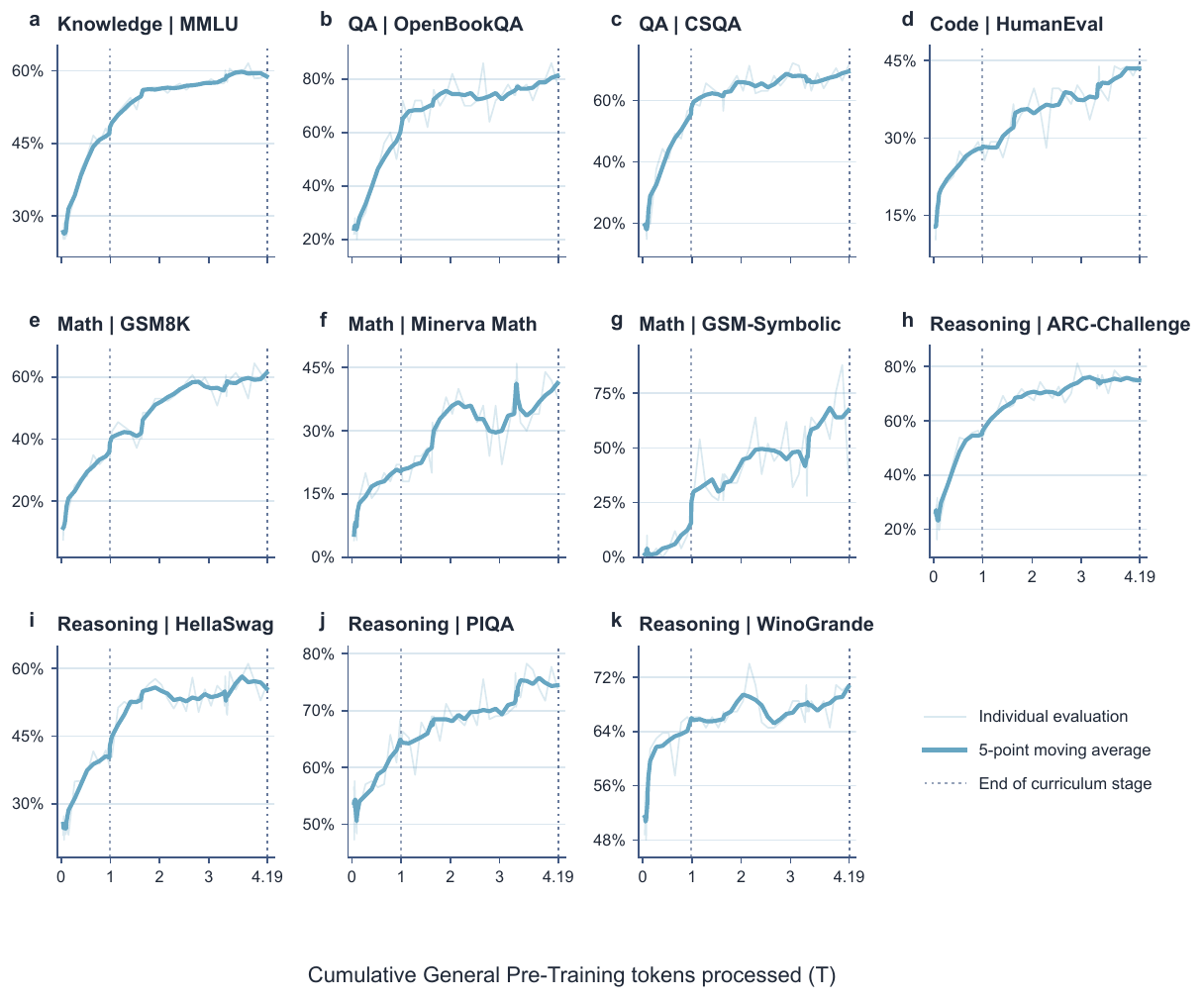}
  \caption{Capability trends during General Pre-Training. Each point is
  obtained by directly evaluating the corresponding base checkpoint without
  supervised fine-tuning or other post-training adaptation. The x-axis reports
  cumulative tokens processed, truncated at 4.19T. The dotted line at 1.0T
  marks the end of the curriculum-ordered stage. Light lines show individual
  evaluations, and dark lines show five-point moving averages.}
  \label{fig:pretraining-capability-trends}
\end{figure}

We directly evaluate 54 saved base checkpoints, from 0.04T to 4.19T cumulative
General Pre-Training tokens processed. Each checkpoint is evaluated in its
base-model form and generates responses without supervised fine-tuning,
instruction tuning, or other post-training adaptation. The benchmark harness
and evaluation settings remain fixed across checkpoints, so the curves reflect
capability changes in the pre-trained model itself. Token positions are read
from the training logs as consumed samples times sequence length, which keeps
them exact across the global-batch-size ramp. The resulting trajectories are
shown in \cref{fig:pretraining-capability-trends}.

Evaluation uses fixed subsets rather than full test sets: 1{,}425 items for
MMLU (25 per subject), 1{,}004 for HellaSwag, and between 50 and 200 items for
the remaining benchmarks. GSM-Symbolic, Minerva Math, and OpenBookQA use 50
items each, so a single item shifts their score by two points. Point-to-point
variation on these three benchmarks reflects sampling noise as much as
capability change, and only their overall trend is informative.

\Cref{tab:appendix-pretraining-capability-summary} summarizes four
five-checkpoint windows. The first and last columns average the first and last
five checkpoints; the two middle columns average five checkpoints centered near
1.0T and 2.5T tokens. Scores are reported as percentages.

\begin{table}[htbp]
  \centering
  \caption{Five-checkpoint window averages during General Pre-Training.
  Higher is better.}
  \label{tab:appendix-pretraining-capability-summary}
  \small
  \begin{tabular}{@{}llcccc@{}}
    \toprule
    \textbf{Benchmark} & \textbf{Capability} &
      \textbf{Early} & \textbf{Around 1.0T} & \textbf{Around 2.5T} &
      \textbf{Final} \\
    & & \textbf{(0.04T--0.07T)} & \textbf{(0.91T--1.16T)}
      & \textbf{(2.29T--2.79T)} & \textbf{(3.67T--4.18T)} \\
    \midrule
    MMLU          & Knowledge & 26.37 & 48.32 & 56.94 & 59.52 \\
    \midrule
    OpenBookQA    & QA        & 24.80 & 60.80 & 72.40 & 78.80 \\
    CSQA          & QA        & 19.02 & 57.70 & 64.26 & 67.87 \\
    \midrule
    HumanEval     & Code      & 13.66 & 28.29 & 36.46 & 43.54 \\
    \midrule
    GSM8K         & Math      & 11.52 & 38.64 & 57.12 & 59.09 \\
    Minerva Math  & Math      &  6.80 & 20.40 & 32.80 & 38.40 \\
    GSM-Symbolic  & Math      &  1.20 & 25.60 & 49.20 & 64.00 \\
    \midrule
    ARC-Challenge & Reasoning & 24.96 & 56.07 & 69.74 & 75.90 \\
    HellaSwag     & Reasoning & 24.56 & 42.75 & 52.69 & 57.15 \\
    PIQA          & Reasoning & 53.80 & 64.78 & 69.78 & 74.89 \\
    WinoGrande    & Reasoning & 50.71 & 65.98 & 66.14 & 68.98 \\
    \bottomrule
  \end{tabular}
\end{table}

Every benchmark improves from the first window to the last, but the rate at
which they improve diverges. Most of the gain arrives before 2.5T tokens: MMLU
covers 92\% of its total improvement by the 2.5T window, ARC-Challenge 88\%,
and WinoGrande 84\%. The final 1.7T tokens add 2.58 points on MMLU, 1.97 on
GSM8K, and 2.84 on WinoGrande. Code and the harder mathematics benchmarks keep
improving over the same interval, with HumanEval adding 7.08 points and
GSM-Symbolic 14.80, though the latter is measured on 50 items and carries wide
uncertainty. These diverging dynamics suggest that broad pre-training alone
provides increasingly uneven marginal returns and motivate the
capability-oriented mixture used in Mid-Training.

\FloatBarrier
\subsection{Capability Development during Mid-Training}
\label{app:midtraining-short-sft-probe}

We use a short-SFT probe to track the downstream capability of successive
training milestones. Each checkpoint is rapidly adapted with the same
carefully curated SFT corpus and then evaluated on a fixed benchmark suite.
The probe measures the capability elicited from each initialization after
lightweight supervised adaptation. It is therefore distinct from the direct
base-checkpoint evaluation in
\cref{app:pretraining-capability-dynamics}.

\begin{figure}[htbp]
  \centering
  \includegraphics[width=\textwidth]{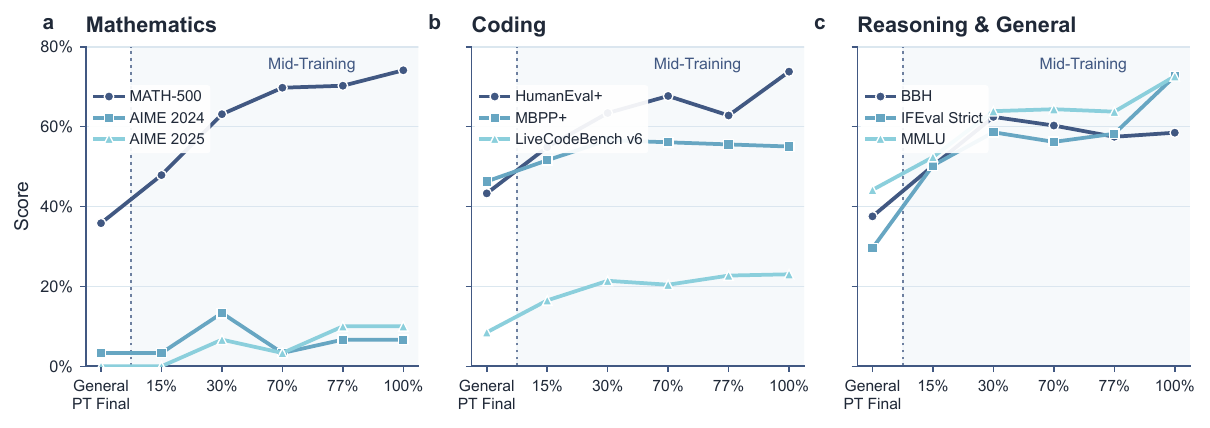}
  \caption{Capability trajectories across General Pre-Training and
  Mid-Training milestones after rapid fine-tuning on the same SFT corpus.
  The General Pre-Training probe and the 15\%/30\%/70\%/77\%
  Mid-Training probes use two SFT epochs; the final (100\%) Mid-Training probe
  uses three. The curves show the downstream capability exposed by short SFT
  rather than direct base-model performance.}
  \label{fig:midtraining-short-sft-probe}
\end{figure}

The probe shows a large shift after entering Mid-Training. From the end of
General Pre-Training to the end of Mid-Training, MATH-500 rises from 35.83\%
to 74.12\%, HumanEval+ from 43.29\% to 73.78\%, IFEval Strict from 29.57\%
to 72.64\%, and MMLU from 44.17\% to 72.63\%. Individual benchmarks are not
monotonic at every checkpoint.

Because the final Mid-Training probe uses three SFT epochs while the preceding
probes use two, these trajectories are diagnostic evidence of improving
downstream potential rather than a strictly matched causal ablation. Together
with the direct base-model trends in
\cref{app:pretraining-capability-dynamics}, they support the use of
capability-oriented Mid-Training before full post-training.

\FloatBarrier
\subsection{Curriculum Pretraining}
\label{app:curriculum-learning}

We compare two 7B runs trained for 10B tokens with identical data mixtures.
The random baseline samples documents in random order. For the curriculum run,
we first exclude documents with extreme lexical-complexity scores as outliers;
these cases are typically corrupted or garbled text rather than meaningful
difficult examples. We then order the remaining natural-language documents
from lower to higher lexical complexity. Code and mathematics do not use
lexical complexity as a difficulty proxy; they are shuffled independently and
interleaved at their target proportions.

\begin{table}[htbp]
  \centering
  \caption{Bits per byte (BPB) after 10B training tokens. Lower is better.}
  \label{tab:appendix-curriculum}
  \small
  \begin{tabular}{@{}lcc@{}}
    \toprule
    \textbf{Benchmark} & \textbf{Random} & \textbf{Curriculum} \\
    \midrule
    Coding   & 1.99 & \textbf{0.81} \\
    Math     & 0.97 & \textbf{0.94} \\
    \midrule
    MMLU     & \textbf{1.03} & 1.12 \\
    ARC      & \textbf{1.05} & 1.09 \\
    HellaSwag & \textbf{1.05} & 1.08 \\
    \bottomrule
  \end{tabular}
\end{table}

Curriculum ordering improves coding BPB (1.99 $\to$ 0.81) and
math BPB (0.97 $\to$ 0.94). General benchmarks (MMLU, ARC, HellaSwag) show a
modest increase of 0.03--0.09 BPB, suggesting that the curriculum
accelerates domain-specific learning at a small cost to general knowledge
acquisition at this token budget.

\subsubsection{Qualitative Examples: Why Code and Mathematics Are Interleaved Separately}
\label{app:curriculum-case-studies}

The examples below illustrate a limitation of using lexical complexity as a
single ordering signal. Percentile positions are recomputed within the
corresponding source shard and are reported only as diagnostic positions; they
are not intrinsic difficulty labels or production metadata.

\paragraph{Code.}
The first example is a WFC-based grid-construction program. It appears near the
low end of the ordering (first 0.01\% of the code shard), despite containing
substantial program structure and environment-specific logic.

\begin{casebox}{Code example: low end of the ordering}
\small\ttfamily
import wfc\_implementation as wfci\\
import mapUtils, interfaceUtils\\
buildArea = interfaceUtils.requestBuildArea()\\
tiles = np.array([[0,0,0,0], [0,1,0,0], [1,1,0,1]])
\end{casebox}

{\noindent\scriptsize Diagnostic position: first 3 of 25,000 documents.\par}

By contrast, the following two-line NumPy snippet appears near the high end of
the ordering (last 0.01\%), although it performs only a simple file-loading
operation.

\begin{casebox}{Code example: high end of the ordering}
\small\ttfamily
import numpy as np\\
count\_data = np.loadtxt("../BookSource/Chapter1\_Introduction/data/txtdata.csv")
\end{casebox}

{\noindent\scriptsize Diagnostic position: last 3 of 25,000 documents.\par}

These two examples show that surface lexical complexity can be dominated by
imports, boilerplate, and token-level variation rather than by the algorithmic
content of a program.

\paragraph{Mathematics.}
The same issue occurs in mathematical text. A long combinatorial programming
problem, ``Manhattan Wiring,'' appears near the low end of the ordering (first
0.11\% of the mathematics shard), even though it requires non-intersecting path
construction under obstacles and a minimum-length objective.

\begin{casebox}{Mathematics example: low end of the ordering}
\small
Connect two pairs of marked cells with non-intersecting horizontal or vertical
lines around obstacles, while minimizing the total length of the two lines.
\end{casebox}

{\noindent\scriptsize Diagnostic position: first 63 of 55,743 documents.\par}

At the opposite end, a one-line statement of the associative law appears near
the high end of the ordering (last 0.01\%), despite requiring little
mathematical reasoning.

\begin{casebox}{Mathematics example: high end of the ordering}
\small
This shows the associative property of addition: $(a+b)+c=a+(b+c)$.
\end{casebox}

{\noindent\scriptsize Diagnostic position: last 3 of 55,743 documents.\par}

The mathematics examples reinforce the same design choice: lexical complexity
is useful for efficiently ordering general-language data, but it is not a
reliable proxy for the reasoning difficulty of code or mathematical content.
We therefore interleave these two domains separately throughout curriculum
pretraining.

\subsection{Mid-Training Mixtures}
\label{app:midtraining-data}

Mid-training uses continued pre-training with a full-sequence causal
objective (not chat-template supervision). Instruction records are serialized
as raw context, and agent actions and environment observations remain
next-token targets.

We apply pool-wide deduplication and mathematics error filtering to produce a
candidate pool of approximately 5.3B records (2.86T tokens). The pool is then
sampled into three cumulative context mixtures:

\begin{table}[htbp]
  \centering
  \caption{Mid-training context mixtures. Each longer-context stage retains
  shorter-context data.}
  \label{tab:appendix-midtraining-length}
  \small
  \begin{tabular}{@{}lcc@{}}
    \toprule
    \textbf{Stage} & \textbf{Sampled tokens} & \textbf{Length composition} \\
    \midrule
    16K  & 180B  & 180B at $\leq$16K \\
    64K  & 240B  & 180.89B at $\leq$16K + 59.11B at 16K--64K \\
    256K & 180.51B  & 127.81B at $\leq$16K + 21.72B at 16K--64K + 30.98B above 64K \\
    \bottomrule
  \end{tabular}
\end{table}

Code and mathematics each occupy approximately 20\% across all three stages.
The 256K stage increases knowledge from 10.5\% to 14.2\% and agentic data
from 1.5\% to 3.3\%, while reducing pre-training replay from 13\% to 9\%.

\subsubsection{Reasoning and Instruction Data}
\label{app:reasoning-instruction-data}

Reasoning sources are normalized to a common schema and classified into
accepted, rewrite, pending, holdout, or rejected categories based on
self-containment, answer evidence, and reasoning quality. The reasoning pool
contains approximately
35M examples (38B tokens). Of these, 34.7M are retained directly via
rule-based cleaning; 201K examples with valuable questions but low-quality
traces are reconstructed by teacher models and re-admitted after validation.

Instruction data is collected from multiple sources (including Tulu- and
FLAN-derived data)~\citep{lambert2024tulu3,wei2021flan} and converted into
pre-training-compatible context.
Selected single-turn examples are rewritten as multi-turn discussions to
provide supervision for sequential reasoning.

\subsubsection{Agentic Data}
\label{app:agentic-midtraining-data}

Agentic mid-training data combines general interaction traces with
software-engineering trajectories.

For general interaction traces, we retain high-quality complete interactions
that cover long-horizon task context, tool calls, and environment feedback.
We additionally reformulate retained traces as Markov decision process-style
single-step decision examples. At each decision point, the task description,
interaction history, and current environment feedback form the state, and the
model predicts an appropriate next action conditioned on that state. This
shifts the learning target from reproducing complete state-transition sequences
to state-conditioned action selection, providing denser per-step supervision
alongside the full-trajectory signal.

For software-engineering trajectories, we apply different filtering strategies
depending on whether a real execution environment is available. Execution-free
trajectories are screened using structural, patch, shell/edit, log, and
behavioral evidence rather than self-reported runtime outcomes.
Execution-grounded trajectories are jointly scored and filtered according to
actual execution results, feedback from the final interaction turn, and the
number of interaction turns. Retained trajectories preserve task context, tool
calls, observations, code changes, and final responses. After deduplication,
the final software-engineering trajectory corpus contains approximately 129K
records (5.9B tokens). These trajectories are trained with full-sequence causal
language modeling, where both agent actions and environment observations are
next-token targets.

\section{Post-Training Implementation Details}
\label{app:posttraining-details}

\subsection{General SFT Data Selection}
\label{app:sft-data-selection}

General SFT data is assembled from open-source datasets and internally
distilled examples spanning instruction following, knowledge, code,
mathematics, science, dialogue, and reasoning. The shared quality-selection
and decontamination procedures are described in \cref{sec:sft}; this appendix
records domain-specific selection and implementation details.

We compare candidate domain mixtures across instruction following, reasoning,
and other capabilities. Excessive long-reasoning shares reduce instruction
following, so we iteratively adjust long-reasoning, direct-response, and
instruction-following proportions before selecting the final composition.

Within the general SFT data, the reasoning component retains 1,307,175 of
2,268,178 candidate examples after structural validation, answer-quality
review, and reasoning-value filtering. Accepted examples must contain
coherent, complete, and substantively useful reasoning in addition to a
reliable final answer. Safety and multi-turn examples pass task-appropriate
structural checks, and reasoning content remains separate from visible final
answers throughout processing.

\subsubsection{Evidence for Quality-First Selection}
\label{app:sft-quality-ablation}

To isolate the effect of data quality, we start from the same base checkpoint
and quickly fine-tune it into a conversational model using three versions of a
common SFT pool. The first version receives only format normalization and basic
structural checks, without model-based quality filtering. The second removes
examples assigned to review or deletion while retaining broad capability
coverage. The third keeps only examples that pass the strictest quality,
style, and capability checks. These counts refer to this controlled ablation
pool rather than to the final SFT corpus.

\begin{table}[H]
  \centering
  \caption{Ablation of SFT data filtering. All variants use the same base
  checkpoint, rapid SFT procedure, and evaluation settings. The final column
  is the mean over the six shared benchmarks.}
  \label{tab:sft-quality-ablation}
  \scriptsize
  \setlength{\tabcolsep}{3.5pt}
  \begin{tabular}{@{}lcccccccc@{}}
    \toprule
    Data processing & Samples & IFEval & MATH-500 & BBH & MMLU
      & HumanEval+ & MBPP+ & Mean (6) \\
    \midrule
    Minimal processing & $\approx$2.08M & 72.64 & 74.12 & 58.49 & 72.63
      & 73.78 & 55.03 & 67.78 \\
    Broader filtering & $\approx$1.833M & 71.20 & 77.63 & 65.08 & 70.60
      & 65.00 & 58.36 & 67.98 \\
    Quality-focused filtering & $\approx$1.145M & 71.94 & 75.43 & 68.59 & 71.50
      & 67.56 & 57.94 & \textbf{68.83} \\
    \bottomrule
  \end{tabular}
\end{table}

The quality-focused version uses 44.9\% fewer examples than the minimally
processed version and 37.5\% fewer than the broadly filtered version. Despite
its smaller size, its mean over the six shared benchmarks rises from 67.78 and
67.98 to 68.83, respectively. Individual tasks do not all move in the same
direction, but the aggregate result shows that high-quality data is more useful
than simply increasing data volume.

\subsection{Agentic SFT Data}
\label{app:sft-traces}

We compare two schedules for introducing agentic supervision. The sequential
schedule first trains on general-capability data and then continues on agentic
data. The joint schedule interleaves general and agentic examples throughout
SFT. Candidate-run evaluations favor the joint schedule, which is used for the
final training run.

\subsubsection{Deep-Research Trajectories}
\label{app:deepresearch-schema}

\paragraph{Interaction schema.}
The Deep Research system instruction is:

\begin{quote}
\small\itshape
You are a deep search assistant. Your primary role is to perform rigorous,
multi-step, multi-source investigations on any topic, covering both broad
open-domain questions and highly specialized academic inquiries. For each user
request, you must actively seek out and cross-check information from credible
and diverse sources, then integrate the findings into a response that is
comprehensive, accurate, well-structured, and objective. When you have gathered
sufficient information and are ready to provide the definitive response,
enclose the entire final answer in <answer></answer> tags.
\end{quote}

The interaction exposes two tools:

\begin{table}[H]
  \centering
  \caption{Tool schema for Deep Research trajectories.}
  \label{tab:deepresearch-tools}
  \small
  \begin{tabular}{@{}L{0.12\linewidth}L{0.34\linewidth}L{0.43\linewidth}@{}}
    \toprule
    Tool & Arguments & Return \\
    \midrule
    \texttt{search} &
      \texttt{query}: nonempty string array;\newline
      \texttt{topn}: positive integer &
      Ranked results containing titles, URLs, and snippets for each query. \\
    \texttt{visit} &
      \texttt{url}: nonempty URL array;\newline
      \texttt{goal}: natural-language information goal &
      Goal-directed information extracted from the requested pages. \\
    \bottomrule
  \end{tabular}
\end{table}

Each example follows the message sequence
\(\texttt{system}\rightarrow\texttt{user}\rightarrow
\texttt{assistant}\rightarrow\texttt{tool}\), with assistant-tool exchanges
repeated as needed before the final assistant response. Tools are declared in
the top-level \texttt{tools} field; assistant actions contain a tool name and JSON
arguments in \texttt{tool\_calls}; and each environment return is a separate
\texttt{role=tool} message paired with the preceding call. Search observations
use a ranked result list, while visit observations provide information relevant
to the stated goal. The final response is enclosed in
\texttt{<answer>...</answer>}.

Think examples include assistant reasoning before actions and the final answer.
Their aligned no-think views remove the reasoning spans while preserving the
user request, tool calls, observations, and answer.

\subsubsection{Software-Engineering Trajectories}
\label{app:swe-trajectories}

\paragraph{Execution-grounded trajectory construction and filtering.}
The execution-grounded source pool contains 34,269 trajectories. We require
valid nonempty tool calls, interaction with external tools, tool observations,
and a complete finish action. We remove trajectories containing private paths,
secret-like strings, empty completion content, or a finish message that still
describes work in progress. This first pass retains 30,914 trajectories and
rejects 3,355: 3,067 for private paths, 254 for secret patterns, and 34 for
in-progress completion language. All 30,914 retained trajectories pass the
GLM-5.1 tokenizer and supervision-boundary preflight without tokenizer errors
or empty targets. Restricting the view to 8K--64K tokens removes 900 overlength
trajectories and yields 30,014 execution-grounded examples.

\paragraph{Execution-free trajectory construction, filtering, and scoring.}
The execution-free source pool contains 318,115 trajectories. The initial
structural and safety gate retains 281,237 candidates and rejects 36,878:
35,259 for private paths, 1,361 for secret patterns, and 258 for incomplete
finish language. GLM-5.1 tokenization produces 280,852 candidates in the
8K--64K range. Structural validity alone is insufficient for execution-free
data, so we construct a source-specific quality rubric through model-assisted
rule iteration. Hard admission criteria require a patch, both shell and edit
actions, a unique nonempty finish action, reasonable interaction and target
lengths, and no unsupported claim that an unexecuted test has passed. This gate
rejects 3,554 trajectories and leaves 277,298 scored candidates.

The scoring function assigns higher scores to trajectories with coherent patch
evidence, complementary shell and editing behavior, balanced tool and reasoning
actions, dense file-path and code-symbol evidence, sufficient assistant target
tokens, informative file and terminal observations, and limited observation
clipping. It assigns lower scores to trajectories with excessive turns or
reasoning actions, heavy clipping, uncertain completion language, or
runtime-success claims unsupported by an execution environment. We select the
top 30,000 trajectories while capping each repository at 800 examples to limit
source concentration. The selected subset has a minimum quality score of 97, a
median of 97, and a 95th percentile of 100. An independent audit confirms that
all selected examples are marked execution-free and contain no illegal tools
or empty finish actions.

\paragraph{Structured-tool projection and validation.}
Quality selection is performed on traceable native records before
training-format conversion. We then project the 30,014 execution-grounded and
30,000 execution-free trajectories to the GLM-5.1 top-level \texttt{tools}
schema. External
\texttt{execute\_bash} and \texttt{str\_replace\_editor} actions become
structured assistant \texttt{tool\_calls}; native \texttt{think} actions become
\texttt{reasoning\_content}; and \texttt{finish} actions become the final
assistant response. Internal acknowledgements such as ``Your thought has been
logged.'' are removed, consecutive assistant messages are merged, valid
observations remain separate tool messages, and a trailing unpaired tool
observation is dropped. The resulting 60,014 software-engineering candidates
pass enhanced structural checks for JSON validity, top-level tool definitions,
tool-call-observation pairing, duplicate identifiers, and complete assistant
termination.

\subsubsection{Terminal Trajectories}
\label{app:terminal-trajectories}

\paragraph{Environment and schema.}
We collect terminal-agent trajectories from terminal task environments through
two independent collection routes, covering Claude Opus 4.6 Thinking and
GLM-5.2. Deduplication is performed
independently within each collection route using the task-environment identity.
Each task runs in an isolated Docker environment with a persistent shell and
one structured \texttt{bash} tool. A rollout permits at most 64 interaction
steps and 4,096 output tokens per assistant turn. Commands have a 180-second
timeout, and each terminal observation is capped at 12,000 characters.

The agent terminates through a predefined submission command, after which
\texttt{tests/test.sh} is executed in the same environment to obtain a verifier
outcome independent of the model's own completion claim. The full trajectory,
run summary, and provenance manifest are retained so that the task, collection
route, execution, interaction trace, and verifier outcome remain traceable.

\paragraph{Construction, filtering, and source normalization.}
Raw rollouts contain repeated task runs, malformed tool protocols, missing
containers or images, setup output, harness-generated format-repair turns,
duplicated visible reasoning, and trailing unpaired observations. Within each
collection route, we retain at most two valid trajectories for a task
environment, preferring verifier-successful and then more complete runs. We
reject repeated protocol failures and infrastructure failures because they do
not represent valid task interactions.

Every assistant turn contains \texttt{content}, while native
\texttt{reasoning\_content} is available only for a subset of turns. Claude
Opus 4.6 Thinking does not expose \texttt{reasoning\_content} on every turn,
whereas GLM-5.2 does. We normalize the two fields across sources without
duplicating supervision. Setup output,
harness \texttt{Format error} repairs, and infrastructure-failure text are
removed from model-visible content. A terminal observation that would otherwise
end the conversation is cropped so that the example closes with a complete
assistant turn. Structurally valid unsuccessful trajectories retain their
outcome labels, enabling both full and success-only materializations.

\paragraph{Materialized training views.}
For a later controlled Agent-SFT experiment, we materialize a paired-source view
from the Claude Opus 4.6 Thinking and GLM-5.2 collection routes. The Claude
source retains 11,653 of 11,944 raw runs, including 8,488 verifier-successful
and 3,165 unsuccessful trajectories. The GLM-5.2 source retains 10,656 of
11,351 pointer-manifest records, including 7,260 successful and 3,396
unsuccessful trajectories.

\subsection{Packing Validation}
\label{app:sft-packing-validation}

Before packing, a tools-aware chat-template and loss-mask preflight verifies
tool rendering, assistant supervision boundaries, action-observation closure,
sequence length, and nonempty targets. Accepted, overlength, and rendering-error
records are routed to separate outputs rather than being silently truncated or
discarded.

\subsection{Thinking-to-Direct Transfer}
\label{app:thinking-to-direct-transfer}

A historical checkpoint comparison examines whether reasoning-oriented
supervision can improve direct-response capability. A baseline checkpoint
trained with direct-response data was subsequently retrained with a mixture of
direct-response examples and explicit reasoning trajectories. Both checkpoints
were evaluated in no-think mode, so the comparison measures answer quality
without exposing an intermediate reasoning trace.

\begin{table}[H]
  \centering
  \caption{No-think performance before and after mixed think/no-think SFT.
  Both columns use matching benchmark versions and metrics; rows are sorted by
  score change.}
  \label{tab:thinking-data-direct-transfer}
  \small
  \begin{tabular}{@{}lccc@{}}
    \toprule
    Benchmark & Before mixed SFT & After mixed SFT & Change \\
    \midrule
    AIME 2025 & 10.00 & 43.33 & \textbf{+33.33} \\
    AIME 2024 & 6.67 & 33.33 & \textbf{+26.66} \\
    MBPP+ & 55.03 & 75.93 & \textbf{+20.90} \\
    BBH & 58.49 & 69.48 & \textbf{+10.99} \\
    LiveCodeBench v3 & 28.00 & 34.31 & \textbf{+6.31} \\
    IFEval & 72.64 & 78.00 & \textbf{+5.36} \\
    MMLU & 72.63 & 70.03 & -2.60 \\
    HumanEval+ & 73.78 & 66.46 & -7.32 \\
    \bottomrule
  \end{tabular}
\end{table}

The largest gains occur in mathematical reasoning and code generation. AIME
2025 increases from 3/30 to 13/30, AIME 2024 increases from 2/30 to 10/30,
and MBPP+ improves by 20.90 points. BBH, LiveCodeBench v3, and IFEval also
improve. These results are consistent with reasoning supervision transferring
useful problem-solving behavior to direct-response inference even when the
intermediate reasoning is not visible.

The transfer is task-dependent rather than universal: MMLU decreases by 2.60
points and HumanEval+ by 7.32 points.

This experiment is an observational checkpoint comparison, not a controlled
data ablation. The mixed-data run may differ in total training tokens, data
composition, checkpoint history, and optimization schedule.

\section{Agentic Evaluation Protocol}
\label{app:agentic-evaluation}

This appendix documents the evaluation harnesses used for the web-environment
results in \cref{tab:posttraining-agentic-comparison}, the Binary Function Search
results in \cref{tab:binary-search}, and our internal software-engineering
and terminal-agent diagnostics.

\subsection{Web-Environment Deep-Research Harness}

The deep-research evaluation uses a thinking-enabled ReAct policy with at most
64 search-and-read steps. Search queries are issued through Serper with Jina
Search as a fallback, and pages are extracted through Jina Reader. Retrieved
material is summarized by \zgcm{} operating in direct-response (no-thinking)
mode, while final answers are judged by Qwen3-30B-A3B-Instruct-2507. If the
interaction reaches the step budget, the harness requests a final answer from
the evidence accumulated so far.

Results under this protocol are reported in
\cref{tab:posttraining-agentic-comparison}. Their interpretation depends on search
date and availability, provider behavior, page extraction, summarization,
judge settings, and retry policy.

\subsection{Binary Function Search}
\label{app:binary-function-search}

\para{Task Definition.}
Binary Function Search evaluates whether an agent can locate a target
function in a stripped ELF binary from a natural-language description. Stripping removes most symbolic clues, while compiler
optimization may transform the source-level structure. Moreover, several
functions may share the same strings, constants, or callees. 
The agent therefore must distinguish functions by combining semantic and structural evidence.

\para{Input and Output.}
Each task specifies the target binary, its architecture, the address-space
convention, and a behavioral description of the target function.
A description may include several kinds of semantic evidence:
\begin{itemize}
    \item the function's main behavior or transformation;
    \item characteristic control-flow decisions or processing stages;
    \item calls to other routines or interactions with subsystems;
    \item distinctive constants, strings, flags, or data formats;
    \item error detection, recovery, and reporting behavior; and
    \item externally visible outputs or state changes.
\end{itemize}

In most cases, the description covers only a subset of the evidence
categories listed above.
The following example presents input of an actual task:
\begin{verbatim}
Binary path: /chroot/build/binutils/nm
Architecture: x86_64
Address space: elf_va

Function description:
Formats one symbol record for output. It selects between
size-based and regular modes, optionally prints source
locations, resolves relocation information, classifies the
symbol, and dispatches the result to the active output-style
printer.
\end{verbatim}

The required output is the exact function-entry ELF virtual address,
together with an optional confidence level and a short rationale:

\begin{verbatim}
{
  "entry_va": "0x...",
  "confidence": "low|medium|high",
  "notes": "optional short rationale"
}
\end{verbatim}

The submitted address must identify the beginning of the target function.
File offsets, runtime addresses, and interior instructions are not valid
outputs. The confidence and notes fields do not affect correctness.

\para{Harness and Tools.}
The Harness provides a Ghidra-backed~\citep{nsa2019ghidra} environment for Binary Function Search.
It exposes \texttt{gcmd}, a structured interface to the results of Ghidra
preprocessing. The lower panel of
\cref{fig:binary-function-search-workflow} summarizes its six actions:

\begin{itemize}
    \item \texttt{functions} returns a compact, filterable list of identified
          functions;
    \item \texttt{function} returns structured metadata for a specified
          function;
    \item \texttt{decompile} returns C-like pseudocode for a specified
          function;
    \item \texttt{search} searches strings, disassembly, decompiled text,
          or byte sequences;
    \item \texttt{refs} finds references to a function, string, data object,
          or address; and
    \item \texttt{listing} returns the assembly listing around a code
          location.
\end{itemize}

The Harness manages the Ghidra session, normalizes addresses to the ELF
virtual-address convention, and verifies that submitted addresses correspond
to function boundaries. It also manages large tool outputs, detects repeated
identical calls, and records the interaction for auditing.

\para{Agentic Workflow.}
Figure~\ref{fig:binary-function-search-workflow} summarizes the workflow,
which consists of four stages.

\textit{\Rmnum 1. Task setup.}
Before the first model turn, the Harness imports the stripped binary into
Ghidra to recover function boundaries, control-flow structures, references,
strings, assembly listings, and decompiled representations. The agent can
access this information through the structured tool interface. The Harness
also provides a summary of the function inventory, such as the number and size
distribution of analyzed functions, without selecting or ranking candidates.

\textit{\Rmnum 2. Candidate exploration.}
The agent searches the set of functions identified by Ghidra using information
from the task description. It enumerates functions, applies structural
filters, and searches strings, disassembly, decompiled text, or byte
sequences. It then follows references from relevant search results to the
functions that use them. This process reduces the function set to a shortlist
of plausible candidates.

\textit{\Rmnum 3. Evidence refinement.}
For each shortlisted candidate, the agent examines function metadata,
decompiled code, references, callers, callees, and assembly listings. It
compares the collected information with the evidence categories present in
the task description, including behavior, control flow, calls, constants,
errors, and outputs. The agent also examines caller-callee relationships to
determine which function directly owns the described behavior.
If the collected evidence is insufficient, the workflow returns to candidate
exploration. The agent performs additional searches, follows new references,
or updates the candidate shortlist. The two stages repeat until the evidence
is sufficient for a final decision.

\textit{\Rmnum 4. Commit and validation.}
Once the agent has collected sufficient evidence and is confident in its
selection, it submits the entry address of the candidate function. The Harness
checks that the address corresponds to a function boundary identified during
Ghidra preprocessing, normalizes it to the ELF virtual-address convention,
and compares it with the hidden oracle.

During candidate exploration and evidence refinement, each agent turn produces
at most one structured \texttt{gcmd} call based on the task and previous
observations. The Harness executes the call and returns its result to the
agent, forming an iterative reasoning-action-observation loop. The workflow
ends when the agent submits an answer or reaches the turn limit.

\para{Benchmark Construction.}
We construct the dataset from thousands of open-source C/C++ projects.
For each retained project, we build paired stripped and unstripped binaries:
the stripped binaries form the benchmark inputs, while symbols and source
mappings in the unstripped binaries identify compiled function entries and
construct the hidden oracle. The validated pool contains 11,767 tasks across
529 projects.

We map source definitions to compiled entries and retain project-native
functions with a clear purpose and meaningful, self-contained behavior, such
as core algorithms, parsers, state transitions, data transformations,
validation logic, resource management, and major request or command workflows.
We exclude tests, bundled dependencies, thunks, aliases, compiler-generated
variants, trivial utilities, placeholder functions, generic error handlers,
and wrappers that delegate the described behavior. Task counts therefore vary
by project. For each selected function, a generation model produces a behavior
description from its source implementation and supporting evidence. The
description may cover inputs, state changes, control-flow decisions, calls,
constants, errors, and externally visible effects, but omits the function name,
source location, address, and recognizable name variants. We reject incomplete
records, leaked identifiers, and addresses that do not map to verified function
entries, and deduplicate tasks using source-function, binary-address, and
description similarity.

The reported evaluation samples 50 tasks from this larger pool. We
randomly select five tasks from each of 10 representative projects---tmux,
tree, GNU Wget2, XZ Utils, YAJL, zlib, libuv, libgit2, nm, and Lua. This
project-balanced subset covers diverse software domains and function
behaviors.

\para{Scoring.}
A prediction is correct only when the submitted ELF virtual address
exactly matches one of the accepted oracle entries. Every accepted alternative
must be a verified function start that independently implements the complete
described behavior; related callers, callees, file offsets, runtime addresses,
and interior instructions are incorrect. A run receives zero credit if it
submits an incorrect entry or fails to produce a valid submission, including
non-convergent repeated tool use and runtime or API failures. We use the same
Ghidra-backed interaction protocol and task budget for all compared models and
compute accuracy over all 50 tasks, without excluding invalid submissions.

\subsection{SWE-bench Verified Mini50 Harness}

We evaluate software-engineering agent capability on a fixed internal 50-task
subset of SWE-bench Verified~\citep{jimenez2024swebench,openai2024swebenchverified}. The evaluation has two
stages. First, mini-SWE-agent v2~\citep{sweagentmini,yang2024sweagent} with tool calling performs an interactive
repair trajectory inside each task's Docker \texttt{/testbed} using a structured
\texttt{bash} tool and submits a unified-diff patch. The official SWE-bench
scorer then applies the patch and executes the task-specific
\texttt{FAIL\_TO\_PASS} and \texttt{PASS\_TO\_PASS} tests. The evaluation uses
a fixed Mini50 manifest, a temperature of zero, 10 shards of five tasks each,
a 250-step interaction budget per task, and a 1,024-token output limit per
turn. The ZGCM evaluation service uses a 256K-token vLLM~\citep{vllm2023} context window with
string-formatted tool observations, automatic tool choice, and GLM tool-call
and reasoning parsers. Under this protocol, the best fully audited ZGCM-family
evaluation resolves 2 of 50 tasks (4.0\%).

\subsection{Terminal-Bench 2.0 Harness}

We evaluate terminal-agent capability on the standard 89-task Terminal-Bench
2.0 benchmark~\citep{merrill2026terminalbench} using task-specific Docker images and Harbor~\citep{harbor2026} for execution and
verification. A thinking-enabled terminal agent using the GLM XML protocol
interacts through a persistent \texttt{bash(command)} tool for at most 64 steps
per task, with a 4,096-token output limit per turn, a temperature of zero,
top-\(p\) of 0.95, a 180-second timeout per command, and a one-hour timeout per
task. The vLLM inference service uses a 128K-token context window with the
GLM-5.1 tool parser and GLM-4.5 reasoning parser. The 89 tasks are partitioned
across four shards, with one attempt per task. Only tasks affected by
predeclared infrastructure failures or verifier timeouts are rerun with
identical parameters after cache warm-up. Under this protocol, the best fully
audited ZGCM-family evaluation resolves 2 of 89 tasks (2.25\%).

\section{Atomic Capability Evaluation Framework}
\label{app:ace}

Atomic Capability Evaluation (\ACE{}) diagnoses model behavior at a finer
granularity than aggregate benchmarks. The framework comprises 18 categories,
183 atomic capabilities, and 2,503 executable probes. This appendix specifies
the taxonomy, scoring method, validation procedure, and category-level results.

\subsection{Design Principles}

\ACE{} follows four design principles. \textbf{(P1) Atomicity.} Each capability
has a narrow semantic boundary, so a failure identifies a specific behavior
rather than only a broad domain. \textbf{(P2) Probe diversity.} Each capability
is tested through multiple surface forms at three author-assigned difficulty
levels (L1--L3), reducing sensitivity to any single wording.
\textbf{(P3) Intent-aware scoring.} The evaluator applies strict response
matching only when formatting is the target behavior; otherwise, it extracts
and scores the substantive answer while normalizing non-semantic variation.
\textbf{(P4) Traceability.} Every decision links to the probe, reference answer,
scoring policy, extracted answer, and failure record.

\subsection{Taxonomy and Scoring}

Across 18 capability categories, \ACE{} decomposes model behavior into 183
atomic capabilities. Each atomic capability targets a specific behavior and
is evaluated with multiple probes that vary in surface form and difficulty.
Table~\ref{tab:ace-taxonomy} summarizes the scope and allocation of this
taxonomy. The 2,503 probes comprise 835 L1, 835 L2, and 833 L3 instances.
These labels denote three levels of increasing probe difficulty.

\begin{table}[h]
 \centering
 \caption{Evaluated \ACE{} taxonomy, category scope, and probe allocation.}
 \label{tab:ace-taxonomy}
 \scriptsize
 \setlength{\tabcolsep}{4pt}
 \renewcommand{\arraystretch}{1.03}
 \begin{tabularx}{\textwidth}{@{}lccZ@{}}
  \toprule
  Category & Caps. & Probes & Primary evaluation scope \\
  \midrule
  Agent & 8 & 120 & Environment interaction, long-horizon execution,
   recovery, memory, and coordinated action. \\
  Arithmetic & 13 & 135 & Numerical operations, comparison, conversion,
   sorting, and exact calculation. \\
  Causal & 5 & 75 & Mechanisms, interventions, counterfactuals, and
   latent-state inference. \\
  Code & 26 & 389 & Program understanding, generation, debugging,
   optimization, testing, and artifact editing. \\
  Dialogue & 6 & 90 & Multi-turn coherence, clarification, coreference,
   intent, persona, and subtext. \\
  Instruction & 13 & 195 & Compliance with format, length, lexical,
   structural, stylistic, and multi-constraint requirements. \\
  Knowledge & 8 & 96 & Recall and application of facts, formulas, APIs,
   conventions, and common mechanisms. \\
  Language & 27 & 321 & Lexical, syntactic, semantic, multilingual, and
   character-level language processing. \\
  Logic & 10 & 102 & Boolean, deductive, inductive, constraint, set,
   strategic, and structural reasoning. \\
  Long Context & 3 & 45 & Position-robust retrieval, multi-span integration,
   and multi-document synthesis. \\
  Mathematics & 11 & 165 & Algebra, discrete mathematics, limits,
   optimization, geometry, number theory, and proof. \\
  Metacognition & 6 & 90 & Ambiguity and error detection, self-checking,
   uncertainty, and plan revision. \\
  Output & 7 & 105 & Final-answer control, exact formatting, fluent prose,
   units, and structured-artifact generation. \\
  Planning & 6 & 90 & Subgoal decomposition, forward and backward
   reasoning, state tracking, experiment design, and persistence. \\
  Reading & 11 & 141 & Targeted retrieval, distractor filtering, passage
   integration, structured lookup, and cross-document reading. \\
  Tool Use & 7 & 105 & Tool selection, argument extraction, schema
   compliance, invocation, and multi-tool coordination. \\
  Truthfulness & 8 & 119 & Calibrated abstention, hallucination resistance,
   source fidelity, infeasibility detection, and misconception resistance. \\
  World Modeling & 8 & 120 & Physical, spatial, temporal, social, procedural,
   and theory-of-mind reasoning. \\
  \midrule
  \textbf{Total} & \textbf{183} & \textbf{2,503} & \\
  \bottomrule
 \end{tabularx}
\end{table}

Table~\ref{tab:ace-level-example} illustrates how these difficulty levels are
instantiated for one atomic capability. The example is drawn from the
\emph{argument extraction} capability in the Tool Use category: L1 extracts a
single scalar argument, L2 combines unit conversion with multiple arguments,
and L3 constructs nested list and object arguments.

\begin{table}[H]
 \centering
 \caption{Example L1--L3 probes for argument extraction.}
 \label{tab:ace-level-example}
 \small
 \setlength{\tabcolsep}{6pt}
 \renewcommand{\arraystretch}{1.08}
 \begin{tabularx}{\textwidth}{@{}lZL{0.31\textwidth}@{}}
  \toprule
  Level & Example probe & Expected tool arguments \\
  \midrule
  L1 & \texttt{set\_timer(minutes: int)}. ``Start a timer for 45 minutes.''
   & \texttt{minutes=45} \\
  L2 & \texttt{ship\_package(weight\_grams: int, speed: standard/overnight)}.
   ``Send a 2.5 kg parcel by overnight delivery.''
   & \texttt{weight\_grams=2500, speed=overnight} \\
  L3 & \texttt{submit\_order(items, shipping)}. ``Order 3 of SKU A100 and 2 of
   SKU B250, and ship them to 12 Oak Lane in Canada.''
   & \texttt{items=[(A100,3),(B250,2)]};\newline
    \texttt{shipping=(12 Oak Lane, Canada)} \\
  \bottomrule
 \end{tabularx}
\end{table}

\ACE{} evaluates each probe with a scorer selected for the target capability.
Among the 2,503 probes, 2,155 (86.1\%) use deterministic scoring. Numeric
scorers recognize answer markers, final lines, boxed expressions, units, and
equivalent fractions. String scorers accept quoted and sentence-embedded
answers, while multiple-choice scorers restrict letter matching to audited
answer positions. Strict full-response matching is reserved for tasks that
explicitly test format, length, or the absence of extra text. Code probes use
abstract-syntax-tree checks or isolated unit tests, and tool-use probes validate
arguments against JSON Schema. For the remaining 348 probes (13.9\%), whose
correctness cannot be expressed through a reliable executable criterion, the
automated pipeline invokes a separate model with a predefined binary rubric.
Both scoring paths execute within the same automated pipeline; human review is
limited to benchmark construction and offline validation. Probe outcomes are
aggregated into atomic-capability scores with L1--L3 breakdowns; these scores
localize improvements and regressions, while category-level aggregation
summarizes broader patterns.

\subsection{AI-Assisted Benchmark Construction}

We build the benchmark through iterative human-LLM co-design. At the taxonomy
level, LLMs propose decompositions, boundary cases, and missing dimensions;
researchers consolidate these proposals into capabilities with explicit
evaluation objectives. At the probe level, LLMs draft questions, reference
answers, difficulty variants, and adversarial cases. Researchers verify that
each item isolates its intended capability and revise ambiguous or confounded
items. At the scoring level, LLMs serve two roles. First, they translate
evaluation specifications into candidate answer extractors, validators, static
analyses, and task-specific tests; researchers validate and freeze these
components before evaluation. Second, at run time, an LLM serves as the
automated judge for rubric-based probes.

\subsection{Evaluation Results}

\Cref{fig:ace-category-heatmap} compares ZGCM-7B with the reference models
across all 18 capability categories.

\newcommand{\acecategoryheatmapfigure}{%
  \centering
  \includegraphics[width=\textwidth]{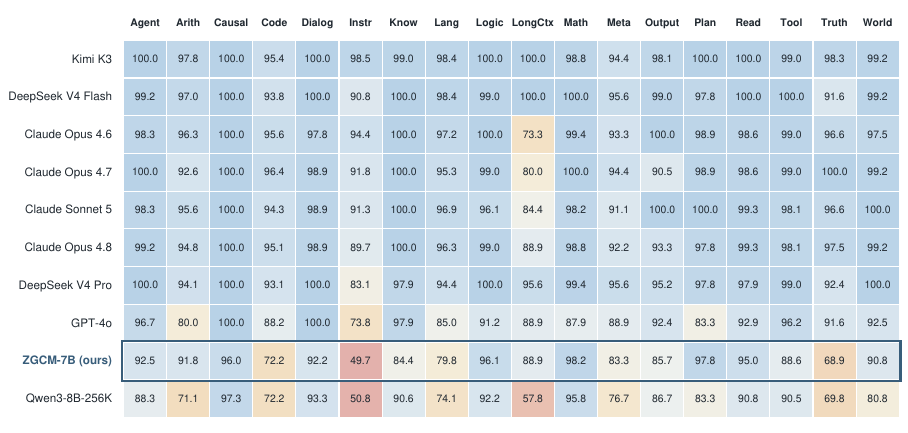}
  \caption{Category-level \ACE{} performance across models. Rows denote
  models and columns denote the 18 evaluated categories; each cell reports the
  corresponding pass rate (\%).}
  \label{fig:ace-category-heatmap}%
}

\makeatletter
\if@twocolumn
 \begin{figure*}[t]
  \acecategoryheatmapfigure
 \end{figure*}
\else
 \begin{figure}[H]
  \acecategoryheatmapfigure
 \end{figure}
\fi
\makeatother

ZGCM-7B performs best in mathematics (98.18\%), planning (97.78\%), logic
(96.08\%), causal reasoning (96.00\%), and reading (95.04\%). Its weakest
categories are instruction following (49.74\%), truthfulness (68.91\%), code
(72.24\%), and language (79.75\%). The heatmap also shows that its gap to the
strongest reference models is concentrated in a small set of categories rather
than distributed across the capability space.

Compared with Qwen3-8B-256K, a model of similar scale, ZGCM-7B shows its largest
advantages in long-context processing (+31.11 percentage points), arithmetic
(+20.74), planning (+14.45), and world modeling (+10.00).

\subsection{Quality and Scope}

The benchmark was refined through repeated scoring audits, cross-model replay,
regression testing, and an independent cross-check. These checks
exposed ambiguities and scoring errors that were repaired before subsequent
evaluation. This refinement process was not fully autonomous: human experts
were required to adjudicate disagreements, verify capability boundaries, inspect
systematic failure patterns, and approve substantive changes to probes and
scorers. In practice, purely model-driven iteration was insufficient because it
could reproduce earlier assumptions or overlook correlated errors; effective
benchmark evolution therefore depended on continued human guidance and
oversight.

\ACE{} measures a structured set of atomic capabilities rather than complete
end-to-end application performance. Its results depend on probe coverage and
scoring quality, and rubric-based LLM judging may retain residual bias. We
therefore use \ACE{} as a diagnostic complement to domain benchmarks and
application-level evaluation, rather than as a single comprehensive measure of
model quality.

\section{AI Autonomy Rubric and Contributor Assessments}
\label{app:ai-autonomy-rubric}

This appendix documents the task-specific autonomy rubric and individual contributor ratings underlying Section~\ref{sec:ai-autonomy-assessment}. The assessment covers 11 task categories grouped into seven R\&D domains. Data engineering comprises data cleaning, data acquisition, and synthetic data generation. Infrastructure engineering comprises operator design, training and inference framework design, and resource platform management. The remaining domains are model architecture design, learning algorithm design, experimentation and monitoring, model evaluation, and deployment engineering.

\subsection{Assessment Scope and Interpretation}
\label{app:ai-autonomy-interpretation}

Nine core contributors each provided one rating for every task category, yielding 99 ratings. Contributors are represented by numeric identifiers in the reporting table, with each identifier referring to the same contributor across all tasks. The assessments capture contributor judgments and should not be interpreted as standardized capability measurements.

The rubric distinguishes five levels according to the allocation of responsibility between humans and AI, the ability to adapt workflows, and the need for human intervention. L1 describes assistance with individual steps; L2 describes execution of predefined workflows; L3 introduces adaptive execution with human approval at consequential decision points; and L4 describes autonomous iteration within human-defined objectives and constraints. L5 extends autonomy to identifying research objectives and coordinating sustained work across R\&D stages.

These levels characterize autonomy rather than productivity, output quality, or scientific novelty. Higher autonomy does not necessarily imply better results, and substantial benefits may arise from assistance at lower levels. L2 behavior may also be implemented through conventional automation; the relevant distinction in an AI assessment is the role played by the agent within that workflow.

L5 serves as a reference for full autonomy, rather than an assertion of demonstrated capability. It does not imply independence from prior knowledge, freedom from operating constraints, or the elimination of external validation. Its defining distinction from L4 is autonomous objective formation and sustained coordination across stages.

The task-specific descriptions below characterize representative behavior at each level. Autonomy demonstrated in a narrow subtask should not be interpreted as equivalent autonomy across an entire task category.

\subsection{Task-Specific Autonomy Criteria}
\label{app:ai-autonomy-criteria}

\subsubsection{Data Cleaning}

\begin{description}
    \item[L1: Basic Assistance.] Humans define cleaning rules and inspect outputs. AI assists with individual regular expressions, formatting operations, or local code corrections.
    \item[L2: Partial Automation.] AI executes predefined cleaning pipelines, including filtering, deduplication, sensitive-data removal, and fixed-format quality reporting.
    \item[L3: Conditional Autonomy.] AI identifies distributional anomalies and low-quality samples, proposes or adapts cleaning strategies, and requests approval before consequential changes to data selection or processing.
    \item[L4: High Autonomy.] Given downstream objectives and constraints, AI develops operational quality criteria and iteratively cleans, selects, and mixes data, validating decisions against downstream outcomes.
    \item[L5: Full Autonomy.] AI identifies evolving data-quality needs and continuously revises cleaning strategies in coordination with acquisition, synthesis, and training, without routine human direction.
\end{description}

\subsubsection{Data Acquisition}

\begin{description}
    \item[L1: Basic Assistance.] Humans identify, select, and integrate data sources. AI assists with source discovery, download scripts, or interface documentation.
    \item[L2: Partial Automation.] AI acquires, parses, and stores data using predefined source lists, interfaces, and collection rules.
    \item[L3: Conditional Autonomy.] AI identifies coverage gaps, discovers candidate sources, and adjusts collection plans. Sources requiring additional authorization or consequential spending are referred to humans.
    \item[L4: High Autonomy.] Given coverage objectives and access constraints, AI independently discovers, evaluates, integrates, and refreshes sources, adjusting acquisition strategies based on observed gaps.
    \item[L5: Full Autonomy.] AI identifies emerging data needs and continuously evolves acquisition workflows in coordination with other R\&D stages, while operating within established access and usage constraints.
\end{description}

\subsubsection{Synthetic Data Generation}

\begin{description}
    \item[L1: Basic Assistance.] Humans design generation templates and quality criteria. AI produces samples in response to individual instructions, with humans selecting the resulting data.
    \item[L2: Partial Automation.] AI generates data at scale using fixed prompts, templates, or generators, followed by predefined filtering and deduplication.
    \item[L3: Conditional Autonomy.] AI selects generation methods based on coverage gaps and iterates on prompts, filters, and data mixtures. Humans approve consequential changes to quality criteria or generation objectives.
    \item[L4: High Autonomy.] Given training objectives, AI autonomously generates, evaluates, filters, and resamples synthetic data, using downstream results to guide successive iterations.
    \item[L5: Full Autonomy.] AI identifies new synthesis objectives and develops generation strategies in coordination with model, algorithm, and evaluation changes, sustaining the process without routine human direction.
\end{description}

\subsubsection{Model Architecture Design}

\begin{description}
    \item[L1: Basic Assistance.] Humans determine model structure, scale, and module configuration. AI provides reference information, code completion, or local implementation suggestions.
    \item[L2: Partial Automation.] AI instantiates established architecture templates, generates configurations and implementation code, and runs predefined architectural variants.
    \item[L3: Conditional Autonomy.] AI proposes, implements, and compares architecture variants based on objectives and experimental results. Major structural or scale changes require human approval.
    \item[L4: High Autonomy.] Given capability targets and resource constraints, AI autonomously searches, implements, trains, and validates architectures through multiple iterations.
    \item[L5: Full Autonomy.] AI identifies architectural research directions and develops model structures in coordination with data, learning algorithms, and infrastructure, without routine human direction.
\end{description}

\subsubsection{Learning Algorithm Design}

\begin{description}
    \item[L1: Basic Assistance.] Humans design training objectives, losses, optimizers, and reinforcement learning methods. AI assists with implementation and local debugging.
    \item[L2: Partial Automation.] AI configures established algorithm templates and evaluates predefined combinations of losses, optimizers, sampling strategies, or reinforcement learning procedures.
    \item[L3: Conditional Autonomy.] AI analyzes training results and proposes algorithmic changes. Consequential changes to learning objectives or training methods require human approval.
    \item[L4: High Autonomy.] Given learning objectives and constraints, AI independently designs, implements, evaluates, and refines learning algorithms across multiple experimental iterations.
    \item[L5: Full Autonomy.] AI identifies algorithmic research questions, develops candidate methods and supporting analyses, and coordinates their validation with other R\&D stages without routine human direction.
\end{description}

\subsubsection{Experimentation and Monitoring}

\begin{description}
    \item[L1: Basic Assistance.] Humans formulate hypotheses, design and launch experiments, and monitor progress. AI assists with code completion or interpretation of individual results, status messages, and errors.
    \item[L2: Partial Automation.] AI launches experiments from fixed templates, runs predefined parameter sweeps, and applies rule-based monitoring, alerts, and retries.
    \item[L3: Conditional Autonomy.] AI designs routine experiments, monitors execution, diagnoses anomalies, and adapts the experimental plan. Major failures, budget changes, and consequential interpretations are referred to humans.
    \item[L4: High Autonomy.] Given research objectives and resource limits, AI autonomously plans, executes, monitors, troubleshoots, and analyzes successive experiments, maintaining records of evidence and conclusions.
    \item[L5: Full Autonomy.] AI identifies research questions, formulates hypotheses, allocates resources within operating constraints, and uses experimental conclusions to determine subsequent research directions.
\end{description}

\subsubsection{Operator Design}

\begin{description}
    \item[L1: Basic Assistance.] Humans design, implement, and debug operators. AI provides code completion, interface explanations, or local performance suggestions.
    \item[L2: Partial Automation.] AI generates or modifies operators using established templates and runs fixed correctness and performance benchmarks.
    \item[L3: Conditional Autonomy.] AI identifies bottlenecks, proposes specialized operators, and implements and validates candidate solutions. Consequential hardware-specific changes require human approval.
    \item[L4: High Autonomy.] Given correctness and performance targets, AI autonomously designs, implements, compiles, tunes, and repairs operators across the specified hardware environments.
    \item[L5: Full Autonomy.] AI identifies operator and hardware co-design opportunities and evolves implementation and optimization strategies alongside changes in models, compilers, and hardware.
\end{description}

\subsubsection{Training and Inference Framework Design}

\begin{description}
    \item[L1: Basic Assistance.] Humans construct frameworks and connect their components. AI assists with configuration, integration code, and error interpretation.
    \item[L2: Partial Automation.] AI configures standard frameworks and pipelines, executing predefined builds, integration tests, and task orchestration.
    \item[L3: Conditional Autonomy.] AI adapts parallelization strategies and runtime configurations to workloads and diagnoses routine framework problems. Major framework changes require human approval.
    \item[L4: High Autonomy.] Given training and inference objectives, AI independently improves framework components and validates changes to parallel execution, compilation, runtime behavior, and pipeline organization.
    \item[L5: Full Autonomy.] AI identifies system design objectives and continuously restructures and validates frameworks across model and hardware changes, coordinating improvements with the broader R\&D workflow.
\end{description}

\subsubsection{Resource Platform Management}

\begin{description}
    \item[L1: Basic Assistance.] Humans request and allocate resources, submit jobs, and maintain environments. AI suggests commands or explains individual failures.
    \item[L2: Partial Automation.] AI executes predefined resource workflows, including environment setup, queueing, job retries, and rule-based preemption.
    \item[L3: Conditional Autonomy.] AI dynamically balances workloads, isolates unhealthy nodes, and migrates jobs. Actions with substantial cross-tenant impact or resource implications require human approval.
    \item[L4: High Autonomy.] Given workload objectives and resource limits, AI forecasts demand, adjusts capacity, recovers from failures, and optimizes placement, communication, and cluster utilization.
    \item[L5: Full Autonomy.] AI identifies platform-level improvement objectives and coordinates scheduling and configuration across clusters and hardware types, continuously adapting the platform to evolving R\&D needs.
\end{description}

\subsubsection{Model Evaluation}

\begin{description}
    \item[L1: Basic Assistance.] Humans select benchmarks, design evaluation procedures, and inspect results. AI assists with individual scripts, metric explanations, or local analysis.
    \item[L2: Partial Automation.] AI runs predefined evaluation pipelines and produces metric comparisons, visualizations, and structured reports.
    \item[L3: Conditional Autonomy.] AI adaptively investigates weaknesses, clusters failure cases, and conducts follow-up diagnostics. Changes to evaluation objectives or consequential judgments require human review.
    \item[L4: High Autonomy.] Given evaluation objectives, AI designs and validates additional tests, conducts adversarial evaluations, and iteratively investigates model weaknesses, including checks for benchmark-specific overfitting.
    \item[L5: Full Autonomy.] AI identifies emerging evaluation needs and continuously develops evaluation criteria and benchmarks in coordination with model development, maintaining validation appropriate to each claim.
\end{description}

\subsubsection{Deployment Engineering}

\begin{description}
    \item[L1: Basic Assistance.] Humans package models, configure environments, deploy services, scale capacity, and perform rollbacks. AI provides scripts or suggestions for individual operations.
    \item[L2: Partial Automation.] AI executes predefined continuous integration and delivery pipelines, serving configurations, health checks, and rollback procedures.
    \item[L3: Conditional Autonomy.] AI selects quantization settings, instance configurations, and rollout strategies based on deployment objectives, and manages staged releases and monitoring. Full releases and consequential rollbacks require human approval.
    \item[L4: High Autonomy.] Given quality, latency, throughput, and cost targets, AI independently performs model compression, quantization, serving optimization, staged deployment, scaling, and rollback within authorized limits.
    \item[L5: Full Autonomy.] AI identifies evolving deployment objectives and continuously adapts the deployment and inference system to model, traffic, and environment changes, coordinating with training, evaluation, and resource management.
\end{description}

\subsection{Individual Contributor Ratings}
\label{app:ai-autonomy-ratings}

Table~\ref{tab:ai-autonomy-individual-ratings} reports the complete contributor ratings. Tasks are presented in the same order as in Figure~\ref{fig:ai-autonomy-assessment}, and columns identify Core Contributors 1--9. The table preserves individual variation that may be obscured by aggregate summaries.

For each task, the figure reports the arithmetic mean and sample standard deviation of the nine ratings after encoding L1--L5 as 1--5. These statistics provide descriptive summaries of ordinal responses; they do not establish equal capability differences between successive levels. Error bars represent variation across contributors rather than confidence intervals. The individual points in the figure correspond directly to the entries below.

\begin{table*}[t]
    \centering
    \small
    \setlength{\tabcolsep}{5pt}
    \renewcommand{\arraystretch}{1.15}
    \caption{Individual assessments of AI autonomy across R\&D tasks. Columns 1--9 denote Core Contributors 1--9, with identifiers held constant across tasks. Ratings follow the task-specific criteria in Appendix~\ref{app:ai-autonomy-criteria}.}
    \label{tab:ai-autonomy-individual-ratings}
    \begin{tabularx}{\textwidth}{@{}X*{9}{c}@{}}
        \toprule
        & \multicolumn{9}{c}{\textbf{Core Contributor}} \\
        \cmidrule(l){2-10}
        \textbf{Task} & \textbf{1} & \textbf{2} & \textbf{3} & \textbf{4} & \textbf{5} & \textbf{6} & \textbf{7} & \textbf{8} & \textbf{9} \\
        \midrule
        Data cleaning & L3 & L2 & L3 & L3 & L2 & L3 & L3 & L2 & L2 \\
        Data acquisition & L4 & L4 & L4 & L4 & L1 & L2 & L4 & L2 & L4 \\
        Synthetic data generation & L4 & L3 & L3 & L3 & L3 & L3 & L3 & L3 & L3 \\
        \addlinespace
        Model architecture design & L3 & L2 & L2 & L2 & L1 & L2 & L2 & L1 & L2 \\
        \addlinespace
        Learning algorithm design & L3 & L2 & L2 & L3 & L2 & L1 & L2 & L1 & L2 \\
        \addlinespace
        Experimentation \& monitoring & L5 & L4 & L4 & L4 & L4 & L4 & L4 & L4 & L3 \\
        \addlinespace
        Operator design & L4 & L4 & L4 & L3 & L4 & L3 & L3 & L4 & L4 \\
        Training \& inference framework design & L3 & L3 & L3 & L2 & L2 & L3 & L2 & L3 & L2 \\
        Resource platform management & L4 & L4 & L3 & L4 & L2 & L4 & L4 & L4 & L3 \\
        \addlinespace
        Model evaluation & L4 & L4 & L4 & L4 & L1 & L4 & L4 & L4 & L3 \\
        \addlinespace
        Deployment engineering & L4 & L4 & L4 & L4 & L2 & L4 & L4 & L4 & L4 \\
        \bottomrule
    \end{tabularx}
\end{table*}

\section{Evaluation Reproducibility and Provenance}
\label{app:reproducibility}

Reproducibility requires binding a numerical result to four layers at once:
the model artifact, the evaluation data, the invocation contract, and the
scoring implementation. For each final benchmark, the evaluation record should
preserve the model hash, official dataset URL and immutable revision, split,
prompt wrapper, few-shot examples, system prompt, chat template, mode-control
setting, answer extraction, metric implementation, random seed, timeout,
inference engine and revision, hardware, and raw-output location. External
comparison values are taken from the cited releases rather than a unified
rerun.

We evaluate \zgcm{} in no-think mode using deterministic decoding with
temperature 0, top-\(p=1.0\), and one run.

\begin{table}[H]
  \centering
  \caption{Selected no-think benchmark results (\%) for \zgcmeval{} and
  selected 7B--8B models. Dark blue cells in bold indicate the best result;
  light blue cells indicate the second-best result.}
  \label{tab:appendix-nothink-comparison}
  \resizebox{\textwidth}{!}{%
  \begin{tabular}{lcccc}
    \toprule
    Benchmark
      & \multicolumn{1}{c}{\zgcmeval{}}
      & \multicolumn{1}{c}{Olmo 3 7B Instruct}
      & \multicolumn{1}{c}{Qwen3-8B (non-thinking mode)}
      & \multicolumn{1}{c}{Qwen2.5-7B} \\
    \midrule
    \multicolumn{5}{l}{\textit{Reasoning} \& \textit{General}} \\
    MATH-500     & \bestscore{89.40} & \secondscore{87.30} & 82.30 & 71.00 \\
    AIME 2024    & \secondscore{33.33} & \bestscore{44.30} & 26.20 & 11.30 \\
    AIME 2025    & \bestscore{43.33} & \secondscore{32.50} & 21.70 & 6.30 \\
    BBH          & 69.48 & \secondscore{71.20} & \bestscore{73.70} & 68.80 \\
    \midrule
    \multicolumn{5}{l}{\textit{Code}} \\
    HumanEval+       & 66.46 & \secondscore{77.20} & \bestscore{79.80} & 74.90 \\
    MBPP+            & \bestscore{75.93} & 60.20 & \secondscore{64.40} & 62.60 \\
    LiveCodeBench v3 & 34.31 & 29.50 & \bestscore{53.20} & \secondscore{34.50} \\
    \midrule
    \multicolumn{5}{l}{\textit{Knowledge}} \\
    MMLU & 70.03 & 69.10 & \bestscore{80.40} & \secondscore{77.20} \\
    GPQA-Diamond & \secondscore{42.42} & 40.40 & \bestscore{44.60} & 35.60 \\
    \midrule
    \multicolumn{5}{l}{\textit{Instruction Following}} \\
    IFEval  & 78.00 & \secondscore{85.60} & \bestscore{86.30} & 73.40 \\
    IFBench~\citep{pyatkin2025ifbench} & \secondscore{31.67} & \bestscore{32.30} & 29.30 & 28.40 \\
    \bottomrule
  \end{tabular}%
  }
\end{table}

\end{document}